%% file: main.tex
\pdfoutput=1
\documentclass[11pt]{article}
\usepackage{amssymb}
\usepackage{amsmath}

\usepackage[preprint]{acl}

\usepackage{times}
\usepackage{CJKutf8}
\usepackage{latexsym}
\usepackage{wrapfig}
\usepackage{textcase}
\usepackage{hyperref}
\usepackage{cleveref}
\usepackage{float}
\usepackage{booktabs}
\usepackage{xcolor}
\usepackage{bm}
\usepackage{soul}
\usepackage{xcolor}
\usepackage{makecell}
\usepackage{enumitem}
\usepackage{placeins}

\newcommand{\valc}[2]{#1\textsubscript{\scriptsize \,±#2}}
\newcommand{\valbc}[2]{\textbf{#1}\textsubscript{\scriptsize \,±#2}}
\newcommand{\valuc}[2]{\underline{#1}\textsubscript{\scriptsize \,±#2}}

\usepackage{tcolorbox} % preamble에 추가 필요
\tcbset{
    colback=white, colframe=black, arc=3mm, boxrule=0.8pt,
    left=5pt, right=5pt, top=5pt, bottom=5pt
}
\tcbuselibrary{breakable}

\usepackage[T1]{fontenc}
\usepackage[utf8]{inputenc}

\usepackage{microtype}

\usepackage{inconsolata}

\usepackage{graphicx}
\usepackage{subcaption}
\usepackage{multirow}
\usepackage{algorithm}
\usepackage{algorithmic}
\usepackage{xcolor}
\usepackage{capt-of}  
\usepackage{multicol}
\usepackage{caption}

\usepackage{CJKutf8}

\newcommand{\myframework}{MindTailor}
\newcommand{\mydataset}{ReddiSupp}

\definecolor{Reframing}{RGB}{46, 139, 87}     % ForestGreen
\definecolor{Regard}{RGB}{204, 85, 0}    % Burnt Orange
\definecolor{Solution}{RGB}{138, 43, 226}  % Violet

\title{
MindTailor: Personalized Emotional Support via Post History-Grounded Case Formulation and Collaborative Refinement
}

\author{Suhyun Han \\
  Sungkyunkwan University \\
  Suwon \\
  South Korea \\
  \texttt{gkstngus01@skku.edu} \\\And
  Kyunghyun Cho \\
  New York University \\
  New York \\
  USA \\
  \texttt{kyunghyun.cho@nyu.edu} \\\And
  JinYeong Bak \\
  Sungkyunkwan University \\
  Suwon \\
  South Korea \\
  \texttt{jy.bak@skku.edu} \\}

\begin{document}
\maketitle
\begin{abstract}

\input{latex/000_abstract}
\end{abstract}

\section{Introduction}
\label{sec:introduction}

\input{latex/010_introduction}

\section{Related Work}
\label{sec:related_work}

\input{latex/020_related_work}

\section{\myframework}
\label{sec:methodology}

\input{latex/030_methodology}
\section{\mydataset}
\label{sec:dataset}

\input{latex/040_dataset}

\section{Experiments}
\label{sec:experiments}

\input{latex/050_experiments}

% \section{Ablation Study}
% \label{sec:ablation_study}
% \input{latex/055_ablation_study}

\section{Analysis}
\label{sec:analysis}

\input{latex/060_analysis}

% \section{User Study}
% \label{sec:user_study}
% \input{latex/070_user_study}

% \section{Case Study}
% \label{sec:case_study}
% \input{latex/080_case_study}

\section{Conclusion}
\label{sec:conclusion}

\input{latex/100_conclusion}

\newpage
\section*{Limitations}

Our dataset comprises posts from 2020-2022, a period shaped by the COVID-19 pandemic, which may have influenced the nature and expression of mental health concerns. The one-year retrospective window for post history, while designed to ensure contextual relevance, may also miss longer-term patterns in some seekers' experiences.

\myframework~employs three counseling strategies which do not cover the full spectrum of therapeutic approaches. Other evidence-based strategies such as motivational interviewing~\cite{miller2012motivational} or dialectical behavior therapy techniques~\cite{linehan1993cognitive} may further enhance response quality for specific populations.

The iterative collaborative refinement process introduces computational overhead that may limit real-time deployment, as each response requires multiple LLM calls across agents and refinement rounds. Adaptive strategies such as early stopping, which terminates refinement once agent outputs converge, offer a promising direction for reducing this cost.

Additionally, we observed occasional failure cases in case formulation, including quote manipulation (merging separate seeker statements into fabricated quotes), risk underestimation (assigning low risk to posts with explicit crisis language), and temporal hallucination (e.g., a systematic +1 year date shift). These errors point to the need for stricter grounding mechanisms in future iterations. For safety-sensitive cases, integrating dedicated crisis detection components alongside the case formulation pipeline remains an important direction for future work.

Finally, our evaluation is conducted exclusively on \mydataset, which is constructed from Reddit. Other platforms such as Twitter/X and TalkLife differ in post length, anonymity norms, and community conventions, which may affect both support-seeking behavior and response effectiveness.
Validating \myframework~across these platforms remains an important direction for future work.

\section*{Ethical Considerations}

\paragraph{Licenses and Terms of Use}
All target posts in \mydataset~were created between January 2020 and December 2022. The data was collected using the Pushshift API~\cite{baumgartner2020pushshift}, which provided public access to Reddit data for academic research at the time of collection. Our data collection was completed prior to Reddit's introduction of the Public Content Policy in May 2024 \cite{reddit_policy_change}, which now requires formal agreements for data access. All data and models used in this work were employed in accordance with their respective terms of service and licenses for non-commercial academic research purposes. Detailed license information for the language models used is provided in Table~\ref{tab:models}.

\paragraph{Privacy and Data Protection}
Reddit usernames are pseudonymous, but original posts may still be retrievable via search engines. To balance the risk of re-identification with the need for reproducibility, we release \mydataset~under a data usage agreement, the details of which are described in Appendix~\ref{apdx:dataset_access}. We manually reviewed a random sample of the dataset to check for personally identifiable information (e.g., real names, contact information) and found no such instances, as Reddit's community norms discourage sharing personal details. Furthermore, all examples presented in this paper have been rephrased to prevent direct retrieval of the original posts while preserving their semantic content. Due to the nature of mental health support data, the dataset contains emotionally sensitive content, including discussions of self-harm and suicidal ideation. This is inherent to the research domain and necessary for developing effective emotional support systems.

\paragraph{Intended Use and Limitations}
\mydataset~and \myframework~are intended solely for academic research on emotional support generation. They should not be deployed in real-world mental health services without professional oversight and appropriate ethical review, as automated responses may not adequately address the complex needs of vulnerable individuals.

We acknowledge several potential risks of this work. Model-generated responses may be inappropriate or even harmful in crisis situations such as suicidal ideation, where professional intervention is essential. There is also a risk that seekers may over-rely on automated support, potentially delaying professional help-seeking. Furthermore, the techniques developed here could be misused to manipulate emotionally vulnerable individuals. To mitigate these concerns, any deployment should include crisis detection mechanisms, clear disclaimers that the system does not replace professional care, and continuous human oversight.

\paragraph{IRB Approval}
All research procedures involving human participants were approved by our Institutional Review Board (IRB).\footnote{IRB Approval Number: SKKU 2025-06-033, 2025-09-011} The human evaluation and user study were reviewed and approved under separate protocols. Detailed information about participant recruitment, compensation, informed consent, and safeguards is provided in Appendix~\ref{apdx:human_eval_details} and Appendix~\ref{apdx:user_study_details}.

% \section*{Acknowledgments}

% Bibliography entries for the entire Anthology, followed by custom entries
%\bibliography{anthology,custom}
% Custom bibliography entries only
\bibliography{custom}

\clearpage

\appendix

\section{Counseling Strategy}
\label{apdx:counseling_strategy}

\input{latex/appendix/101_counseling_strategy}

\section{\mydataset~Construction Procedures}
\label{apdx:dataset_construction}
\input{latex/appendix/105_dataset_construction}

\input{latex/table/110_dataset_statistics}

\section{\mydataset~Statistics}
\label{apdx:dataset_statistics}

\input{latex/appendix/110_dataset_statistics}

\section{\mydataset~Access and Usage Agreeement}
\label{apdx:dataset_access}
\input{latex/appendix/115_dataset_access}

% \section{Seeker-Aware Win Rate}
% \label{apdx:seeker_aware_win_rate}
% \input{latex/appendix/135_seeker_aware_win_rate}

\section{Framework Prompts}
\label{apdx:framework_prompts}
\input{latex/appendix/190_framework_prompts}

\section{Additional Evaluation}
\label{apdx:additional_evaluation}

\input{latex/appendix/123_additional_evaluation}

\section{Reliability of LLM-as-a-Judge Evaluation}
\label{apdx:llm_reliability}

\input{latex/appendix/193_llm_as_a_judge_reliability}

\section{LLM-as-a-Judge Prompts}
\label{apdx:evaluation_prompts}

\input{latex/appendix/195_evaluation_prompts}

\section{Human Evaluation Details}
\label{apdx:human_eval_details}

\input{latex/appendix/125_human_evaluation_guideline}

\section{User Study Details}
\label{apdx:user_study_details}

\input{latex/appendix/150_user_study_guideline}

\input{latex/table/170_implementation_details_model_info}

\section{Additional Case Study}
\label{apdx:case_study}

\input{latex/appendix/160_case_study}

% \section{Evaluation across Refinement Rounds}
% \label{apdx:evaluation_across_refinement_rounds}
% \input{latex/appendix/140_evaluation_across_refinement_rounds}

% \section{Strategy-Specific Score across Refinement Rounds}
% \label{apdx:counseling_strategy_across_refinement_rounds}
% \input{latex/appendix/145_counseling_strategy_across_rounds}

\section{Implementation Details}
\label{apdx:implementation_details}

\input{latex/appendix/170_implementation_details}

% \section{Analysis of Refinement Rounds}
% \label{apdx:analysis_refinement_rounds}
% \input{latex/appendix/175_refinement_rounds}

\section{Computational Costs}
\label{apdx:computational_costs}

\input{latex/appendix/180_computational_costs}

\FloatBarrier
\input{latex/figure/case_formulation_prompt}
\input{latex/figure/draft_generation_prompt}
\input{latex/figure/critique_generation_prompt}
\input{latex/figure/guideline_synthesis_prompt}
\input{latex/figure/response_refinement_prompt}

\input{latex/figure/pairwise_comparion_prompt}

\input{latex/figure/empathy_prompt}
\input{latex/figure/helpfulness_prompt}
\input{latex/figure/understanding_prompt}

\input{latex/figure/personalization_prompt}

\FloatBarrier
\input{latex/table/160_case_study_table}

% \section{Prompts in \myframework}
% \label{apdx:framework_prompts}
% \input{latex/appendix/120_framework_prompts}

% \section{\mydataset~Construction}
% \label{apdx:data_construction}
% \input{latex/appendix/130_data_construction}

\end{document}

%% file: latex/000_abstract.tex
% As mental health issues globally escalate, there is a growing need for personalized digital support systems.
% While recent work has explored personalization in emotional support systems by leveraging seekers' emotional states and personas, these approaches primarily capture current characteristics while overlooking underlying factors that shape present mental health concerns.
% In this work, we propose \myframework, a collaborative critique-based framework that generates personalized emotional support responses leveraging case formulation and iterative refinement.
% Drawing on case formulation, a structured clinical method that integrates a seeker's current state with relevant past experiences, \myframework~ leverages post history to understand underlying factors of emotional distress. 
% Inspired by the human writing process, we employ multiple specialized counselor agents that iteratively refine responses through collaborative critique, with controlled feedback to ensure effective revision.
% To enable research on history-aware emotional support generation, we construct \mydataset, a dataset of 798 Reddit posts paired with seekers' posting histories. 
% Through comprehensive evaluation via LLM-as-a-Judge, expert evaluation with psychology professionals, and a user study with individuals who have sought emotional support online, we demonstrate that \myframework~outperforms baselines in terms of empathy, personalization, and understanding.

As mental health concerns continue to rise globally, social media has emerged as a vital space where individuals seek emotional support.
While prior work on personalized emotional support has leveraged seekers' emotional states, personas, and situational context, these approaches primarily capture the seeker's current state, overlooking the formative experiences that shape present concerns.
In this work, we propose \myframework, a framework that generates personalized emotional support responses by constructing a case formulation from the seeker's post history and iteratively refining responses through collaborative critique among counselor agents grounded in distinct counseling strategies.
To enable research on this history-aware task, we construct \mydataset, a dataset of 798 Reddit posts paired with seekers' prior post histories.
Through LLM-as-a-Judge evaluation, expert human evaluation, and a user study with seekers, we demonstrate that \myframework~ outperforms baselines across these evaluations, improving empathy, personalization, understanding, and achieving the highest overall preference.

%% file: latex/010_introduction.tex
As mental health concerns continue to rise globally~\cite{who2022mentaldisorder_covid, who2022mentaldisorder, who2025mentaldisorder}, social media has become a vital space for individuals in emotional distress, offering anonymity and a non-judgmental environment that lowers barriers to self-disclosure~\cite{naslund2014, luo2020}.
On these platforms, individuals referred to as \textit{seekers} share their struggles and seek \textit{emotional support}, the communicative behaviors that convey care and empathy to help others cope with stress.
We define a \textit{target post} as a post in which a seeker describes emotional struggles and implicitly or explicitly seeks support.
In this work, we tackle the task of generating personalized emotional support responses for target posts on social media, and propose \myframework, a framework grounded in psychotherapy and clinical practice.

While prior studies have leveraged various seeker characteristics for personalized emotional support, including emotional states~\cite{tu-etal-2022-misc, PENG2023110340, chen2024cauesccausalawaremodel}, persona~\cite{cheng-etal-2023-pal, cheng2025autopalautonomousadaptationusers}, situational context~\cite{liu-etal-2021-towards}, and cognitive states~\cite{xu2024dynamicdemonstrationretrievalcognitive}, they primarily capture the seeker's current state while overlooking underlying psychological factors and formative experiences that shape it~\cite{Hofmann2014, johnstone2006formulation, Felitti1998, Zarse01012019}.
To address this, we leverage the seeker's post history, a distinctive affordance of social media, to construct a case formulation~\cite{Sim2005, eells2022handbook} that integrates current and historical context for deeper understanding and more personalized responses.

Emotional support response generation is inherently complex, requiring simultaneous consideration of emotional understanding, empathetic expression, problem exploration, and actionable guidance~\cite{liu-etal-2021-towards}. 
Since emotional functioning involves the interplay of cognition, affect, and behavior~\cite{Ellis2019}, a single perspective is insufficient for emotional support.
To address it, drawing on multidisciplinary case conferences in clinical practice~\cite{unutzer-etal-2002-collaborative}, we employ multiple counselor agents specializing in different counseling strategies to critique the response from diverse perspectives.
Inspired by the human writing process of drafting and iteratively refining~\cite{flower1981cognitive, du-etal-2022-understanding-iterative}, we incorporate these critiques through iterative refinement,  selecting only the most impactful suggestions at each step. We empirically find that this design choice prevents quality degradation from over-correction.

% we incorporate these critiques through iterative refinement, limiting the number reflected in each step to avoid the counterproductive effects of excessive feedback~\cite{huang2024largelanguagemodelsselfcorrect}.
% Since excessive feedback can be counterproductive, we limit the number of critiques reflected in each refinement step.

% In this paper, we propose \myframework, a novel collaborative critique-based framework that generates personalized emotional support responses grounded in clinical psychology principles and approaches, including case formulation and iterative refinement. 
% In this paper, we propose \myframework, a novel collaborative critique-based framework that generates personalized emotional support responses leveraging case formulation and iterative refinement.
To facilitate research on this task, we construct \mydataset, a new dataset that pairs each target post with the seeker's prior post history, addressing a key limitation of existing datasets that lack such history~\cite{liu-etal-2021-towards, bertagnolli2020counsel}.
% To facilitate research on this task, we construct a new dataset, \mydataset, as existing benchmarks either focus on multi-turn conversations~\cite{liu-etal-2021-towards} or lack post history~\cite{bertagnolli2020counsel}. 
% We collect 798 instances from Reddit, each pairing a seeker's post history with their target post.

\myframework~consistently outperforms all baselines in pairwise comparisons using LLM-as-a-Judge~\cite{10.5555/3666122.3668142} across diverse backbone and judge models. 
In expert human evaluation, \myframework~achieves the highest scores in empathy, personalization, understanding, and overall preference, and in our user study, seekers most prefer \myframework~across all dimensions.
These results indicate that \myframework~generates well-balanced responses that are both clinically sound and genuinely valued by seekers.

Our primary contributions are as follows:
\begin{itemize}
    \item We propose \myframework, a framework that generates personalized emotional support by constructing a case formulation from the seeker's post history and refining responses through multi-agent collaborative critique.
    \item We construct \mydataset, a dataset of 798 Reddit posts paired with seekers' post histories, enabling research on personalized emotional support grounded in seekers' broader context.
    \item We validate \myframework~through LLM-as-a-Judge, expert human evaluation, and a user study with seekers, demonstrating effectiveness in empathy, personalization, understanding, and overall preference.
\end{itemize}

%% file: latex/020_related_work.tex
\subsection{Personalized Emotional Support}
Emotional support response generation aims to alleviate the emotional distress of seekers through supportive responses~\cite{liu-etal-2021-towards}. To generate effective and personalized responses, numerous studies have attempted to leverage various characteristics of the seeker.
A prominent line of research focuses on modeling the seeker's emotional state, using commonsense knowledge to infer fine-grained emotions~\cite{tu-etal-2022-misc}, modeling emotion causes and dynamics~\cite{chen2024cauesccausalawaremodel}, tracking turn-level emotion transitions~\cite{zhao-etal-2023-transesc}, or capturing emotional changes through feedback mechanisms~\cite{PENG2023110340}.
Another direction emphasizes the seeker's persona, including dynamically inferring persona from dialogue history~\cite{cheng-etal-2023-pal}, extracting consistent persona through semantic similarity~\cite{han2024personaextractionsemanticsimilarity}, adapting persona during interaction~\cite{cheng2025autopalautonomousadaptationusers}, or leveraging rich user profiles with tool-augmented responses to deliver personalized social support~\cite{huang2026compass}.
Researchers have also explored situational context, such as the problem situation faced by the seeker~\cite{liu-etal-2021-towards, electronics13081484}, and cognitive states, leveraging commonsense knowledge to deepen situational awareness~\cite{xu2024dynamicdemonstrationretrievalcognitive, zhang2025intentionescintentioncenteredframeworkenhancing}. 
Beyond these characteristics, other approaches tailor support strategies through multi-agent collaboration, such as debate among agents~\cite{lee2024} or staged multi-agent frameworks~\cite{xu-etal-2025-multiagentesc}, while another line of work leverages human values to guide responses~\cite{kim-etal-2025-dialogue}.

While these approaches have made significant progress, they primarily capture the seeker's characteristics at the current moment, overlooking the formative experiences that shape present concerns~\cite{Felitti1998, Zarse01012019, Hofmann2014}.
We address this gap by leveraging seekers' post history, uniquely available in social media settings, for more personalized emotional support.

\begin{figure*}[t!]
    \centering
    \includegraphics[width=\linewidth]{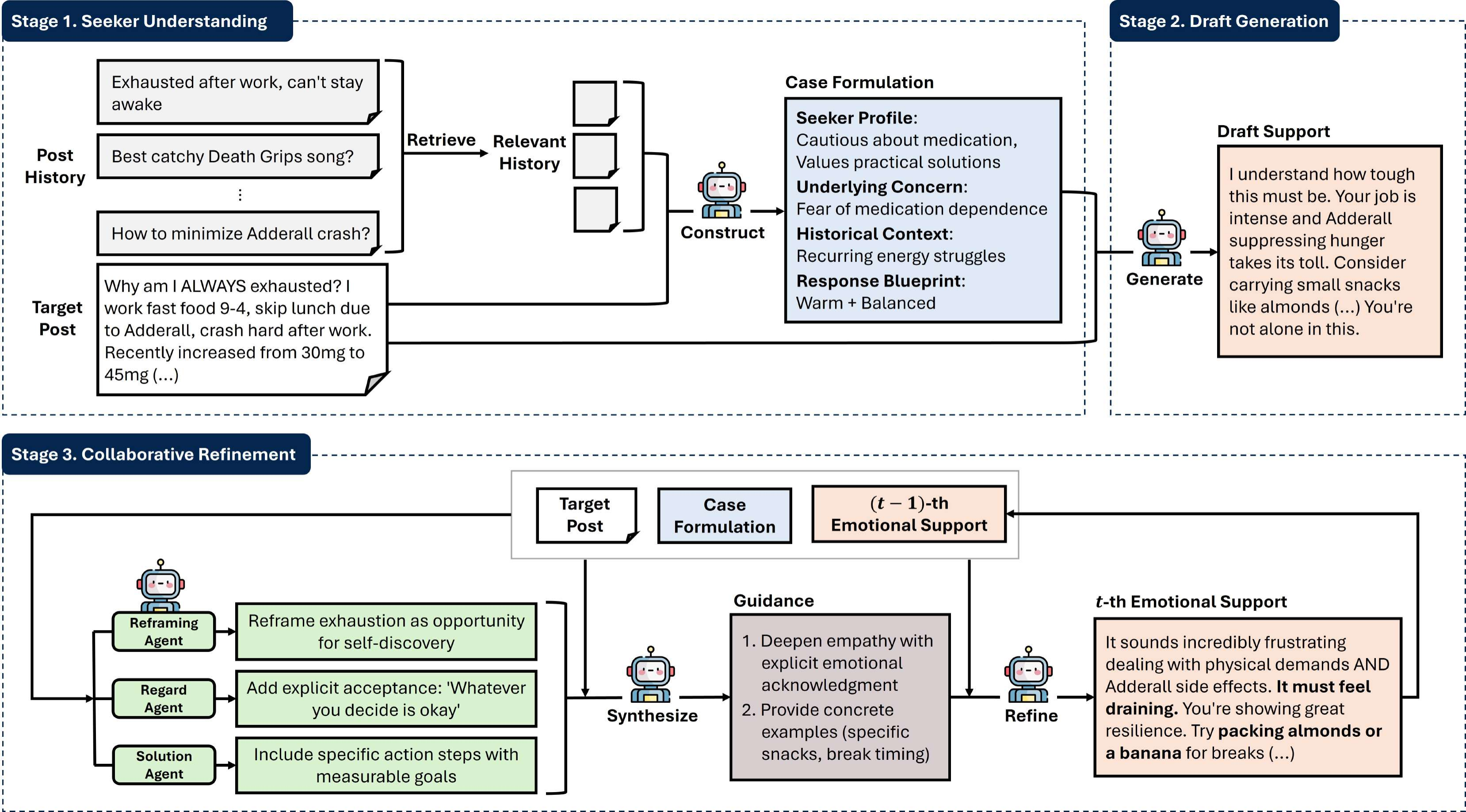}
    \caption{Overview of \myframework.
     Given a seeker's target post and post history, our framework generates a personalized emotional support response through three sequential stages: (1) Seeker Understanding, which retrieves relevant posts from the history and constructs a four-dimensional case formulation; (2) Draft Generation, which produces an initial response grounded in the case formulation; and (3) Collaborative Refinement, in which three counselor agents grounded in Cognitive Reframing, Unconditional Positive Regard, and Solution-Focused counseling (shown as Reframing, Regard, Solution) critique the draft from complementary cognitive, affective, and behavioral perspectives, and their critiques are synthesized into focused guidance to iteratively refine the response.
    % consisting of three stages: (1) Case Analysis: Relevant posts $H_k$ are filtered from the seeker's post history $H$ and analyzed to construct a case formulation $C$. (2) Initial Response Generation: An initial emotional support response $r_0$ is generated based on the target post $p$ and case formulation $C$. (3) Iterative Collaborative Refinement: Three counselor agents independently critique the current emotional support, and their feedback is synthesized into focused guidance to produce a refined emotional support.
    % The right part of the figure illustrates how the response evolves from $r_0$ to $r_t$ through iterative refinement.
    }

    \label{fig:framework_overview}
\end{figure*}

% iterative refinement를 사용한 논문들의 introduction 참조
% iterative text refinement 논문들은 해당 방식을 사용하는 근거로 인간의 행동 특성을 근거로 함
% 초안을 작성하고 이를 정제하면서 완성 시켜나가는 행동, 복잡한 텍스트 작성 작업을 할 때 이러한 과정을 거침
% 이러한 방식을 참고하여 여러 논문에서 iterative text refinement 방법론을 제시함
% emotional support 또한 복합적인 요소를 고려해야 하는 태스크이므로 iterative refinement 방식을 채택
% 여기서 다양한 관점을 고려할 수 있도록 multi-agent 방식을 사용했고, 효과적인 refinement가 이루어지도록 multi-agent의 critique를 종합하고 중요한 critique만을 필터링하는 critique synthesis 방식을 설계함 -> 이 내용이 여기에 넣기에 적절한가..?

\subsection{Iterative Refinement}
Inspired by the human cognitive process of producing high-quality text through initial drafting followed by continuous refinement \cite{du-etal-2022-understanding-iterative, flower1981cognitive}, iterative text refinement has been widely adopted in natural language generation. 
Early approaches relied on a single model that generates self-feedback and iteratively refines outputs \cite{10.5555/3666122.3668141}, stores verbal reflections in episodic memory to improve subsequent trials \cite{10.5555/3666122.3666499}, leverages external tools for verification and correction \cite{gou2024critic}, or applies self-critique based on constitutional principles \cite{bai2022constitutionalaiharmlessnessai}.
More recent work moves beyond single-model refinement by adopting multi-agent collaboration, where diverse agents jointly critique and revise outputs to mitigate the limitations of self-feedback, such as improving faithfulness through multi-agent, multi-model collaboration across refinement subtasks \cite{wan2025mamm} or applying coarse-to-fine refinement guided by error localization across agents \cite{chen2025magicore}.

Building on this line of work, we adopt iterative refinement for emotional support generation, obtaining critiques from multiple counselor agents. To keep each refinement step focused on the most impactful changes, a dedicated agent synthesizes critiques into a small set of high-priority items before each refinement step.

%% file: latex/030_methodology.tex
We propose \myframework, which generates personalized emotional support responses from a seeker's target post and post history in three stages: (1) \textbf{Seeker Understanding}, (2) \textbf{Draft Generation}, and (3) \textbf{Collaborative Refinement}.

% \paragraph{Stage 1: Case Formulation.}
\subsection{Stage 1: Seeker Understanding}

The first stage constructs a case formulation~\cite{eells2022handbook} to gain a comprehensive understanding of the seeker, achieved by examining the target post alongside their post history.

% \textbf{History Filtering.}
\paragraph{History Retrieval.}
Since the post history may contain posts irrelevant to the target post, we retrieve a subset of the most relevant posts. Using text-embedding-3-small \cite{openai_text_embedding_3_small}, we compute semantic embeddings for the target post and each post in the history, then select the top-$k$ posts based on cosine similarity as the relevant history.

% \textbf{Formulation Construction.}
\paragraph{Case Formulation Construction.}
From the target post and the relevant history, we construct a structured case formulation.
Drawing on psychotherapy case formulation frameworks~\cite{eells2022handbook, persons2012case, johnstone2006formulation}, the formulation comprises four dimensions:
(1) \textit{Seeker Profile}, capturing predisposing and protective factors such as self-concept, values, and personal strengths~\cite{eells2022handbook, persons2012case}; 
(2) \textit{Underlying Concern}, characterizing the seeker's current problem and the underlying emotional mechanisms and implicit needs beneath it~\cite{beck1979cognitive, Hofmann2014}; 
(3) \textit{Historical Context}, identifying precipitating events and maintaining patterns from the seeker's history~\cite{johnstone2006formulation, Sim2005}; and 
(4) \textit{Response Blueprint}, linking assessment to intervention through actionable guidance on tone, validation, and personalization~\cite{eells2022handbook}.

% \paragraph{Stage 2: Draft Generation.}
\subsection{Stage 2: Draft Generation}
The second stage produces a draft response that reflects the seeker's personal context identified in the case formulation.

% \paragraph{Stage 3: Collaborative Refinement.}
\subsection{Stage 3: Collaborative Refinement}
The third stage iteratively refines the response by incorporating feedback from complementary therapeutic perspectives.
Inspired by multidisciplinary clinical case conferences~\cite{unutzer-etal-2002-collaborative}, this design aims to produce more balanced, higher-quality responses through diverse viewpoints.
Each iteration consists of the following three sub-steps.

% \textbf{Critique Generation.}
\paragraph{Critique Generation.}
We employ three counselor agents following \citet{lee2024}. This design is grounded in the principle that cognition, affect, and behavior are mutually influential in psychological functioning~\cite{Ellis2019}: Cognitive Reframing~\cite{Beck1989} targets cognitive patterns, Unconditional Positive Regard~\cite{standar_regard} provides affective validation, and Solution-Focused Counseling~\cite{Bannink2007} offers behavioral guidance.
The three agents independently critique the current response from their respective perspectives.

% \textbf{Guidance Synthesis.}
\paragraph{Guidance Synthesis.}
The critiques are aggregated, and at most two high-priority items are selected to produce focused guidance.This keeps each refinement step focused on the most impactful changes, avoiding the dilution that can occur when too many revisions are attempted simultaneously.

% at most two high-priority items are selected to produce focused guidance.
% This prevents quality degradation often observed when models attempt complex self-correction without focused guidance \cite{huang2024largelanguagemodelsselfcorrect}.

% \textbf{Response Refinement.}
\paragraph{Response Refinement.}
Finally, the response is then refined based on the synthesized guidance.

%% file: latex/040_dataset.tex
To enable systematic evaluation of \myframework~on this history-aware emotional support generation task, we constructed \mydataset.
Existing benchmarks typically focus on either multi-turn conversational interactions~\cite{liu-etal-2021-towards} or single-turn responses conditioned solely on the target post~\cite{bertagnolli2020counsel}.
However, multi-turn benchmarks capture only within-session dialogue context, and single-turn settings lack context entirely. Neither format provides access to the seeker's broader history beyond the immediate interaction.
In contrast, \mydataset~pairs target posts with the seeker's prior post history, providing the historical context necessary for research on history-aware personalized emotional support.

\mydataset~is constructed from publicly available Reddit data collected via the Pushshift API \cite{baumgartner2020pushshift}, drawing from mental health-related subreddits used in prior work~\cite{low2020natural}. Each instance consists of a target post, where a seeker expresses emotional distress, paired with their post history from the preceding year. We applied filtering criteria to ensure target posts are semantically rich and genuinely warrant emotional support, and retained only instances with at least two prior posts to guarantee sufficient historical context. The final dataset comprises 798 target posts paired with their associated post histories. Detailed construction procedures and dataset statistics are provided in Appendix~\ref{apdx:dataset_construction} and Appendix~\ref{apdx:dataset_statistics}.

%% file: latex/050_experiments.tex
\input{latex/table/050_experiments_spc}

We conduct comprehensive experiments to evaluate the effectiveness and psychological quality of \myframework's responses.

% \subsection{Evaluation Metrics}
\subsection{Evaluation Approaches}
\label{sec:evaluation_approaches}

\paragraph{LLM-as-a-Judge~\cite{10.5555/3666122.3668142}.}
To assess whether \myframework's response aligns with the seeker's preferences inferred from their post history, we conduct pairwise comparisons against baselines using gpt-4o-mini~\cite{openai2024gpt4omini}, deepseek-v4-flash~\cite{deepseekai2026deepseekv4} and glm-4.5-air~\cite{zeng2025glm} as judges, reporting win, loss, and tie rates.
We mitigate position bias~\cite{wang-etal-2024-large-language-models-fair} by running three evaluations in each of the original and reversed orders.

\paragraph{Expert Human Evaluation.}
To assess whether responses meet professional psychological standards, we recruit five evaluators with graduate-level training in psychology.
Each evaluator rates responses on a 5-point Likert scale across four dimensions: 
Empathy, Helpfulness, Personalization, and Understanding \cite{dey2025gravityframeworkpersonalizedtext,ye2025genericempathypersonalizedemotional}. 
They also provide an overall ranking indicating which response the seeker would most likely prefer.
Details are provided in Appendix~\ref{apdx:human_eval_details}.

\paragraph{User Study with Seekers.}
The preceding evaluations rely on third-party assessments that may not fully capture the preferences of those actually seeking support.
To address this, we conduct a user study with 50 seekers who evaluate responses to their own posts across six dimensions using a 5-point Likert scale: Empathy, Perceived Helpfulness, Personalization, Relevance, Trustworthiness \& Safety, and Willingness to Use, along with an overall ranking~\cite{liu-etal-2021-towards,Abbasian2024,app14135889,informatics12010033}.
Details are provided in Appendix~\ref{apdx:user_study_details}.

\subsection{Baselines}
\label{sec:baselines}

We select baselines spanning three paradigms: direct prompting with full context, multi-agent collaboration, and value-driven generation.

\paragraph{Vanilla.}
This baseline employs standard prompting with both post history and target post as input, explicitly instructing the model to generate personalized emotional support based on characteristics inferred from the seeker's post history.

\paragraph{MentalAgora~\cite{lee2024}.}
A multi-agent framework where counselor agents with distinct strategies collaborate through debate to synthesize personalized responses from the target post. While both approaches leverage multiple counseling perspectives, MentalAgora uses debate to determine strategy weights, whereas \myframework{} uses them to generate critiques for iterative refinement.

\paragraph{ES-VR~\cite{kim-etal-2025-dialogue}.}
A value-driven framework that improves emotional support by reinforcing seekers' positive values within the ongoing conversation. We adapt it to our single-turn setting by merging multi-turn exchanges into single post-response pairs.

\subsection{Backbone Models}
We evaluate our framework across a diverse set of backbone language models, including qwen2.5-14b-instruct, qwen2.5-72b-instruct~\cite{qwen2025qwen25technicalreport}, gemma-3-27b-it~\cite{gemma_2025}, gemma-4-31b-it~\cite{gemma4}, mistral-nemo-instruct-2407~\cite{mistralai2024nemo}, and gemini-2.5-flash-lite~\cite{gemini25flashlite2025} (hereafter Qwen2.5-14B, Qwen2.5-72B, Gemma3, Gemma4, Mistral, and Gemini, respectively).
These models exhibit variation in design and capability, enabling comprehensive evaluation of \myframework's performance across diverse model families and scales.

\subsection{Results}
\label{sec:results}

Across all three evaluation approaches, \myframework~outperforms baselines, demonstrating personalization alignment, balanced response quality, and strong preference from actual seekers.

% \subsubsection{RQ1) Superior Personalization Alignment} 
% To determine whether our framework accurately aligns its response with the seeker's inferred characteristics, we first examine the LLM-as-a-Judge results. 
% As shown in Table~\ref{tab:pas_scores}, \myframework~achieves the highest PAS across all backbones, peaking at 4.91 with \texttt{GEMMA}. 
% This result confirms that our iterative approach effectively bridges the gap between generic advice and seeker-specific characteristics. Furthermore, in SPC metric (Table~\ref{tab:pairwise_comparison}), \myframework~records win rates up to 92.44\%, significantly outperforming specialized baselines like MentalAgora and ES-VR.
% These results provide strong evidence that our multi-agent iterative approach successfully generates responses that are deeply attuned to individual seeker characteristics.

% \subsubsection{RQ1) Post History Enables Better Alignment}
% \paragraph{Ans 1) Post history enables better alignment}
% \paragraph{Ans 1) \myframework~aligns the seeker's characteristics.}
\paragraph{LLM-as-a-Judge.}
\label{sec:results_llm_as_a_judge}
% To determine whether our framework aligns its response with the seeker's inferred characteristics, we examine the LLM-as-a-Judge results.

% As shown in Table~\ref{tab:pas_scores}, \myframework~achieves the highest PAS across all backbones. Notably, Vanilla consistently outperforms MentalAgora and ES-VR despite using only standard prompting. This pattern can be attributed to post history utilization: both \myframework~and Vanilla leverage the seeker's post history, while MentalAgora and ES-VR rely solely on the target post. Access to post history provides richer context for inferring seeker characteristics, enabling more aligned emotional supports.

% Furthermore, the SPC results (Table~\ref{tab:pairwise_comparison}) show that 
% As shown in Table~\ref{tab:pairwise_comparison}, 
% \myframework~achieves win rates of 70-92\% against all baselines. The consistent advantage over Vanilla demonstrates that our structured approach effectively translates post history into personalized emotional support compared to standard prompting.

\input{latex/table/050_experiments_human_eval}
\input{latex/table/050_experiments_user_study}

As shown in Table~\ref{tab:pairwise_comparison}, \myframework~outperforms all baselines in pairwise comparison across all backbone and judge configurations.
% As shown in Table~\ref{tab:pairwise_comparison}, \myframework~consistently outperforms all baselines across backbone and judge configurations.
The advantage over Vanilla shows that \myframework~effectively translates post history into personalized emotional support. The improvements against ES-VR are particularly pronounced, with win rates exceeding 79\% across all settings. \myframework~also maintains an advantage over MentalAgora across all backbones and judges.
To verify the reliability of our automatic evaluation, we conduct inter-judge and judge--human agreement analyses (Appendix~\ref{apdx:llm_reliability}). Inter-judge directional agreement ranges from 0.82 to 0.91, exceeding the 0.81 human--human reference reported by \citet{10.5555/3666122.3668142}. 
Judge--human agreement for gpt-4o-mini, our analysis judge, reaches 0.64 exact and 0.74 directional, matching the 0.63 exact reference while falling slightly short on directional.

\paragraph{Expert Human Evaluation.}

% As shown in Table~\ref{tab:human_eval_5}, \myframework~achieves the highest scores in Empathy, Personalization, and Understanding, while also being ranked as the most preferred method. These results indicate that \myframework~generates responses demonstrating deeper emotional connection to seekers.
% Notably, MentalAgora achieves the highest Helpfulness score but the lowest Empathy score, indicating that its focus on actionable advice comes at the cost of emotional resonance.

As shown in Table~\ref{tab:human_eval_5}, in our expert human evaluation, \myframework~achieves the highest scores in Empathy, Personalization, and Understanding, while also being ranked as the most preferred method.
Notably, MentalAgora achieves the highest Helpfulness score but the lowest Empathy score, suggesting that its focus on actionable advice through multi-agent debate comes at the cost of emotional resonance.
In contrast, \myframework~achieves a better balance between empathy and actionable guidance.
These results indicate that, under expert assessment, \myframework~ generates responses that meet professional psychological standards without sacrificing empathetic quality.
We use Intraclass Correlation Coefficient (ICC) \cite{mcgraw1996forming} to assess inter-evaluator agreement on Likert-scale ratings, obtaining an overall ICC(2,k) of 0.645, indicating good agreement \cite{cicchetti1994guidelines}.

\input{latex/table/060_analysis}

% \subsubsection{RQ3) Seekers Most Prefer \myframework}
% \paragraph{Ans 3) Seekers most prefer \myframework}
\paragraph{User Study with Seekers.}
% To capture perspectives that third-party evaluation may miss, we conduct a user study where 20 participants with lived experience of mental health concerns evaluate the responses. 

As shown in Table~\ref{tab:user_study}, \myframework~achieves the highest scores across all six dimensions, the best average, and the top overall ranking. In contrast, MentalAgora ties with \myframework~for the best Perceived Helpfulness but scores the lowest on Empathy, indicating that competing methods trade off emotional attunement for informational utility. Crucially, these evaluations were conducted by the seekers themselves, which strengthens the validity of the results: from the perspective of those most directly affected, \myframework~is judged as the most empathetic, personalized, and trustworthy system, supporting its effectiveness over existing baselines in our user study.

Qualitative feedback reinforces these findings. When asked why they preferred a particular response, one participant explained: 
``\textit{This response feels like it understands why I said what I said and considers the context that leads me to express myself that way, which makes the response feel more emotionally supportive}.'' 
Importantly, this feedback was given for a response from \myframework, without the participant knowing which method produced it.
This highlights that seekers value responses that acknowledge not just what they said, but why they may have said it. This is precisely what \myframework~achieves by leveraging post history.

% Qualitative feedback from participants further reinforces these findings. In post-study interviews, one participant notes that ``\textit{\myframework~feels like it understands why I said what I said and considers the context that leads me to express myself that way, which makes the response feel more emotionally supportive}.'' This perception directly reflects the design of \myframework, which jointly considers the seeker's post history alongside the target post to infer underlying emotional states and personal circumstances, enabling responses that acknowledge not just what the seeker said, but why they may have said it.

%% file: latex/table/050_experiments_spc.tex
\begin{table*}[t]
    \centering
    \resizebox{\textwidth}{!}{
    \begin{tabular}{cl||rrr|rrr|rrr}
        \toprule
         & & \multicolumn{9}{c}{\textbf{Judge Model}} \\
        \cmidrule(lr){3-11}
        \multirow{2}{*}{\textbf{\makecell{Backbone\\Model}}} & \multicolumn{1}{c||}{\multirow{2}{*}{\textbf{Method}}} & \multicolumn{3}{c|}{\textbf{gpt-4o-mini}} & \multicolumn{3}{c|}{\textbf{deepseek-v4-flash}} & \multicolumn{3}{c}{\textbf{glm-4.5-air}} \\
        \cmidrule(lr){3-5} \cmidrule(lr){6-8} \cmidrule(lr){9-11}
        & \multicolumn{1}{c||}{} & \multicolumn{1}{c}{\textbf{Wins}} & \multicolumn{1}{c}{\textbf{Losses}} & \multicolumn{1}{c|}{\textbf{Ties}} & \multicolumn{1}{c}{\textbf{Wins}} & \multicolumn{1}{c}{\textbf{Losses}} & \multicolumn{1}{c|}{\textbf{Ties}} & \multicolumn{1}{c}{\textbf{Wins}} & \multicolumn{1}{c}{\textbf{Losses}} & \multicolumn{1}{c}{\textbf{Ties}} \\
        \midrule
        \multirow{3}{*}{\makecell{Qwen2.5\\14B}} 
        & vs. Vanilla & \valbc{75.36}{1.09} & \valc{24.64}{1.09} & \valc{0.00}{0.00} \ 
                & \valbc{55.58}{1.20} & \valc{44.36}{1.20} & \valc{0.06}{0.06} \
                & \valbc{73.48}{1.01} & \valc{26.50}{1.01} & \valc{0.02}{0.02} \\
        & vs. MentalAgora & \valbc{85.32}{0.87} & \valc{14.68}{0.87} & \valc{0.00}{0.00} \ 
                & \valbc{79.76}{1.02} & \valc{20.22}{1.02} & \valc{0.02}{0.02} \
                & \valbc{75.15}{1.06} & \valc{24.83}{1.06} & \valc{0.02}{0.02} \\
        & vs. ES-VR & \valbc{79.01}{1.04} & \valc{20.99}{1.04} & \valc{0.00}{0.00} \ 
                & \valbc{84.54}{0.90} & \valc{15.43}{0.90} & \valc{0.02}{0.02} \
                & \valbc{84.21}{0.86} & \valc{15.79}{0.86} & \valc{0.00}{0.00} \\
        \midrule
        \multirow{3}{*}{\makecell{Qwen2.5\\72B}} 
        & vs. Vanilla & \valbc{70.13}{1.09} & \valc{29.87}{1.09} & \valc{0.00}{0.00} \ 
                 & \valbc{61.28}{1.21} & \valc{38.68}{1.21} & \valc{0.04}{0.03} \
                 & \valbc{78.22}{0.91} & \valc{21.78}{0.91} & \valc{0.00}{0.00} \\
        & vs. MentalAgora & \valbc{89.87}{0.63} & \valc{10.13}{0.63} & \valc{0.00}{0.00} \   
                & \valbc{86.93}{0.80} & \valc{12.99}{0.80} & \valc{0.08}{0.05} \
                & \valbc{87.93}{0.70} & \valc{12.05}{0.70} & \valc{0.02}{0.02} \\
        & vs. ES-VR & \valbc{97.81}{0.29} & \valc{2.19}{0.29} & \valc{0.00}{0.00} \ 
                & \valbc{84.57}{0.86} & \valc{15.43}{0.86} & \valc{0.00}{0.00} \
                & \valbc{90.60}{0.63} & \valc{9.40}{0.63} & \valc{0.00}{0.00} \\
        \midrule
        \multirow{3}{*}{Gemma3} 
        & vs. Vanilla & \valbc{70.59}{1.14} & \valc{29.41}{1.14} & \valc{0.00}{0.00} \ 
                & \valbc{73.71}{0.88} & \valc{26.11}{0.89} & \valc{0.19}{0.07} \
                & \valbc{76.75}{0.87} & \valc{23.25}{0.87} & \valc{0.00}{0.00} \\
        & vs. MentalAgora & \valbc{72.45}{1.14} & \valc{27.55}{1.14} & \valc{0.00}{0.00} \ 
                & \valbc{60.30}{1.07} & \valc{39.58}{1.07} & \valc{0.13}{0.05} \
                & \valbc{59.63}{1.06} & \valc{40.37}{1.06} & \valc{0.00}{0.00} \\
        & vs. ES-VR & \valbc{92.44}{0.57} & \valc{7.56}{0.57} & \valc{0.00}{0.00}  \ 
                & \valbc{84.25}{0.74} & \valc{15.75}{0.74} & \valc{0.00}{0.00} \
                & \valbc{79.62}{0.79} & \valc{20.36}{0.79} & \valc{0.02}{0.02} \\
        \midrule
        \multirow{3}{*}{Gemma4} 
        & vs. Vanilla & \valbc{62.43}{1.22} & \valc{37.57}{1.22} & \valc{0.00}{0.00} \ 
                & \valbc{82.81}{0.77} & \valc{16.79}{0.77} & \valc{0.40}{0.10} \
                & \valbc{86.45}{0.69} & \valc{13.53}{0.69} & \valc{0.02}{0.02} \\
        & vs. MentalAgora & \valbc{52.92}{1.03} & \valc{47.08}{1.03} & \valc{0.00}{0.00} \ 
                & \valbc{56.83}{1.05} & \valc{42.77}{1.04} & \valc{0.40}{0.11} \
                & \valbc{57.21}{1.08} & \valc{42.71}{1.08} & \valc{0.08}{0.04} \\
        & vs. ES-VR & \valbc{86.49}{0.90} & \valc{13.51}{0.90} & \valc{0.00}{0.00} \ 
                & \valbc{96.12}{0.42} & \valc{3.86}{0.41} & \valc{0.02}{0.02} \
                & \valbc{95.34}{0.42} & \valc{4.64}{0.42} & \valc{0.02}{0.02} \\
        \midrule
        \multirow{3}{*}{Mistral} 
        & vs. Vanilla & \valbc{70.30}{1.25} & \valc{29.70}{1.25} & \valc{0.00}{0.00} \ 
                & \valbc{52.55}{1.27} & \valc{47.45}{1.27} & \valc{0.00}{0.00} \
                & \valbc{65.73}{1.17} & \valc{34.27}{1.17} & \valc{0.00}{0.00} \\
        & vs. MentalAgora & \valbc{90.41}{0.77} & \valc{9.59}{0.77} & \valc{0.00}{0.00} \ 
                & \valbc{75.29}{1.14} & \valc{24.69}{1.14} & \valc{0.02}{0.02} \
                & \valbc{79.41}{1.00} & \valc{20.57}{1.00} & \valc{0.02}{0.02} \\
        & vs. ES-VR & \valbc{89.79}{0.63} & \valc{10.21}{0.63} & \valc{0.00}{0.00} \ 
                & \valbc{94.90}{0.50} & \valc{5.10}{0.50} & \valc{0.00}{0.00} \
                & \valbc{91.31}{0.54} & \valc{8.69}{0.54} & \valc{0.00}{0.00} \\
        \midrule
        \multirow{3}{*}{Gemini} 
        & vs. Vanilla & \valbc{98.29}{0.34} & \valc{1.71}{0.34} & \valc{0.00}{0.00} \ 
                & \valbc{97.77}{0.33} & \valc{2.19}{0.33} & \valc{0.04}{0.03} \
                & \valbc{97.93}{0.30} & \valc{2.07}{0.30} & \valc{0.00}{0.00} \\
        & vs. MentalAgora & \valbc{51.11}{1.22} & \valc{48.89}{1.22} & \valc{0.00}{0.00} \ 
                & \valbc{74.06}{0.91} & \valc{25.65}{0.91} & \valc{0.29}{0.08} \
                & \valbc{66.42}{1.08} & \valc{33.56}{1.08} & \valc{0.02}{0.02} \\
        & vs. ES-VR & \multicolumn{1}{c}{--} & \multicolumn{1}{c}{--} & \multicolumn{1}{c}{--} & \multicolumn{1}{c}{--} & \multicolumn{1}{c}{--} & \multicolumn{1}{c}{--} & \multicolumn{1}{c}{--} & \multicolumn{1}{c}{--} & \multicolumn{1}{c}{--} \\
        \bottomrule
    \end{tabular}
    }
    \caption{Pairwise comparison results using LLM-as-a-Judge. Values represent the percentage of wins, losses, and ties for \myframework~against each baseline. ES-VR is not applicable to Gemini as it requires fine-tuning.}
    \label{tab:pairwise_comparison}
\end{table*}

%% file: latex/table/050_experiments_human_eval.tex
\begin{table*}[t]
    \centering
    % \resizebox{0.8\textwidth}{!}
    {
    \begin{tabular}{l||rrrrr|r}
    \toprule
    \multicolumn{1}{c||}{\textbf{Method}} & \multicolumn{1}{c}{\textbf{Emp.$\uparrow$}} & \multicolumn{1}{c}{\textbf{Hel.$\uparrow$}} & \multicolumn{1}{c}{\textbf{Per.$\uparrow$}} & \multicolumn{1}{c}{\textbf{Und.$\uparrow$}} & \multicolumn{1}{c|}{\textbf{Avg.$\uparrow$}} & \multicolumn{1}{c}{\textbf{Rank$\downarrow$}} \\
    \midrule
    Vanilla & \valc{3.80}{0.03} & \valc{2.97}{0.03} & \valc{3.01}{0.02} & \valc{3.46}{0.03} & \valc{3.31}{0.01} & \valc{3.26}{0.04} \\
    MentalAgora & \valc{3.37}{0.03} & \valbc{4.03}{0.03} & \valuc{3.36}{0.03} & \valc{3.55}{0.03} & \valuc{3.58}{0.01} & \valuc{2.14}{0.05} \\
    ES-VR & \valuc{3.88}{0.03} & \valc{3.42}{0.03} & \valc{3.28}{0.03} & \valuc{3.72}{0.03} & \valc{3.57}{0.02} & \valc{2.48}{0.05} \\
    \myframework~& \valbc{4.20}{0.03} & \valuc{3.73}{0.03} & \valbc{3.48}{0.03} & \valbc{3.82}{0.03} & \valbc{3.81}{0.01} & \valbc{2.12}{0.04} \\
    \bottomrule
    \end{tabular}
    }
    \caption{Expert human evaluation results using responses generated by Qwen2.5-14B. Higher scores indicate greater response quality across four dimensions: Empathy (Emp.), Helpfulness (Hel.), Personalization (Per.), and Understanding (Und.). 
    Average (Avg.) represents the average score across all four dimensions. Lower ranks indicate a stronger overall preference.
    % \myframework~achieves the highest average score and the best overall rank, demonstrating well-balanced performance across all evaluation dimensions.
    % The inter-annotator agreement, measured using Fleiss' kappa \cite{fleiss1971measuring}, yielded an overall value of x.xx, indicating slight agreement according to the interpretation guidelines proposed by \citet{landis1977measurement}.
    }
    \label{tab:human_eval_5}
\end{table*}

%% file: latex/table/050_experiments_user_study.tex
\begin{table*}[t]
\centering
\resizebox{\textwidth}{!}{%
\begin{tabular}{l||rrrrrrr|r}
\toprule
\multicolumn{1}{c||}{\textbf{Method}} & \multicolumn{1}{c}{\textbf{Emp.$\uparrow$}} & \multicolumn{1}{c}{\textbf{PH$\uparrow$}} & \multicolumn{1}{c}{\textbf{Per.$\uparrow$}} & \multicolumn{1}{c}{\textbf{Rel.$\uparrow$}} & \multicolumn{1}{c}{\textbf{TS$\uparrow$}} & \multicolumn{1}{c}{\textbf{WtU$\uparrow$}} & \multicolumn{1}{c|}{\textbf{Avg.$\uparrow$}} & \multicolumn{1}{c}{\textbf{Rank$\downarrow$}} \\
\midrule
Vanilla & \valc{3.76}{0.15} & \valc{2.82}{0.14} & \valc{3.12}{0.17} & \valc{3.42}{0.13} & \valc{3.28}{0.15} & \valc{2.92}{0.19} & \valc{3.22}{0.06} & \valc{2.96}{0.13} \\
MentalAgora & \valc{3.24}{0.16} & \valbc{3.56}{0.16} & \valc{3.40}{0.17} & \valuc{3.68}{0.13} & \valc{3.48}{0.15} & \valc{3.16}{0.18} & \valc{3.42}{0.06} & \valc{2.58}{0.17} \\
ES-VR & \valuc{3.80}{0.15} & \valc{3.30}{0.14} & \valuc{3.46}{0.14} & \valc{3.56}{0.14} & \valuc{3.68}{0.12} & \valuc{3.22}{0.15} & \valuc{3.50}{0.06} & \valuc{2.42}{0.15} \\
\myframework~ & \valbc{3.98}{0.12} & \valbc{3.56}{0.16} & \valbc{3.70}{0.18} & \valbc{3.76}{0.14} & \valbc{3.80}{0.15} & \valbc{3.52}{0.18} & \valbc{3.72}{0.06} & \valbc{2.04}{0.15} \\
\bottomrule
\end{tabular}
}
\caption{
User study results using responses generated by Qwen2.5-14B. Higher scores indicate greater user satisfaction across six dimensions: Empathy (Emp.), Perceived Helpfulness (PH), Personalization (Per.), Relevance (Rel.), Trustworthiness \& Safety (TS), and Willingness to Use (WtU). Average (Avg.) represents the average score across all dimensions. Lower ranks indicate better overall preference.
% \myframework~achieves the highest scores across all six dimensions and is ranked as the most preferred method, indicating that seekers perceive responses generated by \myframework~as more supportive and satisfying.
}
\label{tab:user_study}
\end{table*}

%% file: latex/table/060_analysis.tex
\begin{table*}[t]
\centering
\resizebox{\textwidth}{!}{
\begin{tabular}{ccl||rrr}
\toprule
\multicolumn{2}{c}{\textbf{Experiment}} & \multicolumn{1}{c||}{\textbf{Method}} & \multicolumn{1}{c}{\textbf{Wins}} & \multicolumn{1}{c}{\textbf{Losses}} & \multicolumn{1}{c}{\textbf{Ties}} \\
\midrule
\multicolumn{2}{c}{\multirow{2}{*}{Stage-Level Ablation}}
& vs. w/o Ref. & \valbc{76.80}{0.85} & \valc{23.16}{0.85} & \valc{0.04}{0.03} \\
\multicolumn{2}{c}{}
& vs. w/o CF \& Ref. & \valbc{83.40}{0.93} & \valc{16.60}{0.93} & \valc{0.00}{0.00} \\
\midrule
\multirow{7}{*}{\makecell{Case Formulation\\Analysis}}
& History Swap & vs. Swapped & \valbc{78.05}{0.64} & \valc{21.95}{0.64} & \valc{0.00}{0.00} \\
\cmidrule{2-6}
& \multirow{2}{*}{Persona Comparison} & vs. Persona + Ref.& \valbc{66.92}{1.18} & \valc{33.08}{1.18} & \valc{0.00}{0.00} \\
& & vs. Persona w/o Ref. & \valbc{70.53}{1.17} & \valc{29.47}{1.17} & \valc{0.00}{0.00} \\
\cmidrule{2-6}
& \multirow{4}{*}{Dimension Ablation} & vs. w/o Seeker Profile & \valbc{52.90}{1.21} & \valc{47.10}{1.21} & \valc{0.00}{0.00} \\
& & vs. w/o Underlying Concern& \valbc{56.06}{1.21} & \valc{43.94}{1.21} & \valc{0.00}{0.00} \\
& & vs. w/o Historical Context & \valbc{53.28}{1.21} & \valc{46.72}{1.21} & \valc{0.00}{0.00} \\
& & vs. w/o Response Blueprint & \valbc{54.89}{1.23} & \valc{45.11}{1.23} & \valc{0.00}{0.00} \\
\midrule
\multicolumn{2}{c}{\multirow{3}{*}{Guidance Synthesis Ablation}}
& vs. Single Agent & \valbc{56.68}{1.14} & \valc{43.19}{1.14} & \valc{0.13}{0.06} \\
\multicolumn{2}{c}{}
& vs. Random Selection & \valbc{53.95}{1.11} & \valc{46.01}{1.11} & \valc{0.04}{0.03} \\
\multicolumn{2}{c}{}
& vs. No Limit & \valbc{62.51}{1.14} & \valc{37.49}{1.14} & \valc{0.00}{0.00} \\
\bottomrule
\end{tabular}
}
\caption{Ablation and analysis studies on \myframework~components. The Stage-Level Ablation removes the case formulation and/or refinement stages; the Case Formulation Analysis includes history swap, persona-based prompting comparison, and per-dimension removal of the four case formulation components; the Guidance Synthesis Ablation compares against Single Agent, Random Selection, and No Limit variants.}
\label{tab:ablation}
\end{table*}

%% file: latex/060_analysis.tex
% \input{ltex/table/061_analysis_stage_ablation}
% \input{latex/table/061_analysis_dimension_ablation}
% \input{latex/table/061_analysis_history_swap}
% \input{latex/table/061_analysis_persona_comparison}

% \input{latex/table/061_analysis_case_formulation}

% \input{latex/table/061_analysis_critique_ablation}

% \begin{figure}[t!]
%     \centering
%         \includegraphics[width=0.8\columnwidth]{latex/image/ours_vs_vanilla_qwen25_14B.pdf}
%     \caption{
%     % Seeker-Aware Pairwise Comparison (SPC) win rates of the final response (Round 6) against responses from each intermediate round.
%     Evolution of Seeker-Aware Pairwise Comparison win rates across refinement rounds for \texttt{Qwen}.
%     \myframework~(blue) shows consistent improvement over Vanilla (red) as rounds progress, with performance plateauing around Round-4. 
%     % This pattern supports adopting Round-4 responses as the final output to balance personalization quality with computational efficiency.
%     }
%     \label{fig:vs_vanilla}
% \end{figure}

% \begin{figure*}[ht]
%     \centering
%     \includegraphics[width=\linewidth]{latex/image/ours_vs_vanilla_all_models.png}
%     \caption{\myframework~vs. Vanilla}
%     \label{fig:ours_vs_vanilla}
% \end{figure*}

In this section, we analyze \myframework~from multiple perspectives to validate the contribution of each component.
All experiments in this section use Qwen2.5-14B as the backbone model, and gpt-4o-mini is used as the judge model.

\subsection{Stage-Level Ablation}
\label{sec:stage_ablation}
To evaluate the contribution of each stage in \myframework, we compare two ablation variants against the full \myframework: (1) w/o Ref., which removes the collaborative refinement stage, and (2) w/o CF \& Ref., which removes both the case formulation and refinement stages.

As shown in Table~\ref{tab:ablation}, \myframework~outperforms both ablation variants. This demonstrates that both case formulation and collaborative refinement stages are essential for effective personalization.

\subsection{Case Formulation Analysis}
\label{sec:case_formulation_analysis}

We conduct three experiments to validate the effectiveness of case formulation from different perspectives: (1) History Swap to verify that seeker-specific information is leveraged, (2) Persona Comparison to demonstrate the advantage of structured case formulation over persona-based prompting, and (3) Dimension Ablation to assess the contribution of each dimension in case formulation.

\paragraph{History Swap.}
To verify that \myframework~leverages seeker-specific information rather than mere contextual augmentation, we replace each seeker's history with the three most dissimilar ones based on case formulation distance. Responses with the original history achieve a 78.05\% win rate, confirming that case formulation captures information that directly improves response quality.

\paragraph{Persona Comparison.}
We examine whether structured case formulation provides advantages over simple persona inference. We compare against persona-based prompting, where the model infers the seeker's persona from the target post and post history instead of constructing a case formulation. \myframework~achieves win rates of 66.92--70.53\%, demonstrating that structured case formulation captures deeper factors, such as underlying concern, longitudinal patterns, and response guidelines, that simple persona inference cannot capture.

\paragraph{Dimension Ablation.}
We examine whether each dimension of our case formulation contributes to generating personalized responses. 
As shown in Table~\ref{tab:ablation}, removing any single dimension degrades performance, with win rates ranging from 52.90\% to 56.06\%. The comparable magnitude of drops across dimensions indicates that all four components contribute similarly to response quality, and that the effectiveness of case formulation stems from their joint use rather than from any single dominant dimension.

\input{latex/table/060_analysis_case_study_sample}

\subsection{Guidance Synthesis Ablation}
\label{sec:critique_ablation}

\myframework~employs three counselor agents and synthesizes their critiques into focused guidance. To investigate the contribution of this mechanism, we compare against three variants: (1) Single Agent, which uses a single agent instead of three counselors; (2) Random Selection, which randomly picks 2 critiques; and (3) No Limit, which applies all critiques without top-2 filtering.

As shown in Table~\ref{tab:ablation}, \myframework~outperforms all variants. The largest gain is against the variant without critique limits, confirming that controlled feedback volume prevents quality degradation. Multi-agent diversity also contributes, and prioritized selection outperforms random selection.

\subsection{Case Study}
\label{sec:case_study}

In Table~\ref{tab:case_study}, we provide an example to compare the responses of \myframework~ with the baselines. 
Vanilla, MentalAgora, and ES-VR all provide relatively generic emotional support, with responses that do not fully leverage the seeker's personal context. While Vanilla acknowledges the seeker's situation, it fails to translate the post history it receives into personalized support, offering generic advice for anyone experiencing work-related fatigue. MentalAgora and ES-VR, which do not have access to the seeker's post history, are inherently limited to the target post. MentalAgora provides comprehensive but clinically structured information that lacks emotional warmth. ES-VR shows more empathy but still misses key personal context.

Whereas \myframework, through its case formulation stage, identifies critical information from the seeker's post history: their \textit{fear of medication dependence due to witnessed substance-related loss} and their \textit{cautious approach to medication with preference for exploring lifestyle adjustments}. This enables \myframework~to generate a response that explicitly validates their ``thoughtful and cautious approach'' to medication, emphasizes \textbf{non-medication alternatives} aligned with their preferences, and provides \textbf{concrete, personalized suggestions} referencing their specific work environment and dietary patterns. Additional case studies demonstrating \myframework's personalization across different scenarios are provided in Appendix~\ref{apdx:case_study}.

%% file: latex/table/060_analysis_case_study_sample.tex
\begin{table*}[t!]
\centering
\small
\begin{tabular}{l|p{0.84\textwidth}}
\toprule
\multicolumn{2}{c}{\textbf{Target Post}} \\
\midrule
Seeker & I've been experiencing persistent fatigue that I can't explain. I work a physically demanding service job in a hot environment, 9 AM to 4 PM. I often skip lunch because my medication suppresses hunger. I'm cautiously adjusting my dosage—taking less than prescribed while gradually increasing. Why do I crash so hard after work? \\
\midrule
\multicolumn{2}{c}{\textbf{Post History}} \\
\midrule
Seeker & I have a deeply-rooted fear of taking medications because I've seen too many people close to me struggle with substance issues. \\
Seeker & Is it normal to be completely exhausted by late afternoon? I come home and can barely do anything. \\
Seeker & How do you deal with after-work exhaustion without relying on substances? \\
% ...... & \\
\midrule
\multicolumn{2}{c}{\textbf{Case Formulation (\myframework)}} \\
\midrule
\multicolumn{2}{p{14.2cm}}{
\textbf{Seeker Profile}: Cautious about medication, values practical and community-based solutions \newline
\textbf{Underlying Concern}: Fear of medication dependence rooted in personal loss \newline
\textbf{Historical Context}: Recurring energy struggles with consistent preference for non-pharmaceutical approaches \newline
\textbf{Response Blueprint}: Warm + Balanced; validate cautious approach; prioritize non-medication suggestions
} \\
\midrule
\multicolumn{2}{c}{\textbf{Response}} \\
\midrule
Vanilla & It sounds like you're struggling with energy levels. Try staying hydrated and packing small snacks. Maybe some light exercise after work could help too. 
% \textit{(Generic Advice)} 
\\
\midrule
MentalAgora & Here are some factors to consider: (1) Physical demands and environment—stay hydrated. (2) Diet—eat balanced meals. (3) Sleep hygiene—maintain consistent schedule. (4) Medication—consult your doctor. 
% \textit{(Comprehensive but Impersonal)} 
\\
\midrule
ES-VR & This sounds really challenging. Your job and medication adjustments are both taking a toll. Consider keeping a log of your daily activities and discussing with your doctor. 
% \textit{(Empathetic but Misses Personal Context)} 
\\
\midrule
\myframework & I really appreciate your resilience and the \textbf{thoughtful, cautious approach} you're taking with your medication. Given your situation, here are some \textbf{non-medication strategies}: try carrying small snacks like almonds despite appetite suppression, take brief cooling breaks during shifts, and \textbf{connect with others in the community} who understand these challenges. 
% \textit{(Personalized with Strong Empathy)} 
\\
\bottomrule
\end{tabular}
\caption{Case study comparing MindTailor with baselines. 
\myframework~leverages post history to identify the seeker's medication hesitancy and preference for exploring lifestyle adjustments, generating personalized support that validates their approach and aligns with their demonstrated coping style.
Posts and responses have been paraphrased to prevent re-identification while preserving original meaning and intent.
% All content has been rephrased or condensed to protect user privacy.
}
\label{tab:case_study}
\end{table*}

%% file: latex/100_conclusion.tex
% We proposed \myframework, a multi-agent framework for generating personalized emotional support responses that leverages seekers' post histories for context analysis and iteratively refines responses through collaboration among counselor agents grounded in distinct therapeutic strategies. We also introduced \mydataset, a dataset pairing Reddit posts with seekers' posting histories. Our comprehensive evaluations, including LLM-based metrics, expert human evaluation, and user studies, demonstrate that \myframework~outperforms baselines in empathy, helpfulness, and personalization.

% We propose \myframework, a collaborative critique-based framework that generates personalized emotional support by leveraging case formulation and iteratively refining responses through collaborative critique and revision among counselor agents grounded in distinct therapeutic strategies. Evaluations using LLM-as-a-Judge, expert human evaluation, and a user study demonstrate that \myframework~outperforms baselines in empathy, personalization, and understanding, with seekers rating it as the most preferred and trustworthy method.

We propose \myframework, a framework that generates personalized emotional support by combining structured case formulation with iterative collaborative critique among counselor agents grounded in distinct counseling strategies.
\myframework~outperforms baselines in pairwise comparison, achieves the highest scores in empathy, personalization, and understanding in expert human evaluation, and is most preferred by seekers themselves.

% Case studies illustrate how \myframework~leverages post history to engage with each seeker's formative experiences and underlying concerns. Extensive analyses further support the contribution of each component.

% We propose \myframework, a framework that generates personalized emotional support through case formulation and iterative collaborative refinement among counselor agents grounded in distinct counseling strategies. Evaluations using LLM-as-a-Judge, expert human evaluation, and a user study demonstrate that \myframework~outperforms baselines in empathy, personalization, and understanding, with seekers rating it as the most preferred and trustworthy method.

% For future work, we plan to incorporate a more diverse set of counselor agents representing additional therapeutic strategies, enabling richer multi-perspective feedback during the refinement process. We also aim to improve computational efficiency through early stopping mechanisms and more streamlined refinement strategies that reduce overhead while maintaining response quality.

% Future Work? 정 자리가 부족하다면 이를 Limitations에 언급하면서 어떻게 future work으로 할 것인지에 대해 얘기하는게 좋겠습니다.

%% file: latex/appendix/101_counseling_strategy.tex
% \begin{figure*}[t!]
%     \centering
%     \begin{subfigure}[b]{0.32\textwidth}
%         \includegraphics[width=\textwidth]{latex/image/ours_vs_vanilla_qwen25_14B.pdf}
%         \caption{\texttt{Qwen}}
%         \label{fig:vs_vanilla_qwen}
%     \end{subfigure}
%     \hfill
%     \begin{subfigure}[b]{0.32\textwidth}
%         \includegraphics[width=\textwidth]{latex/image/ours_vs_vanilla_gemma3_27B.pdf}
%         \caption{\texttt{Gemma}}
%         \label{fig:vs_vanilla_gemma}
%     \end{subfigure}
%     \hfill
%     \begin{subfigure}[b]{0.32\textwidth}
%         \includegraphics[width=\textwidth]{latex/image/ours_vs_vanilla_mistral_nemo.pdf}
%         \caption{\texttt{Mistral}}
%         \label{fig:vs_vanilla_mistral_nemo}
%     \end{subfigure}
%     \caption{
%     % Seeker-Aware Pairwise Comparison (SPC) win rates of the final response (Round 6) against responses from each intermediate round.
%     Evolution of Seeker-Aware Pairwise Comparison win rates across refinement rounds for three backbone models. \myframework~(blue) shows consistent improvement over Vanilla (red) as rounds progress, with performance plateauing around Round-4. This pattern supports adopting Round-4 responses as the final output to balance personalization quality with computational efficiency.
%     }
%     \label{fig:vs_vanilla_full}
% \end{figure*}

\myframework~employs three distinct counseling strategies, each grounded in therapeutic practice:

\paragraph{Reframing} is a cognitive technique that helps individuals view their situation from a different, often more constructive perspective \cite{Beck1989}. By altering the conceptual or emotional framing of a problem without changing the facts themselves, reframing enables individuals to find new meaning or identify previously unrecognized opportunities within their circumstances. This strategy is widely used in cognitive-behavioral therapy to challenge negative thought patterns.

\paragraph{Unconditional Positive Regard} is a concept originally introduced by \citet{standar_regard} and later developed as a cornerstone of person-centered therapy, refers to the therapist's complete acceptance of the client without judgment or evaluation. This approach creates a safe, non-threatening environment where individuals feel valued regardless of their thoughts, feelings, or behaviors, thereby fostering self-exploration and personal growth.

\paragraph{Solution-Focused Counseling} is a goal-oriented therapeutic approach that emphasizes identifying solutions rather than analyzing problems \cite{Bannink2007}. Instead of extensively exploring the origins of difficulties, this strategy directs attention toward the client's strengths, resources, and envisioned future, encouraging them to construct concrete steps toward their desired outcomes.

The counselor agents introduced in \S\ref{sec:methodology} are specialized agents grounded in the aforementioned counseling strategies.

%% file: latex/appendix/105_dataset_construction.tex
The following describes the detailed construction process of \mydataset.

\subsection{Collecting Target Posts}

We first identified ``target posts'', submissions where a seeker explicitly expresses distress or a need for emotional support, by leveraging publicly available Reddit data collected via the Pushshift API \cite{baumgartner2020pushshift} prior to its access restrictions in 2023. Specifically, we retrieved candidate posts from the subreddit collection originally employed to construct the Reddit Mental Health Dataset \cite{low2020natural}. This collection encompasses both condition-specific groups (e.g., r/depression, r/anxiety, r/ptsd) and broader support forums (e.g., r/mentalhealth, r/lonely).

To ensure that the target posts are semantically rich and suitable for emotional support generation, we applied a multi-stage filtering protocol.
First, we retained only posts with lengths between 1K and 2K characters to guarantee adequate informational content for frameworks to identify the seeker's immediate problem. 
Second, we excluded posts created prior to January 1, 2020, to ensure alignment with contemporary mental health concerns. 
Third, to maximize seeker diversity, we removed duplicate submissions from the same author. 
Finally, we employed gpt-4o-mini \cite{openai2024gpt4omini} to conduct a semantic content evaluation, filtering out posts that do not necessitate emotional support (e.g., purely informational posts or spam), thereby ensuring that all retained target posts genuinely warrant a supporter's intervention.

\subsection{Constructing Post History}

For each target post, we aggregated the corresponding seeker's post history. A seeker's post history was defined as the collection of all submissions authored within a one-year retrospective window preceding the target post. This temporal constraint was imposed to ensure contextual relevance, as prior research suggests that individuals' mental health states and expressed concerns evolve over time \cite{macavaney-etal-2018-rsdd}, and excessively outdated posts may no longer reflect the seeker's current circumstances. This longitudinal historical data is crucial for enabling a more nuanced understanding of the seeker's needs. To ensure the availability of sufficient historical context, we retained only those instances where the post history contained a minimum of two prior posts. This curation process yielded a final dataset comprising 798 pairs of target posts and their associated histories, serving as the basis for the evaluation of \myframework.

%% file: latex/table/110_dataset_statistics.tex
\begin{table}[t!]
\centering
    \resizebox{\columnwidth}{!}{
    \begin{tabular}{clr}
    \toprule
    \textbf{Category} & \multicolumn{1}{c}{\textbf{Metric}} & \multicolumn{1}{c}{\textbf{Value}} \\
    \midrule
    \multirow{1}{*}{Dataset Size} & Total instances & 798 \\
    \midrule
    \multirow{5}{*}{Post History} 
        & Mean & 20.33 \\
        & Median & 8.0 \\
        & Std. Dev. & 41.71 \\
        & Min & 2 \\
        & Max & 550 \\
    \midrule
    \multirow{5}{*}{\shortstack[c]{Filtered\\Post History}} 
        & Mean & 4.28 \\
        & Median & 5.0 \\
        & Std. Dev. & 1.14 \\
        & Min & 2 \\
        & Max & 5 \\
    \midrule
    \multirow{5}{*}{\shortstack[c]{Target Post Length\\(chars)}} 
        & Mean & 1,363.9 \\
        & Median & 1,308.0 \\
        & Std. Dev. & 272.1 \\
        & Min & 1,000 \\
        & Max & 2,000 \\
    \midrule
    \multirow{3}{*}{\shortstack[c]{Temporal\\Distribution}}
        & 2020 & 242 (30.3\%) \\
        & 2021 & 263 (33.0\%) \\
        & 2022 & 293 (36.7\%) \\
    \bottomrule
    \end{tabular}
    }
    \caption{Summary statistics of \mydataset, including post history size, target post length, and temporal distribution. Post history statistics are reported both before and after the filtering process described in \S\ref{sec:methodology} and \S\ref{sec:dataset}.}
    \label{tab:dataset_overview}
\end{table}

\begin{table}[t!]
\centering
    \resizebox{\columnwidth}{!}{
    \begin{tabular}{lrr}
    \toprule
    \multicolumn{1}{c}{\textbf{Subreddit}} & \multicolumn{1}{c}{\textbf{Count}} & \multicolumn{1}{c}{\textbf{Percentage}} \\
    \midrule
    r/HealthAnxiety & 65 & 8.1\% \\
    r/COVID19\_support & 64 & 8.0\% \\
    r/mentalhealth & 59 & 7.4\% \\
    r/ptsd & 57 & 7.1\% \\
    r/autism & 57 & 7.1\% \\
    r/EDAnonymous & 56 & 7.0\% \\
    r/BPD & 54 & 6.8\% \\
    r/socialanxiety & 53 & 6.6\% \\
    r/Anxiety & 47 & 5.9\% \\
    r/schizophrenia & 45 & 5.6\% \\
    r/lonely & 45 & 5.6\% \\
    r/BipolarReddit & 42 & 5.3\% \\
    r/depression & 39 & 4.9\% \\
    r/addiction & 38 & 4.8\% \\
    r/alcoholism & 32 & 4.0\% \\
    r/SuicideWatch & 29 & 3.6\% \\
    r/ADHD & 16 & 2.0\% \\
    \midrule
    \textbf{Total} & \textbf{798} & \textbf{100\%} \\
    \bottomrule
    \end{tabular}
    }
    \caption{Distribution of target posts across 17 mental health-related subreddits in \mydataset. The dataset covers both condition-specific communities (e.g., r/ptsd, r/BPD) and general support forums (e.g., r/mentalhealth, r/lonely).}
    \label{tab:subreddit_distribution}
\end{table}

%% file: latex/appendix/110_dataset_statistics.tex
This appendix provides detailed statistics of \mydataset~to facilitate reproducibility and enable readers to better interpret the experimental results presented in this paper. All posts in \mydataset~are written in English, collected from Reddit, an anonymous social media platform. Due to the anonymized nature of Reddit, demographic information about the authors (e.g., age, gender, location) is not available and cannot be verified. The dataset comprises 798 instances, each consisting of a target post paired with the corresponding seeker's post history. Table~\ref{tab:dataset_overview} summarizes the overall dataset characteristics, while Table~\ref{tab:subreddit_distribution} presents the distribution of target posts across different subreddits.

% As shown in \autoref{tab:dataset_overview}, the original post history contains an average of 20.33 posts per seeker, with substantial variance (std.\ dev.\ = 41.71) reflecting the diverse activity levels among Reddit users. After applying the filtering criteria described in \autoref{sec:dataset}, the filtered post history used as input to \myframework~contains an average of 4.28 posts per instance. Target posts range from 1,000 to 2,000 characters in length, ensuring sufficient context for identifying seekers' immediate concerns.

% \autoref{tab:subreddit_distribution} illustrates that \mydataset~covers 17 mental health-related subreddits, spanning condition-specific communities (e.g., r/ptsd, r/BPD, r/schizophrenia) and broader support forums (e.g., r/mentalhealth, r/lonely). This diversity ensures that the dataset captures a wide range of mental health concerns and communication styles. The temporal distribution indicates a relatively balanced collection across 2020--2022, with slightly more posts from recent years.

%% file: latex/appendix/115_dataset_access.tex
While \mydataset~is constructed from publicly accessible posts, the content is sensitive and may enable re-identification of individual seekers. Unrestricted release is therefore not appropriate. Following the data release protocol established by the Reddit Self-reported Depression Diagnosis (RSDD) dataset of \citet{yates-etal-2017-depression}, we adopt a restricted-access model based on a signed data usage agreement to protect user privacy.

Researchers who wish to obtain \mydataset~are required to contact the authors directly via email to request the data usage agreement. Upon verification of the requester's institutional affiliation and intended research purpose, the authors provide the agreement document. Access instructions for \mydataset~are issued only after the signed agreement has been returned to the authors.

The data usage agreement, modeled after \citet{yates-etal-2017-depression}, stipulates that researchers agree to (1) make no attempt to contact any seeker in \mydataset, (2) make no attempt to deanonymize or learn the identity of any seeker in \mydataset, (3) make no attempt to link seekers in \mydataset~with any external information (e.g., an account on another website), and (4) not share any portion of the data, including example posts or excerpts from posts, with any other party.

These conditions are intended to safeguard the privacy of seekers represented in \mydataset~while enabling legitimate research use.

%% file: latex/appendix/190_framework_prompts.tex
This section provides the prompts used at each stage of \myframework.
The framework consists of three sequential stages: Seeker Understanding, Draft Generation, and Collaborative Refinement.

\subsection{Stage 1: Seeker Understanding}

In this stage, we construct a structured case formulation from the seeker's target post and their retrieved post history. The case formulation captures four dimensions: \textit{Seeker Profile}, \textit{Underlying Concern}, \textit{Historical Context}, and \textit{Response Blueprint}. The prompt used for case formulation construction is shown in Figure~\ref{fig:case_formulation_prompt_1}--\ref{fig:case_formulation_prompt_4}.

The case formulation construction prompt is designed around a single criterion: whether the resulting case formulation would enable a downstream response generator to write a reply that feels specifically tailored to this seeker, rather than generically applicable to anyone with similar problems.
The four dimensions reflect this goal.
Seeker Profile is placed first to ground the analysis in the person rather than the problem, and explicitly requires identifying existing strengths and coping resources.
Underlying Concern separates surface presentation from deeper need, capturing the gap between the presenting problem and the core wound, as well as the ``ask behind the as'', which is the implicit need that may diverge from the explicit request.
Historical Context incorporates retrieved post history selectively, only when it directly illuminates the current post, to prevent dilution of the focal analysis.
Finally, Response Blueprint translates the preceding analysis into concrete guidance for generation: validation points anchored in the seeker's own words, personalization anchors, tone calibration, and explicit ``avoid'' items.

\subsection{Stage 2: Draft Generation}

In this stage, we generate an initial draft response that reflects the seeker's personal context identified in the case formulation. The prompt used for draft generation is shown in Figure~\ref{fig:draft_generation_prompt}.

The draft generation prompt is designed to translate the case formulation into a response that reads as peer support rather than clinical guidance.
The core principles emphasize writing as a caring friend rather than a helper following a script, since therapy-speak and structured advice tend to create distance even when the underlying content is accurate. We instruct the model to validate before advising and to mirror the seeker's voice, reflecting the established finding that perceived understanding precedes openness to perspective-taking or suggestions. The \textit{Response Approach} section provides a fixed scaffold (open with something specific, validate, close with warmth) while leaving the middle adaptive to the seeker's identified need (witnessing, advice, grounding, celebration, or sitting with ambivalence), so that structure does not flatten contextual responsiveness.
The \textit{Using Case Formulation} section deliberately frames the formulation as guidance rather than a checklist, and explicitly instructs the model not to reference the analysis or force every element in, since visible adherence to a template undermines the felt sense of a personal reply.

\subsection{Stage 3: Collaborative Refinement}

This stage iteratively refines the response through three sub-steps: critique generation by three counselor agents, guidance synthesis, and response refinement.

\paragraph{Critique Generation.}
Three counselor agents, grounded in Cognitive Reframing, Unconditional Positive Regard, and Solution-Focused Counseling, independently critique the current response from their respective therapeutic perspectives.
All three agents share a common base prompt template, shown in Figure~\ref{fig:critique_generation_prompt_1}--\ref{fig:critique_generation_prompt_2}, into which a counselor-specific persona is inserted. The personas for the Cognitive Reframing, Unconditional Positive Regard, and Solution-Focused Counseling agents are shown in Figures~\ref{fig:agent_persona_prompt}, respectively.

The critique generation step is designed to identify refinement opportunities from multiple therapeutic perspectives.
We instantiate three counselor agents grounded in distinct therapeutic strategies, each examining the current response through a complementary lens.
The shared base prompt asks every agent to provide critique along two separate axes: one grounded in their specialized approach, and one from general counseling best practices such as empathy, rapport, pacing, and tone.
This separation prevents specialized methods from being forced where they do not naturally fit, while still ensuring that broadly applicable refinements are surfaced for every response.
The prompt further requires each agent to first assess the relevance of their specialization before offering suggestions, and to produce concrete, actionable feedback tied to specific changes in the response.

\input{latex/table/123_content_quality_table}

\input{latex/table/125_content_quality_analysis}

\paragraph{Guidance Synthesis.}
The critiques from the three counselor agents are aggregated and distilled into focused guidance containing at most two high-priority items.
The prompt used for guidance synthesis is shown in Figure~\ref{fig:guideline_synthesis_prompt_1}--\ref{fig:guideline_synthesis_prompt_2}.

The guidance synthesis step consolidates the three counselors' critiques into a focused set of revision targets rather than passing all suggestions forward in full.
We cap the output at two items because applying many simultaneous changes tends to degrade coherence and risks exceeding the response length budget, while a tightly scoped set of priorities yields more integrated revisions.
The prompt instructs the model to first list all distinct suggestions and identify consensus signals across counselors, then evaluate each on therapeutic priority and feasibility before selecting either one or two improvements.
We provide explicit criteria for choosing one versus two.
Two items are selected only when they are complementary in scope, such as validation paired with actionability, rather than overlapping, so that the pair addresses different dimensions of the response when applied together.
A prioritization hierarchy of safety, empathy gaps, contextual alignment, actionability, and stylistic refinement anchors the selection in therapeutic impact rather than surface-level polish.

\paragraph{Response Refinement.}
The current response is refined based on the synthesized guidance. The prompt used for response refinement is shown in Figure~\ref{fig:response_refinement_prompt_1}--\ref{fig:response_refinement_prompt_2}.

The response refinement step revises the current response according to the synthesized guidance while preserving what already works. The prompt frames revision as a targeted edit rather than a rewrite, instructing the model to identify effective elements in the current response, locate where each prioritized improvement should be applied, and integrate the changes without disturbing the surrounding text.
We retain the same 450-token cap used in draft generation so that improvements compete for space and the model is forced to make trade-offs rather than accumulating content across iterations.
The prompt also warns against introducing new issues, forcing techniques that do not fit, or producing text that reads as a checklist of applied feedback, since visible evidence of mechanical revision undermines the peer-support register.
An implementation note is requested alongside the revised response, mapping each change back to its corresponding guidance item, which provides a lightweight check that the prioritized improvements were actually incorporated.

%% file: latex/table/123_content_quality_table.tex
\begin{table*}[t]
    \centering
    \resizebox{0.9\textwidth}{!}{
        \begin{tabular}{cl||rrrrr}
        \toprule
          \textbf{Backbone Model} & \multicolumn{1}{c||}{\textbf{Method}} & \multicolumn{1}{c}{\textbf{Emp.}} & \multicolumn{1}{c}{\textbf{Hel.}} & \multicolumn{1}{c}{\textbf{Per.}} & \multicolumn{1}{c}{\textbf{Und.}} & \multicolumn{1}{c}{\textbf{Avg.}} \\
            
            \midrule
            \multirow{4}{*}{\makecell{Qwen2.5\\14B}} 
            & Vanilla & \valc{3.99}{0.01} & \valc{3.85}{0.01} & \valuc{4.27}{0.01} & \valc{4.08}{0.01} & \valc{4.05}{0.01}  \\
            & MentalAgora & \valc{3.97}{0.01} & \valbc{4.14}{0.01} & \valc{4.19}{0.01} & \valc{4.13}{0.01} & \valc{4.11}{0.01}  \\
            & ES-VR & \valbc{4.16}{0.01} & \valc{4.03}{0.01} & \valc{4.20}{0.01} & \valuc{4.25}{0.02} & \valuc{4.16}{0.01} \\
            & \myframework~& \valuc{4.13}{0.01} & \valuc{4.05}{0.01} & \valbc{4.41}{0.01} & \valbc{4.26}{0.01} & \valbc{4.21}{0.01}  \\
            \midrule
            \multirow{4}{*}{\makecell{Qwen2.5\\72B}} 
            & Vanilla & \valuc{4.09}{0.01} & \valc{4.05}{0.01} & \valuc{4.52}{0.01} & \valuc{4.25}{0.01} & \valuc{4.23}{0.01}  \\
            & MentalAgora & \valc{3.92}{0.01} & \valuc{4.16}{0.01} & \valc{4.24}{0.01} & \valc{4.13}{0.01} & \valc{4.11}{0.01}  \\
            & ES-VR & \valc{3.90}{0.01} & \valc{3.86}{0.01} & \valc{4.13}{0.01} & \valc{4.04}{0.02} & \valc{3.98}{0.01}  \\
            & \myframework~& \valbc{4.19}{0.01} & \valbc{4.17}{0.01} & \valbc{4.63}{0.01} & \valbc{4.39}{0.01} & \valbc{4.35}{0.01}  \\
            \midrule
            \multirow{4}{*}{Gemma3} 
            & Vanilla & \valc{4.47}{0.01} & \valc{4.06}{0.01} & \valuc{4.84}{0.01} & \valuc{4.52}{0.01} & \valc{4.47}{0.01}  \\
            & MentalAgora & \valc{4.52}{0.01} & \valbc{4.16}{0.01} & \valc{4.71}{0.01} & \valuc{4.52}{0.01} & \valuc{4.48}{0.01}  \\
            & ES-VR & \valuc{4.64}{0.01} & \valc{4.07}{0.02} & \valc{4.63}{0.01} & \valc{4.29}{0.01} & \valc{4.41}{0.01}  \\
            & \myframework~ & \valbc{4.74}{0.01} & \valuc{4.14}{0.01} & \valbc{4.91}{0.01} & \valbc{4.66}{0.01} & \valbc{4.61}{0.01}  \\
            \midrule
            \multirow{4}{*}{Gemma4}
            & Vanilla & \valc{4.65}{0.01} & \valc{4.41}{0.01} & \valuc{4.87}{0.01} & \valc{4.68}{0.01} & \valc{4.65}{0.01}   \\
            & MentalAgora & \valuc{4.88}{0.01} & \valbc{4.81}{0.01} & \valc{4.82}{0.01} & \valuc{4.71}{0.01} & \valbc{4.81}{0.01}   \\
            & ES-VR & \valc{3.96}{0.04} & \valc{3.76}{0.03} & \valc{4.05}{0.03} & \valc{3.87}{0.04} & \valc{3.91}{0.03}   \\
            & \myframework~& \valbc{4.89}{0.01} & \valuc{4.50}{0.01} & \valbc{4.92}{0.01} & \valbc{4.74}{0.01} & \valuc{4.76}{0.01}   \\
            \midrule
            \multirow{4}{*}{Mistral}
            & Vanilla & \valc{3.85}{0.01} & \valc{3.87}{0.01} & \valuc{4.30}{0.01} & \valc{4.02}{0.01} & \valc{4.01}{0.01}  \\
            & MentalAgora & \valc{3.70}{0.02} & \valuc{4.03}{0.01} & \valc{4.12}{0.01} & \valc{3.95}{0.02} & \valc{3.95}{0.01}  \\
            & ES-VR  & \valbc{4.35}{0.02} & \valc{3.96}{0.02} & \valc{4.16}{0.01} & \valuc{4.13}{0.02} & \valuc{4.15}{0.01} \\
            & \myframework~& \valuc{4.13}{0.02} & \valbc{4.10}{0.01} & \valbc{4.36}{0.01} & \valbc{4.25}{0.01} & \valbc{4.21}{0.01}  \\
            \midrule
            \multirow{4}{*}{Gemini}
            & Vanilla & \valc{3.57}{0.03} & \valc{2.94}{0.03} & \valc{3.32}{0.04} & \valc{3.11}{0.04} & \valc{3.24}{0.03}  \\
            & MentalAgora & \valuc{4.50}{0.01} & \valbc{4.26}{0.02} & \valuc{4.72}{0.01} & \valuc{4.53}{0.01} & \valuc{4.50}{0.01}  \\
            & ES-VR  & \multicolumn{1}{c}{--} & \multicolumn{1}{c}{--} & \multicolumn{1}{c}{--} & \multicolumn{1}{c}{--} & \multicolumn{1}{c}{--} \\
            & \myframework~& \valbc{4.61}{0.01} & \valuc{4.15}{0.01} & \valbc{4.86}{0.01} & \valbc{4.64}{0.01} & \valbc{4.56}{0.01}  \\
            \bottomrule
        \end{tabular}
    }
    \caption{LLM-as-a-Judge evaluation results on a 5-point Likert scale. Emp., Hel., Per., and Und. denote Empathy, Helpfulness, Personalization, and Understanding, respectively. Avg. denotes the average score across the four dimensions.}
    \label{tab:content_quality}
\end{table*}

%% file: latex/table/125_content_quality_analysis.tex
\begin{table*}[t]
\centering
\resizebox{\textwidth}{!}{
\begin{tabular}{ccl||rrrrr}
\toprule
\multicolumn{2}{c}{\textbf{Experiment}} & \multicolumn{1}{c||}{\textbf{Method}} & \multicolumn{1}{c}{\textbf{Emp.}} & \multicolumn{1}{c}{\textbf{Hel.}} & \multicolumn{1}{c}{\textbf{Per.}} & \multicolumn{1}{c}{\textbf{Und.}} & \multicolumn{1}{c}{\textbf{Avg.}} \\
\midrule
\multicolumn{2}{c}{\multirow{2}{*}{Stage-Level Ablation}}
& vs. w/o Ref.   & \valc{4.12}{0.01} & \valc{3.86}{0.01} & \valc{4.39}{0.01} & \valc{4.23}{0.01} & \valc{4.15}{0.01} \\
\multicolumn{2}{c}{}
& vs. w/o CF \& Ref. & \valc{3.93}{0.01} & \valc{3.87}{0.01} & \valc{4.27}{0.01} & \valc{4.10}{0.01} & \valc{4.04}{0.01} \\
\midrule
\multirow{7}{*}{\makecell{Case Formulation\\Analysis}}
& History Swap & vs. Swapped & \valc{3.55}{0.01} & \valc{3.42}{0.01} & \valc{3.56}{0.01} & \valc{3.33}{0.02} & \valc{3.46}{0.01} \\
\cmidrule{2-8}
& \multirow{2}{*}{Persona Comparison} & vs. Persona + Ref. & \valc{4.04}{0.01} & \valc{4.02}{0.01} & \valc{3.99}{0.01} & \valc{4.24}{0.01} & \valc{4.07}{0.01} \\
& & vs. Persona w/o Ref.  & \valc{4.10}{0.01} & \valc{3.81}{0.01} & \valc{4.03}{0.01} & \valc{4.26}{0.01} & \valc{4.05}{0.01} \\
\cmidrule{2-8}
& \multirow{4}{*}{Dimension Ablation} & vs. w/o Seeker Profile 
 & \valc{4.10}{0.01} & \valc{4.02}{0.01} & \valc{4.02}{0.01} & \valc{4.30}{0.01} & \valc{4.11}{0.01} \\
& & vs. w/o Underlying Concern  & \valc{4.11}{0.01} & \valc{4.01}{0.01} & \valc{4.01}{0.01} & \valc{4.26}{0.01} & \valc{4.10}{0.01} \\
& & vs. w/o Historical Context  & \valc{4.11}{0.01} & \valc{4.02}{0.01} & \valc{4.01}{0.01} & \valc{4.26}{0.01} & \valc{4.10}{0.01} \\
& & vs. w/o Response Blueprint  & \valc{4.08}{0.01} & \valc{4.05}{0.01} & \valc{4.02}{0.01} & \valc{4.27}{0.01} & \valc{4.10}{0.01} \\
\midrule
\multicolumn{2}{c}{\multirow{3}{*}{Guidance Synthesis Ablation}}
& vs. Single Agent  & \valc{4.06}{0.01} & \valc{4.05}{0.01} & \valc{4.01}{0.01} & \valc{4.22}{0.01} & \valc{4.09}{0.01} \\
\multicolumn{2}{c}{}
& vs. Random Selection  & \valc{4.06}{0.01} & \valc{3.99}{0.01} & \valc{4.01}{0.01} & \valc{4.20}{0.01} & \valc{4.07}{0.01} \\
\multicolumn{2}{c}{}
& vs. No Limit  & \valc{3.90}{0.02} & \valc{3.88}{0.01} & \valc{3.97}{0.01} & \valc{4.11}{0.01} & \valc{3.97}{0.01} \\
\midrule
\multicolumn{3}{c||}{Full \myframework} & \valc{4.13}{0.01} & \valc{4.05}{0.01} & \valc{4.41}{0.01} & \valc{4.26}{0.01} & \valc{4.21}{0.01} \\
\bottomrule
\end{tabular}
}
\caption{Ablation and analysis studies on \myframework~components. The Stage-Level Ablation removes the case formulation and/or refinement stages; the Case Formulation Analysis includes history swap, persona-based prompting comparison, and per-dimension removal of the four case formulation components; the Guidance Synthesis Ablation compares against Single Agent, Random Selection, and No Limit variants.}
\label{tab:ablation_scoring}
\end{table*}

%% file: latex/appendix/123_additional_evaluation.tex
\subsection{Likert-Scale Scoring}
\label{sec:likert_scale_scoring}

This evaluation assesses each quality dimension of the generated emotional support response on a 5-point Likert scale \cite{likert1932technique} using gpt-4o-mini as an LLM-as-a-Judge, following \citet{ye2025genericempathypersonalizedemotional} and \citet{dey2025gravityframeworkpersonalizedtext}.
We evaluate across four key dimensions: (1) Empathy, how well the response acknowledges and validates the seeker's emotions; (2) Helpfulness, how practical and actionable the advice is for the seeker; (3) Personalization, how well an emotional support response aligns with the seeker's needs; and (4) Understanding, how well the response demonstrates comprehension of the seeker's situation.
This evaluation takes the post history, target post, and generated emotional support response as input, and assigns a score on a 5-point Likert scale for each dimension.
To mitigate the stochasticity inherent in LLM-based evaluation, we run the judge three times for each item and average the resulting scores.

\subsubsection{Baseline Comparison}

We compare \myframework~against the baselines introduced in \S\ref{sec:baselines} across all six backbone models. Table~\ref{tab:content_quality} reports the per-dimension and averaged Likert scores, and we summarize the key findings as follows.

Across nearly all backbones, \myframework~attains the highest overall rating, with the only exception being Gemma4. This consistency across model families suggests that the gains are driven by the framework itself rather than backbone-specific advantages, and aligns with the pairwise comparison results in \S\ref{sec:results}.

Improvements concentrate in Personalization and Understanding, where \myframework~ is the top method on every backbone, with Empathy following closely behind.
This pattern is consistent with our design: Stage 1 explicitly constructs a case formulation covering the seeker's profile, underlying concerns, and historical context, which directly supports the Personalization and Understanding dimensions, while the Unconditional Positive Regard agent in Stage 3 contributes to Empathy.
Helpfulness is the only dimension where \myframework~ does not consistently lead, with MentalAgora often achieving the highest score.
However, \myframework~ remains competitive on this dimension and attains the highest averaged score on nearly all backbones.
The same pattern holds in our human evaluation and user study (\S\ref{sec:results}), indicating that \myframework's advantage comes from balanced strength across dimensions rather than dominance in any single one.

\subsubsection{Analysis Experiments}

We further apply Likert-scale scoring to the analysis experiments introduced in \S\ref{sec:analysis}: stage-level ablation, case formulation analysis, and guidance synthesis ablation, to examine whether the trends observed under pairwise comparison are also reflected in absolute scoring ratings across the four dimensions. Following the setup in \S\ref{sec:analysis}, we use Qwen2.5-14B as the backbone, reuse the same generated responses, and apply the Likert-scale protocol described above. Table~\ref{tab:ablation_scoring} reports the per-dimension and averaged scores.

\paragraph{Stage-Level Ablation.}
Removing the refinement stage (w/o Ref.) reduces the average score from 4.21 to 4.15, and further removing case formulation (w/o CF \& Ref.) brings it down to 4.04. The monotonic degradation across all four dimensions aligns with the pairwise comparison results in \S\ref{sec:stage_ablation}, reinforcing that both stages contribute complementary value: case formulation establishes seeker-grounded understanding, while iterative refinement polishes responses based on that foundation.

\paragraph{Case Formulation Analysis.}
The history swap variant exhibits a substantial drop to 3.46 on average, with the largest decrease in Understanding (3.33). This sharp degradation, far exceeding any other ablation, indicates that case formulation is not merely a generic context augmentation but encodes seeker-specific information whose mismatch actively harms response quality. Both persona-based variants (4.05--4.07) underperform \myframework~(4.21), consistent with the pairwise finding that structured case formulation captures deeper longitudinal and motivational factors beyond surface-level persona traits. For per-dimension ablation, removing any single component reduces the average score to 4.10--4.11, with relatively uniform drops across dimensions. The Likert-scale scoring shows comparable contributions from all four dimensions. This suggests that each dimension provides a roughly equivalent absolute quality contribution when responses are rated independently.

\paragraph{Guidance Synthesis Ablation.}
The No Limit variant shows the largest performance drop (3.97), notably lower than both Single Agent (4.09) and Random Selection (4.07). This pattern aligns with the pairwise comparison results and confirms that unconstrained feedback volume is the most detrimental factor, as applying all critiques without prioritization introduces conflicting or low-priority signals that degrade response quality. The smaller gap between Single Agent and Random Selection further indicates that multi-agent diversity and prioritized selection both contribute, but controlling feedback volume through top-$k$ filtering is the dominant design choice.

\paragraph{Summary.}
Across all three analysis experiments, the Likert-scale scoring trends are consistent with the pairwise comparison findings. The convergence of these two evaluation protocols strengthens the validity of our component-level design choices in \myframework.

\subsection{Effect of Diverse Therapeutic Perspectives}

\input{latex/table/123_additional_evaluation_counselor_ablation}

\input{latex/figure/refinement_rounds}

In the Collaborative Refinement stage, \myframework~employs three counselor agents specializing in distinct therapeutic strategies, which are Cognitive Reframing, Unconditional Positive Regard, and Solution-Focused, to provide critiques from diverse perspectives. These critiques are synthesized into focused guidance, which is then used to iteratively refine the response. To investigate whether this multi-perspective feedback mechanism contributes to generating personalized emotional support, we conduct an ablation study by varying the composition of counselor agents.

% Table~\ref{tab:pas_scores_ablation_counselor} presents the results using Personalization Alignment Score (PAS), which infers the seeker's characteristics from their post history and measures how well the generated response aligns with the seeker's preferences on a 1–5 scale.

Table~\ref{tab:pairwise_comparison_ablation_counselor} reports the results of pairwise comparison using LLM-as-a-Judge, which directly compares responses from the full \myframework~against each ablation variant from the seeker's perspective, reporting the win, loss, and tie rates. \myframework~achieves higher win rates against all ablation variants, with win rates ranging from 51.48\% to 61.90\%. The comparable magnitude across variants indicates that all three counseling perspectives contribute to response quality, with the effectiveness stemming from their joint use rather than from any single dominant perspective.

\subsection{Analysis of Refinement Rounds}
\label{sec:analysis_refinement_round}

To examine how response quality evolves across refinement rounds, we compare \myframework's response at each round against three baselines (Vanilla, MentalAgora, and ES-VR) using the pairwise comparison.
As shown in Figure~\ref{fig:vs_vanilla_mentalagora_es_vr}, \myframework~consistently outperforms all baselines across the three backbone models (Qwen2.5-14B, Gemma3, Mistral), with the margin widening as refinement progresses.
The trend is most pronounced when compared against Vanilla and MentalAgora: on Qwen2.5-14B and Gemma3, \myframework~starts near or even below the baseline at Round-0 but steadily surpasses it, reflecting that early-round responses have yet to incorporate seeker-specific characteristics.
Against ES-VR, \myframework~already achieves a substantial lead from Round-0 and maintains it throughout, indicating that even minimally refined personalized responses are preferred over ES-VR's outputs.
Across all comparisons, the improvement begins to plateau after Round-4, with marginal gains in subsequent rounds.
Considering that computational cost scales linearly with the number of rounds (see Appendix~\ref{apdx:computational_costs}), we adopt the Round-4 response as the final output in the experiments, balancing personalization quality with efficiency.

%% file: latex/table/123_additional_evaluation_counselor_ablation.tex
% \begin{table}[t]
%     \centering
%     \resizebox{0.7\columnwidth}{!}
%     {
%     \begin{tabular}{l|r}
%     \toprule
%      & \multicolumn{1}{c}{\texttt{Qwen}}      \\ 
%     \midrule
%     \myframework & \valc{4.41}{0.01} \\
%     no Reframing & \valc{4.36}{0.01}  \\
%     no Regard & \valc{4.33}{0.01}  \\
%     no Solution & \valuc{4.47}{0.01} \\
%     only Reframing & \valc{4.42}{0.01}  \\
%     only Regard & \valbc{4.50}{0.01}  \\
%     only Solution & \valc{4.28}{0.01} \\
%     \bottomrule
% \end{tabular}
%     }
%     \caption{Ablation study on counselor agent composition using Personalization Alignment Score.}
%     \label{tab:pas_scores_ablation_counselor}
% \end{table}

\begin{table}[t!]
    \centering
    \resizebox{\columnwidth}{!}{
    \begin{tabular}{l|rrr}
        \toprule
        \multicolumn{1}{c|}{\textbf{Method}} & \multicolumn{1}{c}{\textbf{Wins}} & \multicolumn{1}{c}{\textbf{Losses}} & \multicolumn{1}{c}{\textbf{Ties}} \\
        \midrule
        vs. no Reframing & \valbc{51.48}{1.13} & \valc{48.45}{1.13} & \valc{0.06}{0.04} \\
        vs. no Regard & \valbc{59.31}{1.13} & \valc{40.62}{1.13} & \valc{0.06}{0.04} \\
        vs. no Solution & \valbc{51.90}{1.12} & \valc{47.97}{1.11} & \valc{0.13}{0.05} \\
        vs. only Reframing & \valbc{60.11}{1.12} & \valc{39.85}{1.12} & \valc{0.04}{0.03} \\
        vs. only Regard & \valbc{54.85}{1.18} & \valc{45.13}{1.18} & \valc{0.02}{0.02} \\
        vs. only Solution & \valbc{61.90}{1.08} & \valc{38.10}{1.08} & \valc{0.00}{0.00} \\

        \bottomrule
    \end{tabular}
    }
    \caption{Pairwise comparison results using LLM-as-a-Judge between \myframework~and ablation variants. Win, Loss, and Tie indicate the percentage of cases where \myframework~was preferred, the variant was preferred, or neither was preferred, respectively.
    % All differences are statistically significant (p < 0.001).
    }
    \label{tab:pairwise_comparison_ablation_counselor}
\end{table}

%% file: latex/figure/refinement_rounds.tex
\begin{figure*}[t!]
    \centering
    % Row 1: vs Vanilla
    \begin{subfigure}[b]{0.32\textwidth}
        \includegraphics[width=\textwidth]{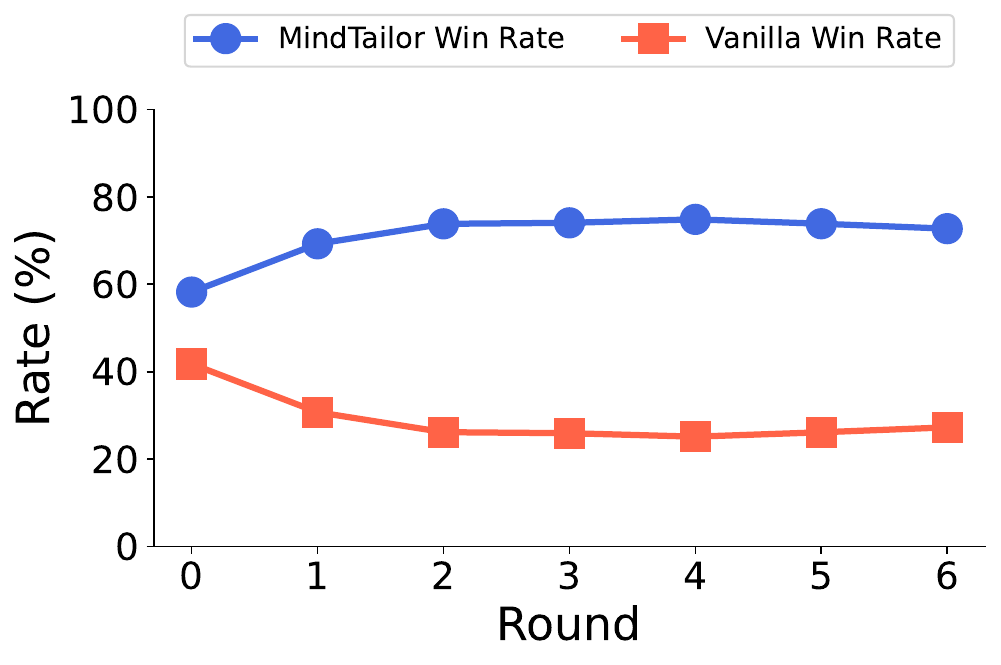}
        \caption{vs.\ Vanilla (Qwen2.5-14B)}
        \label{fig:vs_vanilla_qwen}
    \end{subfigure}
    \hfill
    \begin{subfigure}[b]{0.32\textwidth}
        \includegraphics[width=\textwidth]{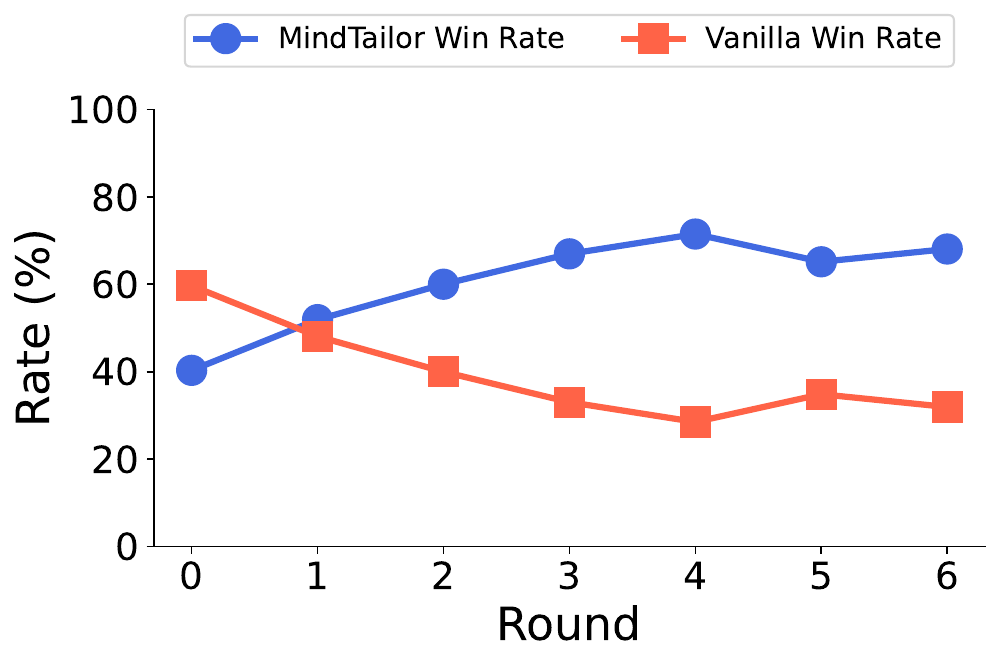}
        \caption{vs.\ Vanilla (Gemma3)}
        \label{fig:vs_vanilla_gemma}
    \end{subfigure}
    \hfill
    \begin{subfigure}[b]{0.32\textwidth}
        \includegraphics[width=\textwidth]{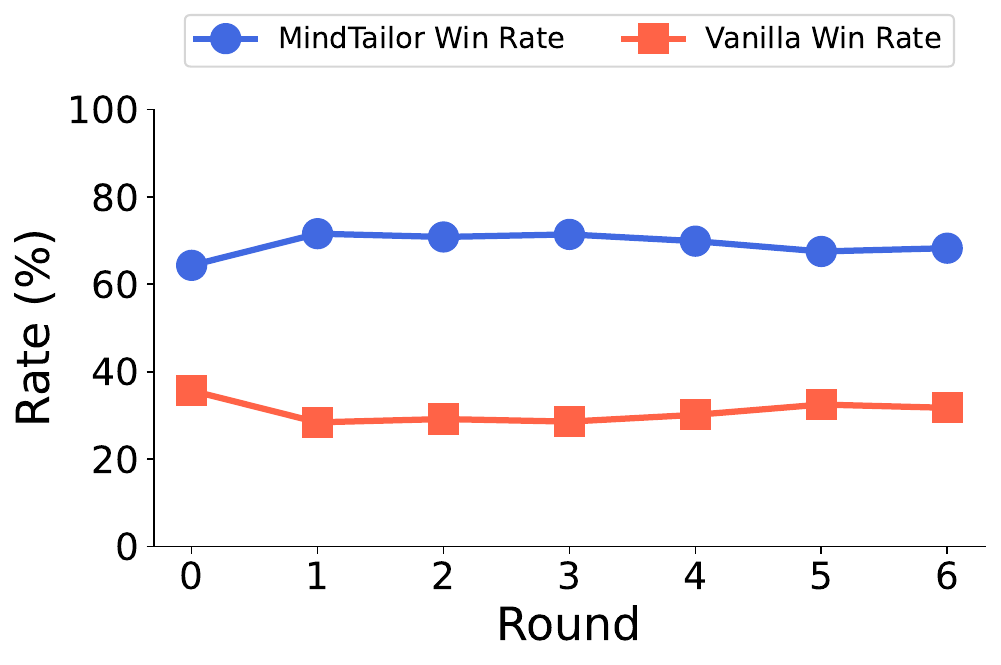}
        \caption{vs.\ Vanilla (Mistral)}
        \label{fig:vs_vanilla_mistral_nemo}
    \end{subfigure}
    \vskip\baselineskip
    % Row 2: vs MentalAgora
    \begin{subfigure}[b]{0.32\textwidth}
        \includegraphics[width=\textwidth]{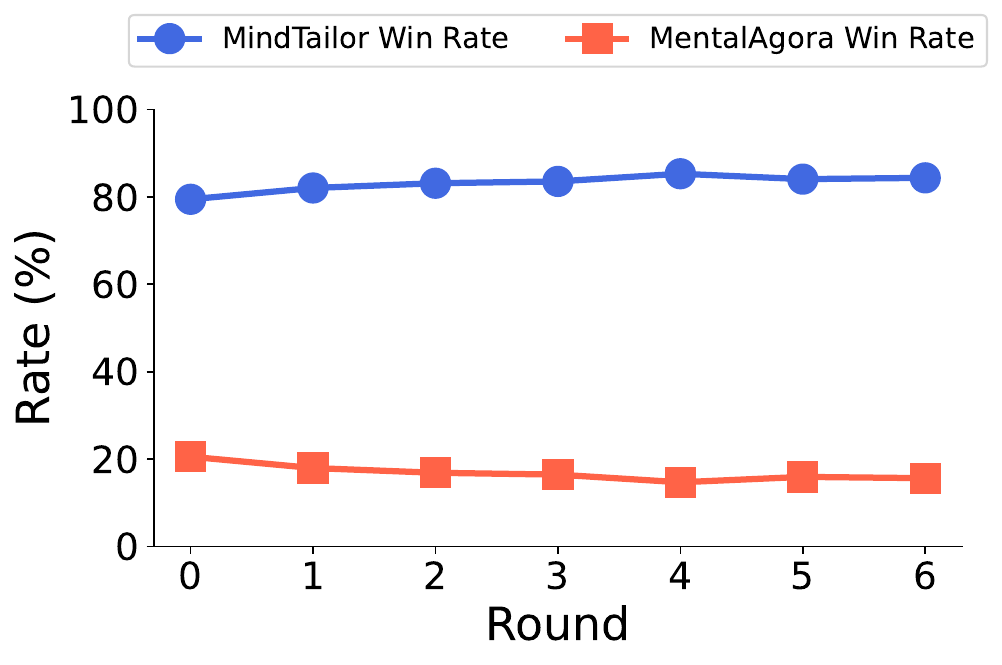}
        \caption{vs.\ MentalAgora (Qwen2.5-14B)}
        \label{fig:vs_mentalagora_qwen}
    \end{subfigure}
    \hfill
    \begin{subfigure}[b]{0.32\textwidth}
        \includegraphics[width=\textwidth]{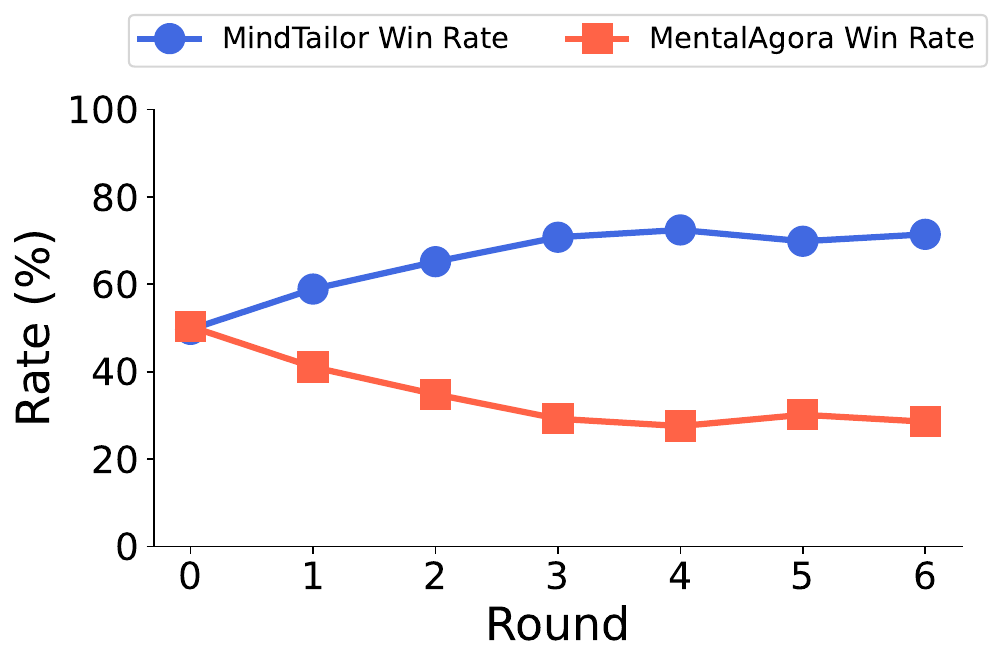}
        \caption{vs.\ MentalAgora (Gemma3)}
        \label{fig:vs_mentalagora_gemma}
    \end{subfigure}
    \hfill
    \begin{subfigure}[b]{0.32\textwidth}
        \includegraphics[width=\textwidth]{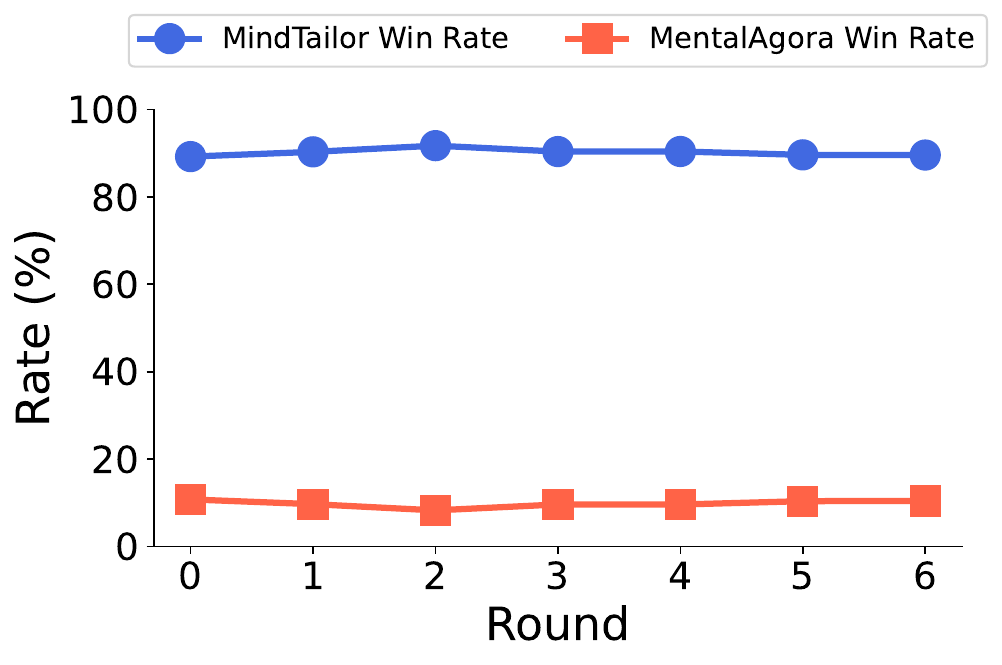}
        \caption{vs.\ MentalAgora (Mistral)}
        \label{fig:vs_mentalagora_mistral}
    \end{subfigure}
    \vskip\baselineskip
    % Row 3: vs ES-VR
    \begin{subfigure}[b]{0.32\textwidth}
        \includegraphics[width=\textwidth]{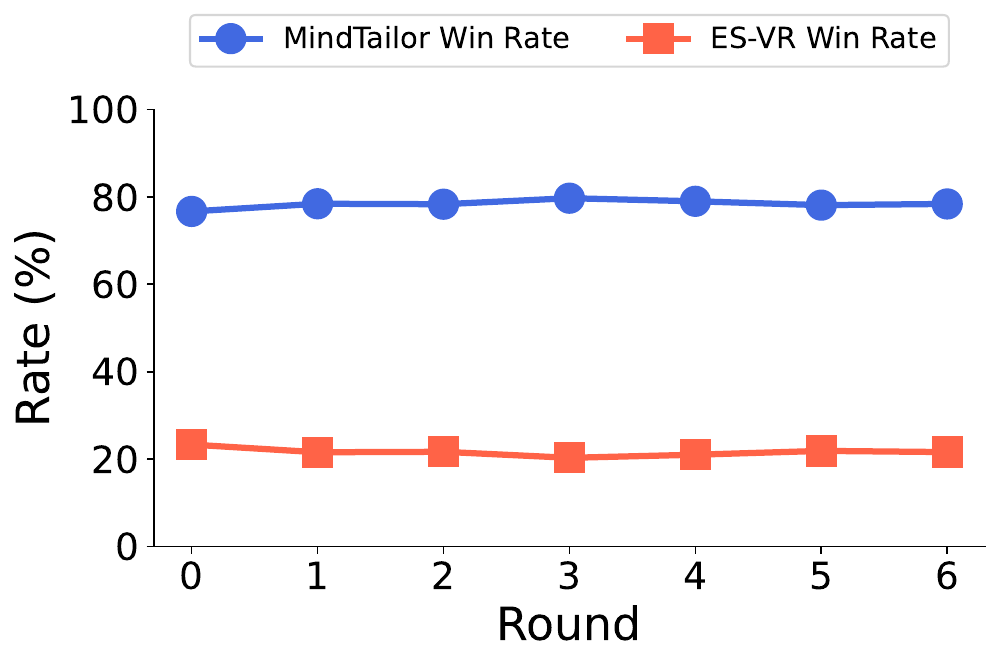}
        \caption{vs.\ ES-VR (Qwen2.5-14B)}
        \label{fig:vs_esvr_qwen}
    \end{subfigure}
    \hfill
    \begin{subfigure}[b]{0.32\textwidth}
        \includegraphics[width=\textwidth]{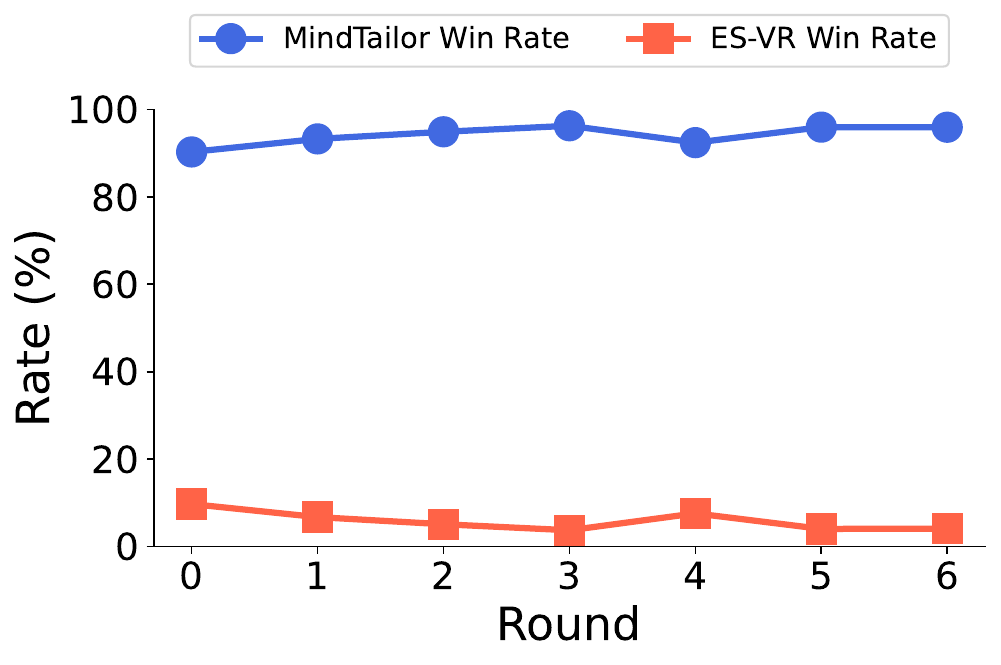}
        \caption{vs.\ ES-VR (Gemma3)}
        \label{fig:vs_esvr_gemma}
    \end{subfigure}
    \hfill
    \begin{subfigure}[b]{0.32\textwidth}
        \includegraphics[width=\textwidth]{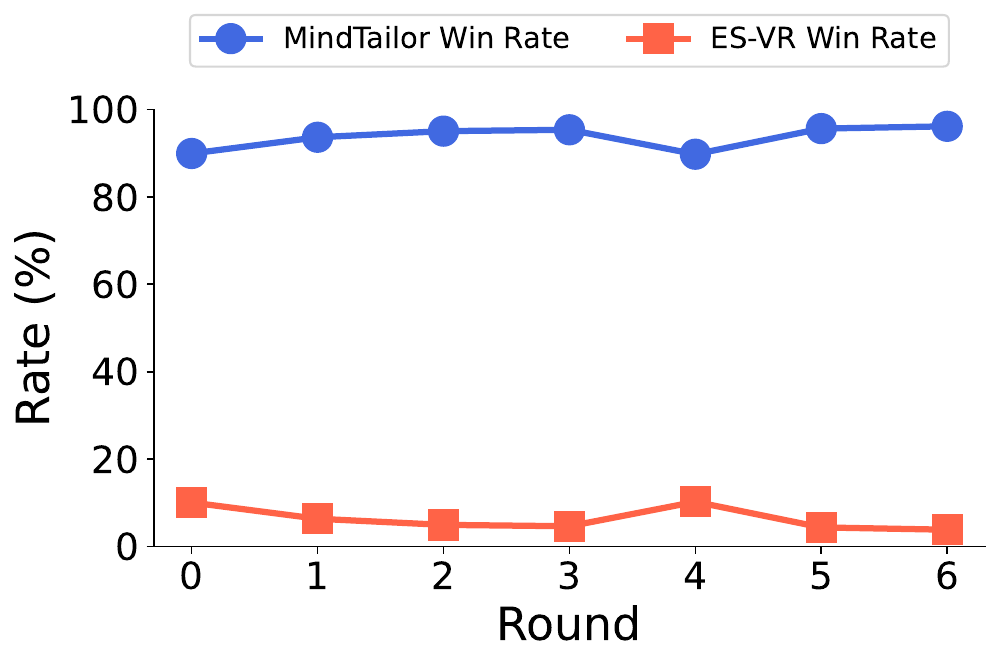}
        \caption{vs.\ ES-VR (Mistral)}
        \label{fig:vs_esvr_mistral}
    \end{subfigure}
    \caption{
    Pairwise comparison win rates of \myframework~against three baselines (Vanilla, MentalAgora, ES-VR) across refinement rounds for three backbone models. \myframework~(blue) shows consistent improvement over each baseline (red) as rounds progress, with performance plateauing around Round-4. This pattern supports adopting Round-4 responses as the final output to balance personalization quality with computational efficiency.
    }
    \label{fig:vs_vanilla_mentalagora_es_vr}
\end{figure*}

%% file: latex/appendix/193_llm_as_a_judge_reliability.tex
Since our main results rely heavily on LLM-as-a-Judge for pairwise comparison, we assess the reliability of these evaluations along two axes following \citet{10.5555/3666122.3668142}: (i) \textit{inter-judge agreement}, measuring consistency across different judge models, and (ii) \textit{judge--human agreement}, measuring alignment between LLM judges and expert human evaluators.
We report two complementary metrics for each axis: \textit{exact agreement rate}, the proportion of items on which the two annotators (LLM judges, or an LLM judge and a human evaluator) assign the same label among \{MindTailor wins, baseline wins, tie\}, and \textit{directional agreement rate}, the same proportion computed only over items where both annotators give a non-tie label, following the ``without tie'' setup in \citet{10.5555/3666122.3668142}.
As a reference, \citet{10.5555/3666122.3668142} report human--human agreement of approximately 0.63 (with ties) and 0.81 (without ties), which we use as a reference point for both metrics throughout this section.

\subsection{Inter-Judge Agreement}
\label{sec:inter-judge}

We compare the three judge models used throughout the paper (gpt-4o-mini, deepseek-v4-flash, and glm-4.5-air) on the pairwise comparison setting, aggregating across all baselines and backbones. Table~\ref{tab:inter-judge} reports pairwise exact and directional agreement rates.

\begin{table}[t]
\centering
\resizebox{\columnwidth}{!}{
\begin{tabular}{lcc}
\toprule
\textbf{Judge Pair} & \textbf{Exact} & \textbf{Directional} \\
\midrule
gpt-4o-mini vs. deepseek-v4-flash & 0.66 & 0.82 \\
gpt-4o-mini vs. glm-4.5-air       & 0.71 & 0.86 \\
deepseek-v4-flash vs. glm-4.5-air & 0.76 & 0.91 \\
\bottomrule
\end{tabular}
}
\caption{Inter-judge agreement rates on pairwise comparisons, aggregated across all baselines and backbone models. Exact agreement includes ties as a label, while directional agreement is computed only over items where neither judge labels a tie.}
\label{tab:inter-judge}
\end{table}

% \begin{table}[t]
% \centering
% \resizebox{\columnwidth}{!}{
% \begin{tabular}{lcc}
% \toprule
% \textbf{Judge Pair} & \textbf{Exact} & \textbf{Directional} \\
% \midrule
% gpt-4o-mini vs. deepseek-v4-flash & 0.66 & 0.82 \\
% human--human & 0.63 & 0.81 \\
% \bottomrule
% \end{tabular}
% }
% \caption{Inter-judge agreement rates on pairwise comparisons, aggregated across all baselines and backbone models. Exact agreement includes ties as a label, while directional agreement is computed only over items where neither judge labels a tie.}
% \label{tab:inter-judge}
% \end{table}

All three pairs achieve exact agreement above 0.65 and directional agreement above 0.80, with the highest agreement between deepseek-v4-flash and glm-4.5-air (0.76 exact, 0.91 directional).
Both metrics are comparable to the human--human agreement levels reported by \citet{10.5555/3666122.3668142}, suggesting that our pairwise comparison results generalize across judge models rather than depending on any single one.
% Both metrics exceed the human--human agreement levels reported by \citet{10.5555/3666122.3668142}, suggesting that our pairwise comparison results generalize across judge models rather than depending on any single one.

\subsection{Judge--Human Agreement}
\label{sec:judge-human}

The expert human evaluation uses 5-point Likert ratings across four dimensions plus an overall ranking among four methods, whereas LLM-as-a-Judge produces pairwise \{win, lose, tie\} labels. To enable direct comparison, we convert each human evaluation into pairwise judgments using two independent procedures:

\begin{itemize}
    \item \textbf{Average-based conversion.} For each item and each baseline, we compute the difference between the human-rated average score (across the four dimensions) of \myframework~and the baseline. A positive difference is mapped to a \myframework win, a negative difference to a baseline win, and an exact tie to tie.
    \item \textbf{Rank-based conversion.} For each item, the relative ordering of \myframework~and a baseline in the expert's overall ranking is mapped to win/lose/tie analogously.
\end{itemize}

For each judge model, we compute both the exact agreement rate (including ties) and the directional agreement rate (excluding ties) between the judge's pairwise label and the converted human label, averaged across the three baselines. Results are shown in Table~\ref{tab:judge-human}.

% \begin{table}[t]
% \centering
% \resizebox{\columnwidth}{!}{
% \begin{tabular}{lcccc}
% \toprule
% & \multicolumn{2}{c}{\textbf{Avg.-based}} & \multicolumn{2}{c}{\textbf{Rank-based}} \\
% \cmidrule(lr){2-3} \cmidrule(lr){4-5}
% \textbf{Judge Model} & Exact & Direct. & Exact & Direct. \\
% \midrule
% gpt-4o-mini         & 0.64 & 0.74 & 0.57 & 0.66 \\
% deepseek-v4-flash   & 0.60 & 0.70 & 0.49 & 0.58 \\
% human--human        & 0.63 & 0.81 & 0.63 & 0.69 \\
% \bottomrule
% \end{tabular}
% }
% \caption{Judge--human agreement rates, averaged across baselines, under two conversion procedures from human Likert/rank annotations to pairwise \{win, lose, tie\} labels. Exact agreement includes ties, while directional agreement excludes them.}
% \label{tab:judge-human}
% \end{table}

\begin{table}[t]
\centering
\resizebox{\columnwidth}{!}{
\begin{tabular}{lcccc}
\toprule
& \multicolumn{2}{c}{\textbf{Avg.-based}} & \multicolumn{2}{c}{\textbf{Rank-based}} \\
\cmidrule(lr){2-3} \cmidrule(lr){4-5}
\textbf{Judge Model} & Exact & Direct. & Exact & Direct. \\
\midrule
gpt-4o-mini         & 0.64 & 0.74 & 0.57 & 0.66 \\
deepseek-v4-flash   & 0.60 & 0.70 & 0.49 & 0.58 \\
glm-4.5-air         & 0.69 & 0.77 & 0.63 & 0.69 \\
\bottomrule
\end{tabular}
}
\caption{Judge--human agreement rates, averaged across baselines, under two conversion procedures from human Likert/rank annotations to pairwise \{win, lose, tie\} labels. Exact agreement includes ties, while directional agreement excludes them.}
\label{tab:judge-human}
\end{table}

Across all three judges, exact agreement falls in the 0.49--0.69 range and directional agreement in the 0.58--0.77 range.
Under the average-based conversion, judge--human agreement reaches 0.69 (exact) and 0.77 (directional), which falls slightly below the human--human reference. glm-4.5-air aligns most closely with experts under both conversion procedures, while gpt-4o-mini is also competitive on the average-based conversion.

\subsection{Summary}

Two findings support the reliability of our automatic evaluation. First, three heterogeneous LLM judges agree with each other above human--human agreement levels reported in prior work, on both exact and directional metrics, indicating that our pairwise results are robust to the choice of judge model. Second, while judge--human agreement (up to 0.69 exact, 0.77 directional) falls slightly below the human--human reference, all three judges consistently identify \myframework~as the preferred method, mirroring the conclusions drawn by expert evaluators. Together with the convergent direction of results across LLM-as-a-Judge, expert human evaluation, and the user study in \S\ref{sec:results}, this provides converging evidence that \myframework's improvements are not an artifact of any single evaluation protocol.

In particular, gpt-4o-mini, which we use as the judge for all analysis experiments (\S\ref{sec:analysis}), achieves exact agreement of 0.64 and directional agreement of 0.74 with expert annotations under the average-based conversion. While the directional score falls slightly below the human--human reference (0.81 directional), the exact agreement matches the corresponding reference (0.63), and gpt-4o-mini's pairwise judgments consistently align with experts in identifying \myframework~as the preferred method.

%% file: latex/appendix/195_evaluation_prompts.tex
In this appendix, we provide the full prompts used for our LLM-as-a-Judge evaluations.
We conduct two types of automatic evaluations to assess the quality of emotional support responses generated by \myframework and the baselines: (1) \textbf{Pairwise comparison} (\S\ref{sec:evaluation_approaches}), where a judge model selects the better response between \myframework and a baseline given the seeker's post history and target post, and (2) \textbf{Likert-scale scoring} (\S\ref{sec:likert_scale_scoring}), where a judge model rates each response on a 5-point scale across four quality dimensions (Empathy, Helpfulness, Personalization, and Understanding). To mitigate position bias~\cite{wang-etal-2024-large-language-models-fair}, the pairwise comparison is conducted three times in the original order and three times in the reversed order. The following subsections present the exact prompts provided to the judge models for each evaluation setting. Likewise, to account for the stochasticity of the judge model, Likert-scale scoring is also conducted three times per response, and the average score is reported.

\subsection{Prompt for Pairwise Comparison}
\label{appendix:prompt_pairwise}

Figure~\ref{fig:pairwise_comparison_prompt} is used for pairwise comparison between \myframework and a baseline. Following the evaluation perspective of \citet{dey2025gravityframeworkpersonalizedtext}, we design a prompt that guides the judge model through three stages: (1) analyzing the seeker's profile from their post history, (2) reading the target post and the two candidate responses, and (3) selecting the response that better aligns with the inferred profile, or returning a tie. The judge outputs one of three labels: \texttt{A is better}, \texttt{B is better}, or \texttt{Tie}, along with a brief justification.

\subsection{Prompts for Likert-Scale Scoring}
\label{appendix:prompt_likert}

For Likert-scale scoring, we use four separate prompts, one for each quality dimension. The prompts for Empathy, Helpfulness, and Understanding are minor adaptations of those used by \citet{ye2025genericempathypersonalizedemotional}, and are shown in Figures~\ref{fig:empathy_prompt}, \ref{fig:helpfulness_prompt}, and \ref{fig:understanding_prompt}, respectively. The prompt for Personalization, shown in Figure~\ref{fig:personalization_prompt}, is designed following the evaluation perspective of \citet{dey2025gravityframeworkpersonalizedtext}. Specifically, it guides the judge model through three stages: (1) analyzing the author's profile from their post history, (2) reading the target post and the generated support response, and (3) rating how well the response aligns with the inferred profile on a 5-point Likert scale.

%% file: latex/appendix/125_human_evaluation_guideline.tex
We recruited 5 psychology experts with graduate-level training (master's or above) to ensure professional assessment quality. The evaluators consisted of four females and one male, ranging in age from their mid-20s to early 30s. Evaluators were recruited through professional networks based on these eligibility criteria. All evaluators were Korean nationals with sufficient English proficiency to comprehend the evaluation materials, which were presented in English.  They were compensated at approximately \$3.50 USD per sample, based on an estimated 20 minutes per sample, corresponding to an hourly rate of approximately \$10.50 USD, which exceeds the local minimum wage in South Korea.

\subsection{Evaluation Setup}

Each evaluator assessed responses generated by our framework and baseline methods using Qwen2.5-14B as the backbone model for 100 samples from \mydataset. For each sample, evaluators were presented with the seeker's post along with four generated emotional support responses. To mitigate potential order bias, the presentation order of responses was randomized for each sample. Evaluators scored each response on four dimensions using a 5-point Likert scale, and then ranked all four responses.

\subsection{Evaluation Dimensions}
\label{sec:human_eval_dimensions}

\paragraph{Empathy} evaluates how well the response acknowledges and validates the seeker's emotions. A score of 1 indicates dismissive or ignorant responses, while a score of 5 indicates deep emotional connection and validation.

\paragraph{Helpfulness} assesses how practical and actionable the advice is for the seeker. A score of 1 indicates no practical value, while a score of 5 indicates highly practical guidance with clear action steps.

\paragraph{Personalization} evaluates how well the generated support aligns with the seeker's preferences. A score of 1 indicates generic responses ignoring preferences, while a score of 5 indicates responses perfectly tailored to individual preferences.

\paragraph{Understanding} assesses how well the response demonstrates comprehension of the seeker's situation. A score of 1 indicates a complete misunderstanding, while a score of 5 indicates a deep comprehension of all aspects and nuances.

\paragraph{Rank} assesses the overall preference ordering of different support responses based on the seeker's personal characteristics and situation. A rank of 1 indicates the most preferred response, while a rank of 4 indicates the least preferred response.

\subsection{Two-Stage Evaluation with Refined Criteria}
To ensure evaluation reliability, we adopted a two-stage evaluation procedure, following established practices in qualitative coding research where iterative codebook revision and recoding are standard for improving inter-coder reliability~\citep{hruschka2004reliability}.
After observing low inter-annotator agreement in the initial round, we identified high-disagreement items by (1) selecting the top 40\% of samples by score standard deviation across evaluators, and (2) within these samples, designating evaluators whose scores deviated from the median by more than 1 point as re-evaluators (or the single evaluator with the largest deviation if none exceeded the threshold; all five evaluators if three or more exceeded it). For the re-evaluation, we refined the scoring rubrics with more operationalized descriptions and anchor examples for each Likert level. The final inter-annotator agreement, measured by ICC(2,k), was 0.645, indicating good agreement~\citep{cicchetti1994guidelines}.

\subsection{Evaluator Safeguards}

All evaluators provided informed consent prior to the evaluation and were informed that their responses would be used for academic research purposes. Evaluators were informed that the evaluation materials may contain content related to mental health concerns or personal distress, which could cause emotional discomfort. They were assured that they could withdraw from the study at any time without penalty. Evaluator confidentiality was guaranteed throughout the study.

%% file: latex/appendix/150_user_study_guideline.tex
We recruited 50 participants for the user study. Participants consisted of 32 females and 18 males, ranging in age from their 20s to 30s. Participants were recruited through personal networks and online recruitment announcements. All participants were Korean nationals with sufficient English proficiency to comprehend the study materials, which were presented in English. They were compensated with approximately \$7.00 USD upon completing all procedures. The average completion time was approximately 30 minutes, corresponding to an hourly rate of approximately \$14.00 USD, which exceeds the local minimum wage in South Korea.

\subsection{Data Collection}
Participant data was collected via Google Forms in two stages. First, participants submitted 3--5 social media posts that reveal their daily life, thoughts, and emotions. Second, participants described a current mental health-related concern they were experiencing in 1,000--2,000 characters, with instructions to write as specifically and in as much detail as possible. All data was used solely for the purposes of this study and was discarded upon completion of the research.

\subsection{Evaluation Setup}
Using the collected data, responses were generated by each baseline method and our proposed framework with Qwen2.5-14B as the backbone model. Participants evaluated these responses through a custom web interface. To mitigate potential order bias, the presentation order of responses was randomized for each participant. Participants rated each response on a 5-point Likert scale across six dimensions, and then ranked all four responses.

\subsection{Evaluation Dimensions}

We adopted six self-report dimensions drawn from prior work on conversational and supportive AI evaluation. Participants rated each support response on a 5-point Likert scale, where 1 indicates the lowest and 5 the highest level of agreement with the corresponding statement.

\paragraph{Emapthy} \cite{Abbasian2024} captures the degree to which participants felt that the response understood their emotional state and offered sincere comfort. A score of 1 indicates that the participant felt no empathy or comfort from the response at all, while a score of 5 indicates that the participant felt deeply understood and sincerely comforted.

\paragraph{Perceived Helpfulness} \cite{informatics12010033} captures the degree to which participants felt that the response actually helped address their concerns or emotions. A score of 1 indicates that the participant felt no help from the response at all, while a score of 5 indicates that the participant felt the response was genuinely useful in resolving their concerns or easing their emotional distress.

\paragraph{Personalization} \cite{Abbasian2024} captures the degree to which participants felt that the response was tailored to them rather than offering generic advice. A score of 1 indicates that the participant felt the response was something that could be said to anyone, while a score of 5 indicates that the participant felt the response was distinctly customized to them.

\paragraph{Relevance} \cite{liu-etal-2021-towards} captures the degree to which participants felt that the response resonated with them personally and fit their own situation. A score of 1 indicates that the participant felt the response was unrelated or off-topic, while a score of 5 indicates that the participant felt the response was strikingly personal, as if it understood them precisely.

\paragraph{Trustworthiness \& Safety} \cite{Abbasian2024} captures the degree to which participants felt psychologically safe and able to trust the response. A score of 1 indicates that the participant felt uncomfortable or even unsettled by the response, while a score of 5 indicates that the participant felt highly trusting and psychologically at ease.

\paragraph{Willingness to Reuse} \cite{app14135889} captures the degree to which participants would want to receive this kind of response again in the future. A score of 1 indicates that the participant would not want to receive such a response again, while a score of 5 indicates that the participant would strongly want to receive this kind of response in the future.

\paragraph{Rank} captures each participant's overall preference ordering across the four support responses they received. After rating all responses on the six Likert-scale dimensions above, participants were asked to consider all responses together and assign a unique rank from 1 (most preferred) to 4 (least preferred), with each rank used exactly once.

\subsection{Participant Safeguards}
All participants provided informed consent prior to the study and were informed that their responses would be used for academic research purposes. Participants were informed that the study materials involve content related to mental health concerns or personal distress, which could cause emotional discomfort. They were assured that they could withdraw from the study at any time without penalty. Participant confidentiality was guaranteed throughout the study.

%% file: latex/table/170_implementation_details_model_info.tex
\begin{table*}[t!]
\centering
\small
\resizebox{\textwidth}{!}{
\begin{tabular}{lcccc}
\toprule
\multicolumn{1}{c}{\textbf{Models}} & \textbf{Parameter Size} & \textbf{Instruction Tuned} & \textbf{Is Proprietary} & \textbf{License} \\
\midrule
\texttt{Qwen2.5-14B-Instruct} & 15B & O & X & Apache-2.0 \\
\texttt{Qwen2.5-72B-Instruct} & 73B & O & X & Qwen \\
\texttt{gemma-3-27b-it} & 27B & O & X & Gemma Terms of Use \\
\texttt{gemma-4-31b-it} & 33B & O & X & Apache-2.0 \\
\texttt{Mistral-Nemo-Instruct-2407} & 12B & O & X & Apache-2.0 \\
\texttt{GLM-4.5-Air} & 110B & O & X & MIT \\
\texttt{deepseek-v4-flash} & 158B & O & X & MIT \\
\texttt{gemini-2.5-flash-lite} & Not Disclosed & O & O & Google Cloud Platform Terms of Service \\
\texttt{gpt-4o-mini} & Not Disclosed & O & O & OpenAI Terms of Use \\
\texttt{text-embedding-3-small} & Not Disclosed & -- & O & OpenAI Terms of Use \\

\bottomrule
\end{tabular}
}
\caption{Overview of language models used in this paper, including their parameter sizes, instruction-tuning status, and licensing information. Models with an ``O'' in the Instruction Tuned column have been fine-tuned for following instructions, while those with an ``X'' in the Is Proprietary column are open-source.}
\label{tab:models}
\end{table*}

%% file: latex/appendix/160_case_study.tex
This section presents additional examples to further illustrate how \myframework's context profile enables personalized emotional support across different types of situations.

% \subsection{Case 1: Anxiety During Unstructured Time}
\paragraph{Case 1: Anxiety During Unstructured Time}

In Table~\ref{tab:case_study_appendix}, the seeker describes worsening anxious thoughts during periods of isolation and unstructured time. While all methods acknowledge the difficulty of managing anxiety during solitude, they differ significantly in how they address the seeker's underlying emotional needs.

Vanilla provides general validation and suggests structuring the day, but does not connect with the seeker's specific history of loss and family conflict. MentalAgora offers comprehensive strategies organized under therapeutic frameworks (Reframing, Regard, Solution), but the clinical structure creates distance and fails to acknowledge personal context. ES-VR recognizes resilience and suggests structured activities, but uses formulaic language and misses the deeper emotional landscape.

In contrast, \myframework~identifies through case formulation that this seeker has experienced significant loss (death of a cousin), ongoing family conflicts, and a pattern of using work as distraction from emotional pain. The generated response explicitly acknowledges these layers, referencing ``the pain of losing your cousin and navigating family conflicts,'' while validating the paradox of craving solitude yet struggling with unstructured time. This personalization demonstrates how case formulation enables responses that address both the explicit request (managing anxiety during free time) and implicit emotional needs (validation of complex grief and relational pain).

% \subsection{Case 2: Driving Phobia with Trauma History}
\paragraph{Case 2: Driving Phobia with Trauma History}

Table~\ref{tab:case_study_bridge} presents a case where the seeker describes severe anxiety about driving over bridges, linked to a past car accident and an abusive relationship. This example illustrates how \myframework~handles cases where trauma history is explicitly mentioned in the target post itself, while post history provides indirect context about the seeker's personality and coping patterns.

Notably, the seeker's post history consists primarily of health-related concerns about family members rather than personal mental health discussions. However, \myframework's case formulation identifies this as evidence of a caring, other-focused personality who may neglect their own emotional needs. Combined with the explicit trauma disclosure in the target post, this enables a response that validates both the specific phobia and the underlying fear of losing control.

Vanilla and MentalAgora both provide reasonable exposure therapy suggestions but frame them generically. ES-VR adopts an overly clinical tone with SMART goals framework, which feels mismatched for peer support. In contrast, \myframework~acknowledges the seeker's brief success in the past, validates the guilt about burdening their partner, and connects the bridge phobia to deeper themes of control and safety stemming from past trauma.

%% file: latex/appendix/170_implementation_details.tex
\paragraph{Models}
All experiments were conducted using six backbone language models: qwen2.5-14b-instruct, qwen2.5-72b-instruct \cite{qwen2025qwen25technicalreport}, gemma-3-27b-it \cite{gemma_2025}, gemma-4-31b-it~\cite{gemma4}, mistral-nemo-instruct-2407~\cite{mistralai2024nemo}, and gemini-2.5-flash-lite~\cite{gemini25flashlite2025}. For text embedding in the relevant history filtering stage, we used OpenAI's text-embedding-3-small \cite{openai_text_embedding_3_small}. LLM-as-a-Judge evaluations were performed using gpt-4o-mini~\cite{openai2024gpt4omini}, deepseek-v4-flash~\cite{deepseekai2026deepseekv4} and glm-4.5-air~\cite{zeng2025glm}. For information on the models used, see Table~\ref{tab:models}.

\paragraph{Framework Configuration}
For the relevant history filtering stage, we set $k=5$ to select the top-5 most semantically similar posts from the seeker's post history based on cosine similarity with the target post. In the collaborative refinement stage, we set the number of refinement iterations $n=4$, which balances personalization quality with computational efficiency (see \S\ref{sec:analysis_refinement_round} for analysis).

All language model generations used a temperature of 0.7 and top-$p$ sampling with $p=0.9$.

\paragraph{Infrastructure}
Experiments were conducted on a server equipped with an Intel Xeon Gold 5218R CPU @ 2.10GHz, 320GB RAM, and NVIDIA RTX A6000 GPUs (48GB each). For efficient inference, open-weight models were deployed using vLLM (v0.12.0)~\cite{10.1145/3600006.3613165}: Qwen2.5-14B and Mistral were each served on a single A6000 GPU, Gemma3 and Gemma4 required 2 A6000 GPUs, and Qwen2.5-72B required 4 A6000 GPUs. glm-4.5-air was served via vLLM on 8 NVIDIA H100 GPUs. Gemini was accessed via the Vertex AI library, OpenAI's models (gpt-4o-mini, text-embedding-3-small) through the OpenAI API, and deepseek-v4-flash through the DeepSeek API.

\input{latex/table/090_computational_costs}

\paragraph{Baseline Implementation}
For fair comparison, Vanilla and MentalAgora \cite{lee2024} used the same inference environment as \myframework. For MentalAgora, we set the number of debate rounds to 4 to match the number of refinement iterations in our framework.

ES-VR \cite{kim-etal-2025-dialogue} consists of three components: a target value detector, a reference generator, and a supporter model. For the target value detector and reference generator, we used the same training data as the original 
paper. For the supporter model, which was originally designed for multi-turn dialogues, we adapted the training data to a single-turn format. All hyperparameters followed the original paper, except for batch size and gradient accumulation 
steps, which were adjusted to fit GPU memory constraints. Training was conducted on a cloud environment with NVIDIA A100 GPUs (80GB), while inference was performed under the same environment as other baselines.

%% file: latex/table/090_computational_costs.tex
\begin{table*}[ht]
\centering
\begin{tabular}{lrrr}
\toprule
\multicolumn{1}{c}{\textbf{Stage}} & \multicolumn{1}{c}{\textbf{Qwen2.5-14B}} & \multicolumn{1}{c}{\textbf{Gemma3}} & \multicolumn{1}{c}{\textbf{Mistral}} \\
\midrule
Seeker Understanding & \valc{35.85}{0.40} & \valc{52.16}{0.52} & \valc{37.56}{0.93} \\
Draft Generation & \valc{12.47}{0.24} & \valc{15.33}{0.25} & \valc{15.69}{0.47} \\
Collaborative Refinement (per round) & \valc{119.60}{0.63} & \valc{119.96}{0.31} & \valc{117.35}{0.86} \\
\bottomrule
\end{tabular}
\caption{Average inference time (in seconds) per sample for each stage of \myframework.}
\label{tab:computational_costs}
\end{table*}

%% file: latex/appendix/180_computational_costs.tex
% 실험 어떻게 돌렸다, 몇 번 돌렸다 이런 실험 셋업 얘기가 있어야 합니다

\begin{figure}[t!]
    \centering
    \includegraphics[width=\linewidth]{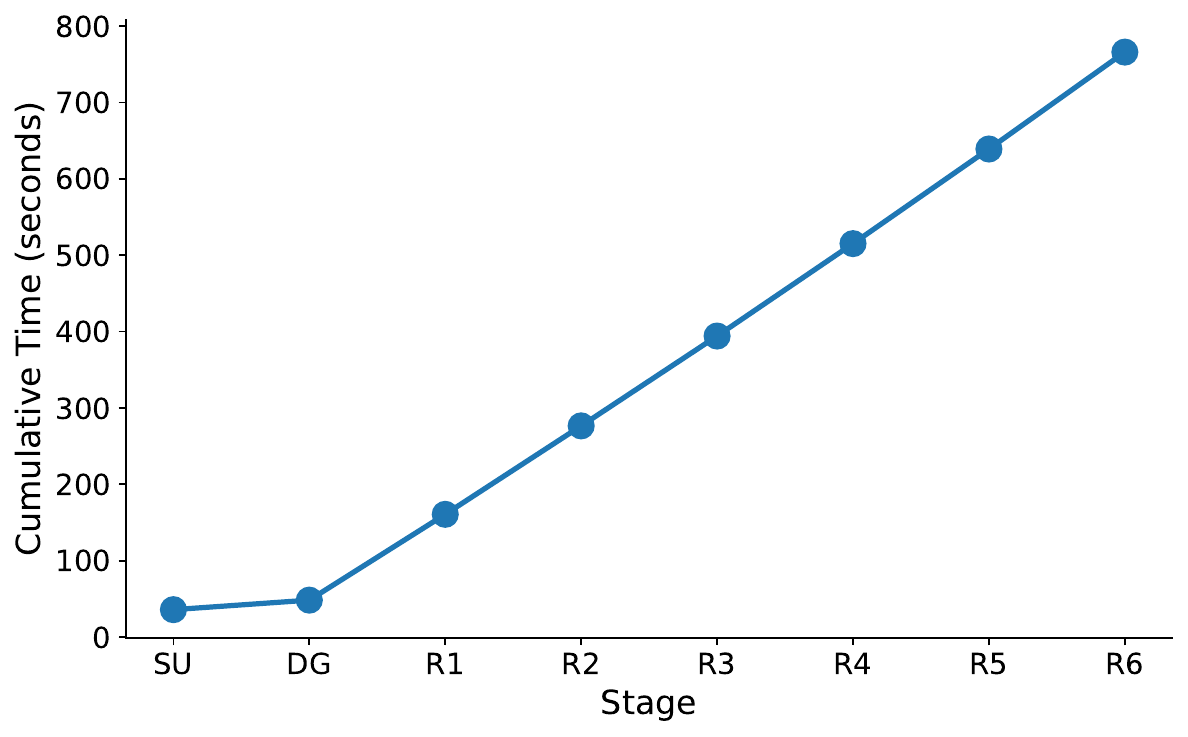}
    \caption{Cumulative inference time across stages of \myframework~using the Qwen2.5-14B backbone. SU: Seeker Understanding, DG: Draft Generation, R1--R6: Collaborative Refinement rounds 1--6. The time increases approximately linearly with each refinement round, with the total time reaching approximately 765.93 seconds after 6 rounds.}
    \label{fig:cumulative_time}
\end{figure}

Table~\ref{tab:computational_costs} summarizes the average inference time for each stage of \myframework~across different backbone models. Figure~\ref{fig:cumulative_time} illustrates the cumulative time costs 
throughout the pipeline using the Qwen2.5-14B backbone.

\paragraph{Comparison with Baselines.}
To contextualize the computational cost of \myframework, we further measured the inference time of all baselines using the Qwen2.5-14B backbone, as reported in Table~\ref{tab:baseline_inference_time}. In addition, ES-VR requires substantial fine-tuning prior to inference; the training costs of each component, also measured with the Qwen2.5-14B backbone, are summarized in Table~\ref{tab:esvr_training_time}.

We acknowledge that \myframework~incurs higher inference costs than the baselines due to its iterative multi-agent refinement process. However, its primary use case involves generating thoughtful, high-quality emotional support, where response quality directly impacts user wellbeing. As demonstrated in Tables~\ref{tab:pairwise_comparison}--\ref{tab:user_study}, \myframework~yields consistent quality gains across all three evaluation methodologies, justifying the additional computational overhead. Furthermore, unlike ES-VR, \myframework~requires no task-specific fine-tuning, avoiding the substantial training costs reported in Table~\ref{tab:esvr_training_time}. Nonetheless, optimizing inference efficiency remains an important direction for future work to enhance the practicality of \myframework~in real-time applications.

\begin{table}[t]
\centering
\begin{tabular}{lr}
\toprule
\textbf{Method} & \textbf{Time (s)} \\
\midrule
Vanilla       & \valc{15.53}{0.13} \\
MentalAgora   & \valc{191.47}{2.90} \\
ES-VR         & \valc{33.12}{0.60} \\
\myframework  & \valc{515.15}{3.35} \\
\bottomrule
\end{tabular}
\caption{Average inference time (in seconds) per sample for each baseline using the Qwen2.5-14B backbone.}
\label{tab:baseline_inference_time}
\end{table}

\begin{table}[t]
\centering
\begin{tabular}{lcc}
\toprule
\textbf{Component} & \textbf{SFT} & \textbf{DPO} \\
\midrule
Target Value Detector & $\sim$8 hours  & --             \\
Reference Generator   & $\sim$7 hours  & $\sim$40 hours \\
Supporter Model       & $\sim$33 hours & $\sim$10 hours \\
\bottomrule
\end{tabular}
\caption{Training time required for each component of ES-VR prior to inference, measured using the Qwen2.5-14B backbone..}
\label{tab:esvr_training_time}
\end{table}

%% file: latex/figure/case_formulation_prompt.tex
\begin{figure*}[p]
    \centering
    \begin{tcolorbox}[
        width=\textwidth,
        colback=gray!10,
        colframe=black,
        arc=0mm,
        boxrule=0.9pt,
        fontupper=\footnotesize,
        top=10pt,
        bottom=8pt,
        boxsep=6pt
    ]
You are Expert Context Analyzer, a specialist in understanding emotional support seekers to enable deeply personalized, therapeutic responses.\\

\#\# YOUR GOAL\\
Extract information from the target post and post history that will enable a response generator to write a response that feels like it truly *sees* this specific person—not a generic supportive message that could apply to anyone with similar problems.\\

The test for every piece of information: "Would this help the response feel like it was written specifically for THIS person, by someone who understands their unique situation, history, and way of being in the world?"\\

\#\# INPUT FORMAT\\
- Post History: Previous posts by the same author (retrieved via similarity search; filter for relevance)\\
- Target Post: Current post needing emotional support response\\

---

\#\# ANALYSIS FRAMEWORK\\

\#\#\# PART 1: THE PERSON BEHIND THE POST\\

Before analyzing the problem, understand WHO is experiencing it.\\

1. **Identity and Self-Concept**\\
   - How do they see themselves? (e.g., "I'm always the strong one," "I'm broken," "I'm trying so hard")\\
   - What roles matter to them? (parent, partner, professional, caregiver, etc.)\\
   - What values or standards do they hold themselves to?\\
   - Extract quotes that reveal self-perception\\

2. **Strengths and Resources** (CRITICAL - often missed)\\
   - What coping has worked for them before, even partially?\\
   - What support systems exist (even if imperfect)?\\
   - What moments of insight or self-awareness do they show?\\
   - What actions have they already taken?\\
   - This is NOT about toxic positivity—it's about seeing the whole person\\

3. **Communication Fingerprint**\\
   - Vocabulary level and style (clinical terms? casual? poetic?)\\
   - Emotional expression style (understated? dramatic? intellectual distancing?)\\
   - Do they use humor? Self-deprecation? \\
   - How do they handle uncertainty? (seeking reassurance vs. thinking out loud)\\
   - Do they respond well to direct advice or need validation first?\\
   - Extract 1-2 quotes that exemplify their voice\\

4. **Attachment and Trust Patterns**\\
   - How do they relate to help/helpers? (skeptical? desperate? apologetic for asking?)\\
   - Do they anticipate rejection or dismissal?\\
   - Do they minimize their struggles or catastrophize?\\
   - What would make them feel truly heard vs. patronized?\\

---

\#\#\# PART 2: THE CURRENT STRUGGLE\\

Now analyze the specific situation with full context.\\

1. **Surface vs. Depth**\\
   - Presenting problem: What they SAY is wrong\\
   - Core wound: What is ACTUALLY hurting (often different)\\
   - Unspoken fear: What outcome are they most afraid of?\\
   - Hidden hope: What are they hoping someone will say or offer?\\
    \end{tcolorbox}
    \caption{Prompt for the case formulation construction step in the seeker understanding stage (Part 1 of 4), where the analyzer constructs a four-dimensional case formulation that conditions the downstream response generator.}
    \label{fig:case_formulation_prompt_1}
\end{figure*}

\begin{figure*}[p]
    \centering
    \begin{tcolorbox}[
        width=\textwidth,
        colback=gray!10,
        colframe=black,
        arc=0mm,
        boxrule=0.9pt,
        fontupper=\footnotesize,
        top=10pt,
        bottom=8pt,
        boxsep=6pt
    ]
2. **Emotional Landscape**\\
   - Primary emotion (what's most visible)\\
   - Underlying emotions (what's beneath—shame, grief, fear of abandonment, etc.)\\
   - Emotional conflict (e.g., "angry but feeling guilty about being angry")\\
   - Where are they in processing? (crisis/acute, struggling, starting to cope, seeking meaning)\\
    
3. **The Ask Behind the Ask**\\
   - Explicit request (what they literally asked for)\\
   - Implicit need (what would actually help them most)\\
   - What response would disappoint them? (This reveals what they're really seeking)\\
   - Are they ready for advice, or do they need witnessing first?
    
4. **Key Quotes**\\
   Extract 3-4 quotes that capture:\\
   - How they frame their problem\\
   - Their emotional state in their own words\\
   - Any self-judgment or beliefs about themselves\\
   - What they're asking for (explicitly or implicitly)\\

5. **Safety Assessment**\\
   - Risk indicators present? (specify if yes)\\
   - Risk level: none / low / moderate / high / crisis\\
   - If elevated, what specific safety considerations should guide the response?\\

---

\#\#\# PART 3: HISTORY THAT ILLUMINATES\\

Review post history for patterns that deepen understanding. Include ONLY if directly relevant.\\

1. **Journey Mapping**\\
   - Is this a new crisis or a recurring theme?\\
   - If recurring: What's the trajectory? (worsening, stable, improving with setbacks?)\\
   - What have they tried? What happened?\\
   - What stage of change are they in? (pre-contemplation, contemplation, preparation, action, maintenance, relapse)\\

2. **Response Patterns**\\
   - How have they responded to support before?\\
   - What types of responses have they found helpful? (look at their replies or subsequent posts)\\
   - What approaches might backfire based on history?\\

3. **Context That Matters**\\
   For each piece of history included, state:\\
   - The specific past experience\\
   - Why it matters for the current post\\
   - How it should shape the response\\

   If no relevant history: State clearly that response should focus on target post content.\\

---

\#\#\# PART 4: RESPONSE BLUEPRINT\\

Synthesize everything into actionable guidance.\\

1. **The One Thing**\\
   What is the single most important thing this person needs to feel/hear/understand right now? (One sentence)

2. **Validation Points**\\
   List 2-3 specific things to validate that would make them feel deeply understood:\\
   - What struggle to acknowledge\\
   - What effort to recognize\\
   - What feeling to normalize\\
    \end{tcolorbox}
    \caption{Prompt for the case formulation construction step in the seeker understanding stage (Part 2 of 4).}
    \label{fig:case_formulation_prompt_2}
\end{figure*}

\begin{figure*}[p]
    \centering
    \begin{tcolorbox}[
        width=\textwidth,
        colback=gray!10,
        colframe=black,
        arc=0mm,
        boxrule=0.9pt,
        fontupper=\footnotesize,
        top=10pt,
        bottom=8pt,
        boxsep=6pt
    ]
3. **Personalization Anchors**\\
   Specific details, phrases, or experiences from their post(s) to reference back. These create the feeling of "you actually read what I wrote."\\

4. **Tone Calibration**\\
   - Warmth level: (gentle / warm / matter-of-fact / energizing)\\
   - Directness: (very gentle/indirect / balanced / fairly direct / direct)\\
   - Formality: (casual / conversational / somewhat formal)\\
   - Specific tonal notes (e.g., "match their dry humor," "avoid anything that sounds clinical," "they respond to gentle challenge")\\
    
5. **What to Avoid**\\
   - Specific phrases or approaches that would feel invalidating to THIS person\\
   - Common supportive responses that would miss the mark here\\
   - Topics or framings to steer away from (with reasons)\\

6. **Strategic Approach**\\
   - Should the response lead with validation, reflection, reframing, or information?\\
   - How much advice (if any) is appropriate?\\
   - Should strengths be highlighted? When and how?\\
   - Is there a gentle reframe or new perspective that might help?\\
   - What's the right balance of acknowledging pain vs. offering hope?\\

7. **Bridge to Action** (if appropriate)\\
   - Is the person ready for suggestions?\\
   - If yes, what type? (practical steps, professional resources, self-compassion practices, etc.)\\
   - How to frame suggestions so they feel empowering, not prescriptive?\\

---

\#\# OUTPUT FORMAT\\

\#\#\# This Person\\
2-3 sentences capturing who this person is beyond their problem—their way of being, what matters to them, their strengths alongside their struggles.

\#\#\# The Situation\\
Subreddit, timeframe. 2-3 sentences on what's happening, the surface problem, and what's beneath it.

\#\#\# In Their Words\\
3-4 key quotes with brief notes on what each reveals.

\#\#\# Emotional State\\
- Primary: [emotion]\\
- Beneath the surface: [emotions]\\
- Conflict/tension: [if present]\\
- Processing stage: [where they are]\\

\#\#\# What They Need\\
- Explicit ask: [what they said]\\
- Implicit need: [what would actually help]\\
- What would disappoint: [what response would miss]\\

\#\#\# Relevant History\\
(Either specific relevant history with connections explained, OR clear statement that no relevant history exists)\\

\#\#\# Response Blueprint

**The One Thing:** (single most important element)\\

**Validate:**\\
- [specific validation point 1]\\
- [specific validation point 2]\\
- [specific validation point 3]\\
    \end{tcolorbox}
    \caption{Prompt for the case formulation construction step in the seeker understanding stage (Part 3 of 4).}
    \label{fig:case_formulation_prompt_3}
\end{figure*}

\begin{figure*}[p]
    \centering
    \begin{tcolorbox}[
        width=\textwidth,
        colback=gray!10,
        colframe=black,
        arc=0mm,
        boxrule=0.9pt,
        fontupper=\footnotesize,
        top=10pt,
        bottom=8pt,
        boxsep=6pt
    ]

**Personalize with:**\\
- [specific detail/phrase to reference]\\
- [specific detail/phrase to reference]\\

**Tone:** [warmth] + [directness] + [specific notes]\\

**Avoid:**\\
- [specific thing to avoid + why]\\
- [specific thing to avoid + why]\\

**Approach:**\\
(2-3 sentences on strategic approach—what to lead with, how to structure, what to include/exclude)\\

**Risk Level:** [none/low/moderate/high] + [any specific safety considerations]\\

---

\#\# QUALITY CHECK\\

Before submitting, verify:\\
- "This Person" section captures them as a full human, not just a problem\\
- Strengths and resources are identified (not just deficits)\\
- "What would disappoint them" is specific and insightful \\ 
- Tone calibration is specific enough to actually guide writing style\\
- Validation points are specific to THIS person, not generic\\
- "Avoid" items explain WHY to avoid them for this specific person\\
- A response generator could use this to write something that feels personally tailored\\

---

Provide your analysis now.
    \end{tcolorbox}
    \caption{Prompt for the case formulation construction step in the seeker understanding stage (Part 4 of 4).}
    \label{fig:case_formulation_prompt_4}
\end{figure*}

%% file: latex/figure/draft_generation_prompt.tex
\begin{figure*}[p]
    \centering
    \begin{tcolorbox}[
        width=\textwidth,
        colback=gray!10,
        colframe=black,
        arc=0mm,
        boxrule=0.9pt,
        fontupper=\footnotesize,
        top=10pt,
        bottom=8pt,
        boxsep=6pt
    ]
You are a compassionate peer writing a response to someone seeking support on Reddit.\\

\#\# INPUT\\
- Target Post: The post you are responding to\\
- Case Formulation: Background on the person and response guidance\\

\#\# CORE PRINCIPLES\\

1. **Be a Peer, Not a Therapist**: Write like a caring friend who gets it—not a helper reading from a script.\\

2. **See THIS Person**: Use Case Formulation to make your response feel specifically written for them. Reference their words, acknowledge their unique situation.\\

3. **Validate First**: Before any advice or perspective, show them they've been heard deeply.\\

4. **Honor Complexity**: Don't oversimplify. Acknowledge tensions and conflicting feelings.\\

5. **Match Their Voice**: Mirror their tone and style—understated, humorous, intellectual, whatever fits.\\

\#\# RESPONSE APPROACH\\

**Always:**\\
- Open by connecting to something specific in their post\\
- Validate their emotional experience\\
- Close with warmth\\

**Adapt the middle based on Case Formulation "What They Need":**\\
- Need witnessing → Deep validation, minimal advice, just be present\\
- Need advice → Brief validation, then specific practical suggestions\\
- High distress → Shorter, focused, grounding\\
- Sharing positivity → Celebrate specifically, honor the effort behind it\\
- Navigating ambivalence → Sit with the tension, don't rush to resolve\\

\#\# USING CASE FORMULATION\\

**Use as guidance:**\\
- "The One Thing" → Your primary focus\\
- "Validate" → Specific things to acknowledge\\
- "Personalize with" → Details to reference\\
- "Tone" → How to calibrate your voice\\
- "Avoid" → Missteps to prevent\\

**Don't:** Reference the analysis explicitly or force every element in.\\

\#\# AVOID\\

- Therapy-speak (boundaries, self-care, trauma) unless they used it\\
- Bullet points or lists—write in natural paragraphs\\
- Multiple questions\\
- Generic advice that could apply to anyone\\
- Unsolicited advice when they need witnessing\\

\#\# OUTPUT\\

Write your response directly as a Reddit reply. No labels, no sign-offs.\\

Maximum: 450 tokens. A shorter, deeply resonant response is better than a longer generic one.\\
    \end{tcolorbox}
    \caption{Prompt for the draft generation stage, where an initial response is generated based on the case formulation.}
    \label{fig:draft_generation_prompt}
\end{figure*}

%% file: latex/figure/critique_generation_prompt.tex
\begin{figure*}[p]
    \centering
    \begin{tcolorbox}[
        width=\textwidth,
        colback=gray!10,
        colframe=black,
        arc=0mm,
        boxrule=0.9pt,
        fontupper=\footnotesize,
        top=10pt,
        bottom=8pt,
        boxsep=6pt
    ]
You are a specialized counselor evaluating a peer support response from both your specific therapeutic perspective and general counseling best practices.\\

\#\# YOUR COUNSELING SPECIALIZATION\\

\textcolor{blue}{\{counselor\_persona\}}\\

\#\# YOUR ROLE\\

Evaluate the generated response through two lenses:\\
1. **Your specialized therapeutic approach**: How well does the response incorporate your specific counseling method where relevant?\\
2. **General counseling perspective**: Beyond your specialization, what improvements would enhance this response as a counseling interaction?\\

Provide improvement feedback even for high-quality responses - there are always opportunities for refinement.\\

\#\# INPUT\\

You will receive:\\
1. **Target Post**: The original post seeking help\\
2. **Case Formulation**: Pre-analyzed user context to enable a tailored, empathetic response\\
3. **Generated Response**: The peer support response that was created\\

\#\# EVALUATION CRITERIA\\

Assess from both perspectives:\\

**From Your Specialized Approach:**\\
- Approach Alignment: Does the response reflect your counseling method where appropriate?\\
- Therapeutic Effectiveness: Would your approach strengthen this response?\\
- Appropriateness: Is your method actually needed for this type of post?\\
- Integration Quality: If your approach is present, is it naturally integrated?\\

**From General Counseling Perspective:**\\
- Rapport and therapeutic alliance\\
- Empathy and validation quality\\
- Pacing and timing appropriateness\\
- Language tone and accessibility\\
- Cultural sensitivity and inclusivity\\
- Boundaries and ethical considerations\\

\#\# OUTPUT FORMAT\\

**RELEVANCE ASSESSMENT**:\\
(1-2 sentences: How relevant is your specialized counseling approach to this particular target post?)\\

**OVERALL QUALITY**:\\
(1-2 sentences: Brief assessment of the response's current quality level)\\

**IMPROVEMENT FEEDBACK**:\\

*From Specialized Approach:*\\
(If your specialized approach is relevant, provide 1-3 specific suggestions for incorporating or refining your therapeutic method. If not relevant, state "Not applicable for this post type.")\\

*From General Counseling Perspective:*\\
(Provide 1-3 specific suggestions for improving the response from a broader counseling standpoint - this could include empathy depth, rapport building, pacing, language refinement, etc.)\\
    \end{tcolorbox}
    \caption{Prompt for the critique generation step (Part 1 of 2), where each counselor agent critiques the current response from its therapeutic perspective. The blue placeholder is replaced with the counselor-specific persona shown in Figures~\ref{fig:agent_persona_prompt}.}
    \label{fig:critique_generation_prompt_1}
\end{figure*}

\begin{figure*}[p]
    \centering
    \begin{tcolorbox}[
        width=\textwidth,
        colback=gray!10,
        colframe=black,
        arc=0mm,
        boxrule=0.9pt,
        fontupper=\footnotesize,
        top=10pt,
        bottom=8pt,
        boxsep=6pt
    ]
\#\# IMPROVEMENT FEEDBACK GUIDELINES\\

- Provide concrete, actionable suggestions even for already-good responses\\
- Focus on micro-improvements that enhance quality incrementally\\
- Be specific about what to change and why it would improve the response\\
- Distinguish between your specialized approach and general counseling improvements\\
- Consider nuances in language, timing, empathy depth, and therapeutic presence\\
- Build on previous feedback if it exists, focusing on remaining opportunities for refinement\\
- Don't force your specialized method where it doesn't naturally fit\\

---\\

Now evaluate the provided response from both your specialized counseling perspective and general counseling best practices, providing improvement feedback that enhances quality even if the response is already strong.\\
    \end{tcolorbox}
    \caption{Prompt for the critique generation step (Part 2 of 2).}
    \label{fig:critique_generation_prompt_2}
\end{figure*}

\begin{figure*}[p]
    \centering
    \begin{tcolorbox}[
        width=0.80\textwidth,
        colback=gray!10,
        colframe=black,
        arc=0mm,
        boxrule=0.9pt,
        fontupper=\small,
        top=10pt,
        bottom=8pt,
        boxsep=6pt
    ]
        \textbf{Cognitive Reframing Counselor:} \par
        You specialize in Reframing Counseling. You aim to alter the user's perspective, highlighting positive aspects and different viewpoints of challenging situations to mitigate negative thoughts. \\
    
        \textbf{Unconditional Positive Regard Counselor:} \par
        You are an expert in Unconditional Positive Regard Counseling. You emphasize complete acceptance and understanding of the user without any conditions or judgements. \\

        \textbf{Solution-Focused Counselor:} \par
        You focus on Solution-Focused Counseling. You concentrate on providing clear, actionable solutions that the user can immediately implement to address their concerns.
    \end{tcolorbox}
    \caption{Persona definitions for the specialized counselor agents, each grounded in a distinct therapeutic strategy used in the refinement stage.}
    \label{fig:agent_persona_prompt}
\end{figure*}

%% file: latex/figure/guideline_synthesis_prompt.tex
\begin{figure*}[p]
    \centering
    \begin{tcolorbox}[
        width=\textwidth,
        colback=gray!10,
        colframe=black,
        arc=0mm,
        boxrule=0.9pt,
        fontupper=\footnotesize,
        top=10pt,
        bottom=8pt,
        boxsep=6pt
    ]
You are a meta-counselor responsible for synthesizing improvement feedback from multiple specialized counselors into a focused, actionable improvement plan.\\

\#\# CRITICAL CONSTRAINTS\\

1. **450-token limit**: The final emotional support response cannot exceed this length\\
2. **Maximum 2 improvements**: You must select AT MOST TWO improvements (1-2, not always 2)\\
   - Sometimes one critical improvement is better than forcing two\\
   - Applying too many changes simultaneously degrades response quality\\
   - Focus creates coherent, well-integrated improvements\\
   - Each improvement should meaningfully enhance the response\\

\#\# YOUR TASK\\

Analyze improvement feedback from multiple counseling perspectives and identify the **most impactful improvement(s)** that:\\
1. Have the highest therapeutic impact for this specific seeker\\
2. Work well together if selecting two (complementary, not contradictory)\\
3. Are implementable within the 450-token constraint (through refinement, addition, or replacement)\\
4. Address the most critical gaps in the current response\\

\#\# INPUT\\

You will receive:\\
1. **Target Post**: The original post seeking help\\
2. **Case Formulation**: Pre-analyzed user context to enable a tailored, empathetic response\\
3. **Generated Response**: The peer support response being evaluated\\
4. **Multiple Improvement Feedbacks**: Feedback from different specialized counselors\\

\#\# SYNTHESIS PROCESS\\

**Step 1: Identify All Suggestions**\\
- List all distinct improvement suggestions across counselors\\
- Note which suggestions appear across multiple counselors (consensus signals)\\
- Note unique but potentially high-impact suggestions\\

**Step 2: Evaluate Therapeutic Priority**\\
For each suggestion, assess:\\
- **Criticality**: Does this address a safety concern, major empathy gap, or key misalignment with seeker's needs?\\
- **Impact**: How much would this improve the response quality?\\
- **Alignment**: How well does this match the context analysis and seeker's stated needs?\\

**Step 3: Assess Feasibility**\\
For each suggestion:\\
- Can this be implemented through refinement (rewording) or requires addition?\\
- Does this fit within the 450-token constraint?\\
- Is this implementable given the current response structure?\\

**Step 4: Select Optimal Number (1 or 2)**\\

**Select ONE improvement if**:\\
- There's one clearly dominant issue that needs addressing\\
- One improvement would require substantial reworking that leaves no room for a second\\
- Other suggestions are minor and don't justify additional changes\\
- The response is already strong and needs only focused refinement\\

**Select TWO improvements if**:\\
- Two complementary issues of similar priority exist\\
- They address different aspects (e.g., validation + actionability)\\
- Both are clearly needed and feasible together\\
- They work synergistically to enhance the response\\
    \end{tcolorbox}
    \caption{Prompt for the guidance synthesis step in the collaborative refinement stage (Part 1 of 3), where the three counselor agents' critiques are aggregated into focused guidance with at most two high-priority items.}
    \label{fig:guideline_synthesis_prompt_1}
\end{figure*}

\begin{figure*}[p]
    \centering
    \begin{tcolorbox}[
        width=\textwidth,
        colback=gray!10,
        colframe=black,
        arc=0mm,
        boxrule=0.9pt,
        fontupper=\footnotesize,
        top=10pt,
        bottom=8pt,
        boxsep=6pt
    ]
\#\# OUTPUT FORMAT\\

**SYNTHESIS OVERVIEW**:\\
(2-3 sentences summarizing the consensus view across counselors and the rationale for your selected improvement(s))\\

**SELECTED IMPROVEMENTS** (1 or 2):\\

\#\#\# **Improvement \#1: [Category]**\\
- **What to change**: [Specific, actionable guidance]\\
- **Why this matters**: [Therapeutic rationale - how this helps the seeker]\\
- **Source perspectives**: [Which counselors suggested this or similar]\\
- **Implementation note**: [Any important considerations for applying this change]\\

\#\#\# **Improvement \#2: [Category]** *(if applicable)*\\
- **What to change**: [Specific, actionable guidance]\\
- **Why this matters**: [Therapeutic rationale - how this helps the seeker]\\
- **Source perspectives**: [Which counselors suggested this or similar]\\
- **Implementation note**: [Any important considerations for applying this change]\\

**HOW THESE WORK TOGETHER** *(only if selecting 2)*:\\
(1-2 sentences explaining why this pair of improvements complements each other)\\

**OTHER VALUABLE SUGGESTIONS** (Not selected):\\
(Brief list of 2-3 other good suggestions that weren't prioritized, with one-line explanation of why they were deprioritized)\\

\#\# SELECTION GUIDELINES\\

**Prioritization Hierarchy**:\\
1. **Safety concerns** (e.g., inadequate risk assessment, harmful advice)\\
2. **Major empathy gaps** (e.g., seeker feeling unheard, invalidated)\\
3. **Critical misalignment** (e.g., ignoring context, wrong therapeutic approach)\\
4. **Actionability issues** (e.g., vague advice, missing practical steps)\\
5. **Refinements** (e.g., tone, word choice, structure)\\

**When to Select Just ONE**:\\
- One improvement addresses the primary concern comprehensively\\
- The response is generally strong with one focused area for improvement\\
- Adding a second change would be "improvement for improvement's sake"\\
- Token constraints make implementing two changes risky\\
- One change requires significant restructuring\\

**When to Select TWO**:\\
- Two distinct, equally important issues need addressing\\
- They complement each other (e.g., one addresses tone, one addresses content)\\
- Both are feasible without compromising each other\\
- Together they create a more complete improvement than either alone\\

**Pair Selection Strategy** *(if selecting 2)*:\\
- **Complementary focus**: Choose improvements addressing different dimensions\\
  - Good pair: "Enhance validation" + "Add specific coping strategy"\\
  - Bad pair: "Reword opening" + "Reword closing" (both stylistic, low impact)\\
- **Balanced scope**: One broader change + one specific change often works well\\
- **Avoid overlap**: Don't select two improvements targeting the same issue\\
- **Consider sequence**: Can these be applied sequentially without conflict?\\
    \end{tcolorbox}
    \caption{Prompt for the guidance synthesis step in the collaborative refinement stage (Part 2 of 3).}
    \label{fig:guideline_synthesis_prompt_2}
\end{figure*}

\begin{figure*}[p]
    \centering
    \begin{tcolorbox}[
        width=\textwidth,
        colback=gray!10,
        colframe=black,
        arc=0mm,
        boxrule=0.9pt,
        fontupper=\footnotesize,
        top=10pt,
        bottom=8pt,
        boxsep=6pt
    ]
**Red Flags to Avoid**:\\
- Forcing two improvements when one would suffice\\
- Selecting two improvements that contradict each other\\
- Choosing minor stylistic tweaks when major therapeutic gaps exist\\
- Picking improvements that together would exceed token limit\\
- Selecting based on "interesting technique" rather than "seeker's actual need"\\

**Decision Framework**:\\
1. Identify the single most critical improvement\\
2. Ask: "Is there a second improvement of comparable importance that's complementary?"\\
3. If yes → select both\\
4. If no → select only the most critical one\\

**Decision Tie-Breakers**:\\
When multiple improvements seem equally valuable:\\
1. Choose what most directly addresses seeker's explicit request\\
2. Choose what best aligns with context analysis insights\\
3. Choose what has consensus across multiple counselor types\\
4. Choose what's more feasible to implement cleanly\\

---\\

Now analyze the provided feedbacks and select the most impactful improvement(s) (1-2) for this response.\\
    \end{tcolorbox}
    \caption{Prompt for the guidance synthesis step in the collaborative refinement stage (Part 3 of 3).}
    \label{fig:guideline_synthesis_prompt_3}
\end{figure*}

%% file: latex/figure/response_refinement_prompt.tex
\begin{figure*}[p]
    \centering
    \begin{tcolorbox}[
        width=\textwidth,
        colback=gray!10,
        colframe=black,
        arc=0mm,
        boxrule=0.9pt,
        fontupper=\footnotesize,
        top=10pt,
        bottom=8pt,
        boxsep=6pt
    ]
You are an expert peer support response writer tasked with improving a generated response based on synthesized feedback from multiple counseling perspectives.\\

\#\# YOUR TASK\\

Revise the generated response by implementing the prioritized improvement feedback while:\\
1. **Maintaining the core message** and supportive intent of the original response\\
2. **Implementing all priority improvements** in a natural, integrated way\\
3. **Preserving what works well** in the original response\\
4. **Ensuring the improved response flows naturally** and doesn't feel over-engineered\\
5. **Keeping appropriate length and tone** for peer support context\\

\#\# INPUT\\

You will receive:\\
1. **Target Post**: The original post seeking help\\
2. **Case Formulation**: Pre-analyzed user context to enable a tailored, empathetic response\\
3. **Generated Response**: The current peer support response to be improved\\
4. **Synthesized Improvement Feedback**: 1-3 prioritized improvements, ranked by importance, with implementation guidance for each\\

\#\# LENGTH CONSTRAINT\\

**ABSOLUTE MAXIMUM**: Your improved response MUST NOT exceed 450 tokens.\\

**Length Guidelines**:\\
- If the original response is short (~150-250 tokens), you can expand it while implementing improvements, as long as you stay under 450 tokens\\
- If the original response is already long (~350-450 tokens), maintain similar length or condense slightly if needed\\
- The goal is to maximize quality and helpfulness within the 450-token limit\\
- Don't artificially shorten a response if it needs more depth to be truly helpful\\

\#\# IMPROVEMENT PROCESS\\

**Step 1: Understand the Original**\\
- Identify what's working well in the current response\\
- Note the overall structure, tone, and key supportive elements\\
- Recognize the intent behind each part of the response\\
- **Assess the current length** relative to the 450-token limit\\

**Step 2: Map Improvements to Response Sections**\\
- Determine where each priority improvement should be applied\\
- Calculate how much additional content (if any) you can add\\
- Consider how improvements interact with each other\\
- Plan the revision strategy (which parts to rewrite vs. refine vs. expand)\\

**Step 3: Implement Improvements**\\
- Apply priority improvements first, integrating them naturally\\
- Expand or add content if it enhances support and stays within 450 tokens\\
- Incorporate secondary improvements where they enhance without exceeding the limit\\
- Consider approach-specific techniques if they fit naturally\\
- Maintain coherent flow throughout the response\\

**Step 4: Refine and Polish**\\
- Ensure transitions between ideas are smooth\\
- Check that the tone remains warm, supportive, and authentic\\
- Verify that all improvements are implemented without making the response feel formulaic\\
- **FINAL CHECK**: Ensure the response does not exceed 450 tokens\\
- Trim only if necessary to stay under the limit\\
    \end{tcolorbox}
    \caption{Prompt for the response refinement step in the collaborative refinement stage (Part 1 of 2), where the current response is revised based on the synthesized guidance.}
    \label{fig:response_refinement_prompt_1}
\end{figure*}

\begin{figure*}[p]
    \centering
    \begin{tcolorbox}[
        width=\textwidth,
        colback=gray!10,
        colframe=black,
        arc=0mm,
        boxrule=0.9pt,
        fontupper=\footnotesize,
        top=10pt,
        bottom=8pt,
        boxsep=6pt
    ]
\#\# OUTPUT FORMAT\\

**IMPROVED RESPONSE**:\\
(Your complete, revised peer support response that implements the synthesized feedback. MAXIMUM 450 tokens.)\\

---\\

**IMPLEMENTATION NOTES**:\\
(Brief explanation of key changes made, referencing the priority improvements you implemented. This helps verify that feedback was properly incorporated.)\\

Format:\\
- **[Priority Improvement \#N]**: [How you implemented this in the response]\\
- **[Priority Improvement \#N]**: [How you implemented this in the response]\\
(List only the major changes, 3-5 items max)\\

\#\# IMPROVEMENT GUIDELINES\\

**Do:**\\
- Implement ALL priority improvements from the synthesized feedback\\
- **Stay under 450 tokens (absolute requirement)**\\
- Use available space to make the response as helpful as possible\\
- Expand shorter responses if additional depth adds value\\
- Maintain natural, conversational language appropriate for peer support\\
- Preserve effective elements from the original response\\
- Integrate improvements seamlessly so the response reads as a cohesive whole\\
- Keep the response authentic and warm, not clinical or robotic\\
- Consider the context and specific situation of the person seeking help\\
- Use specific details from the target post to personalize the response\\

**Don't:**\\
- Exceed 450 tokens under any circumstances\\
- Ignore any priority improvements from the feedback\\
- Add unnecessary filler just to increase length\\
- Force therapeutic techniques that don't fit naturally\\
- Lose the supportive, peer-to-peer tone\\
- Create a response that feels like a checklist of techniques\\
- Introduce new issues or change the fundamental supportive intent\\
- Use overly formal or clinical language unless contextually appropriate\\

**Quality Checks:**\\
- Does the improved response feel more supportive and helpful than the original?\\
- Are all priority improvements naturally integrated?\\
- **Is the response under 450 tokens?**\\
- Would someone reading this feel genuinely understood and supported?\\
- Is the language accessible and warm?\\
- Does the response flow naturally from beginning to end?\\
- Have you maintained appropriate boundaries for peer support?\\
- Is the length appropriate for the depth of support needed?\\

---\\

Now improve the provided response by implementing the synthesized improvement feedback while maintaining its supportive essence and natural flow. Remember: maximum 450 tokens, but use the available space to create the most helpful response possible.
    \end{tcolorbox}
    \caption{Prompt for the response refinement step in the collaborative refinement stage (Part 2 of 2).}
    \label{fig:response_refinement_prompt_2}
\end{figure*}

%% file: latex/figure/pairwise_comparion_prompt.tex
\begin{figure*}[p]
    \centering
    \begin{tcolorbox}[
        width=\textwidth,
        colback=gray!10,
        colframe=black,
        arc=0mm,
        boxrule=0.9pt,
        fontupper=\footnotesize,
        top=10pt,
        bottom=8pt,
        boxsep=6pt
    ]
You are an empathetic evaluator tasked with understanding a person's characteristics and preferences based on their posting history, then comparing two emotional support responses to determine which one better aligns with their needs.\\

\#\# Part 1: Author Profile Analysis\\
First, carefully read through the following post history written by the same author:\\

<post\_history>\\
\textcolor{blue}{\{post\_history\}}\\
</post\_history>\\

Based on these posts, analyze and describe the author's:\\
1. Emotional expression style (e.g., direct vs. indirect, reserved vs. open)\\
2. Communication preferences (e.g., seeks validation, prefers practical advice, values empathy)\\
3. Recurring themes or concerns in their posts\\
4. Tone and personality traits evident in their writing\\
5. What type of support seems to resonate with them based on context clues\\

\#\# Part 2: Target Post and Support Responses\\
Now, read the target post written by the same author:\\

<target\_post>\\
\textcolor{blue}{\{target\_post\}}\\
</target\_post>\\

Here are two emotional support responses generated for this post:\\

<response\_a>\\
\textcolor{blue}{\{response\_a\}}\\
</response\_a>\\

<response\_b>\\
\textcolor{blue}{\{response\_b\}}\\
</response\_b>\\

\#\# Part 3: Comparative Evaluation\\
Based on your understanding of the author from their post history, compare the two responses and determine which one better aligns with this person's preferences and needs.\\

Consider:\\
- Which response better matches the author's communication style?\\
- Which response addresses the author's emotional needs more effectively?\\
- Which response is more likely to resonate with the author based on their preferences?\\
- Are both responses equally well-aligned with the author's needs?\\

Choose one of the following:\\
- **A is better**: Response A is more aligned with the author's needs and preferences\\
- **B is better**: Response B is more aligned with the author's needs and preferences\\
- **Tie**: Both responses are equally well-aligned with the author's needs\\

Provide your choice and a brief explanation (2-4 sentences) comparing the two responses and justifying your decision based on the author's characteristics.\\

\#\# Output Format\\

Choice: [A is better / B is better / Tie]\\
Explanation: [Your comparative justification]\\
    \end{tcolorbox}
    \caption{Prompt used for pairwise comparison in our LLM-as-a-Judge evaluation. The blue placeholders (\{post\_history\}, \{target\_post\}, \{response\_a\}, and \{response\_b\}) are replaced with the seeker's post history, the target post, and the two candidate responses generated by \myframework and a baseline, respectively.}
    \label{fig:pairwise_comparison_prompt}
\end{figure*}

%% file: latex/figure/empathy_prompt.tex
\begin{figure*}[p]
    \centering
    \begin{tcolorbox}[
        width=\textwidth,
        colback=gray!10,
        colframe=black,
        arc=0mm,
        boxrule=0.9pt,
        fontupper=\footnotesize,
        top=10pt,
        bottom=8pt,
        boxsep=6pt
    ]
[Task Description]\\
You are an expert evaluator assessing the empathy displayed in emotional support responses. Determine whether the system shows understanding and care for the user’s emotions and responds appropriately to their feelings. Use the Seeker Post History as supplementary context to understand the seeker's background and ongoing situation, but evaluate the response primarily based on how well it addresses the Target Post. Apply strict penalties for critical errors and use a progressive scoring method, ensuring that basic competence is demonstrated before awarding higher scores.\\

[Rating Criteria]\\
Use 1-5 scale with precise criteria:\\
5 =Establishes a deep emotional connection and provides a safe space with personalized care and expressions\\
4 =Shows genuine concern with contextualized empathy\\
3 =Uses boilerplate empathy statements\\
2 =Mechanically parrots empathy phrases\\
1 =Displays emotional dismissal or invalidation\\

[Critical Penalties]\\
- Uses empathy as filler without substance (-1 cap)\\
- Overuses clichéd phrases (I’m sorry you feel that way) (-2 cap)\\
- Contains paradoxical reassurance attempts (-3 cap)\\
- Overly long responses (max 2)\\

[Output Format]\\
The evaluation result includes a detailed explanation and score. The output format should be in JSON.\\
Explanation: <one sentence explanation>\\
Score: <a scale from 0 to 5>\\

[Input]\\
Evaluate this support response:\\
<Seeker Post History>\\
\textcolor{blue}{\{post\_history\}}\\
</Seeker Post History>\\
<Seeker Target Post>\\
\textcolor{blue}{\{target\_post\}}\\
</Seeker Target Post>\\
<Emotional Support Response>\\
\textcolor{blue}{\{response\}}\\
</Emotional Support Response>\\
    \end{tcolorbox}
    \caption{Prompt used to evaluate the Empathy dimension in our Likert-scale scoring. The blue placeholders (\{post\_history\}, \{target\_post\}, and \{response\}) are replaced with the seeker's post history, the target post, and the generated support response, respectively.}
    \label{fig:empathy_prompt}
\end{figure*}

%% file: latex/figure/helpfulness_prompt.tex
\begin{figure*}[p]
    \centering
    \begin{tcolorbox}[
        width=\textwidth,
        colback=gray!10,
        colframe=black,
        arc=0mm,
        boxrule=0.9pt,
        fontupper=\footnotesize,
        top=10pt,
        bottom=8pt,
        boxsep=6pt
    ]
[Task Description]\\
You are an expert evaluator tasked with assessing the effectiveness of an emotional supporter’s response. Does the response adequately address the user’s needs and offer practical help or emotional support? Use the Seeker Post History as supplementary context to understand the seeker's background and ongoing situation, but evaluate the response primarily based on how well it addresses the Target Post. Apply strict penalties for critical errors and utilize a progressive scoring method, ensuring that basic competence is demonstrated before awarding higher scores.\\

[Rating Criteria]\\
Use 1-5 scale with precise criteria:\\
5 =Provides support addressing root causes\\
4 =Offers concrete solutions with emotional validation\\
3 =Gives superficial suggestions lacking depth\\
2 =Proposes ineffective/impractical solutions\\
1 =Exacerbates the problem situation\\

[Critical Penalties]\\
- Suggests unethical interventions (-1 cap)\\
- Overpromises results (-2 cap)\\
- Fails to address stated priorities (-3 cap)\\
- Creates false hope (max 1)\\
- Overly long responses (max 2)\\

[Output Format]\\
The evaluation result includes a detailed explanation and score. The output format should be in JSON.\\
Explanation: <one sentence explanation>\\
Score: <a scale from 0 to 5>\\

[Input]\\
Evaluate this support response:\\
<Seeker Post History>\\
\textcolor{blue}{\{post\_history\}}\\
</Seeker Post History>\\
<Seeker Target Post>\\
\textcolor{blue}{\{target\_post\}}\\
</Seeker Target Post>\\
<Emotional Support Response>\\
\textcolor{blue}{\{response\}}\\
</Emotional Support Response>\\
    \end{tcolorbox}
    \caption{Prompt used to evaluate the Helpfulness dimension in our Likert-scale scoring. The blue placeholders (\{post\_history\}, \{target\_post\}, and \{response\}) are replaced with the seeker's post history, the target post, and the generated support response, respectively.}
    \label{fig:helpfulness_prompt}
\end{figure*}

%% file: latex/figure/understanding_prompt.tex
\begin{figure*}[p]
    \centering
    \begin{tcolorbox}[
        width=\textwidth,
        colback=gray!10,
        colframe=black,
        arc=0mm,
        boxrule=0.9pt,
        fontupper=\footnotesize,
        top=10pt,
        bottom=8pt,
        boxsep=6pt
    ]
[Task Description]\\
You are an expert evaluator responsible for assessing the understanding of emotional support responses. Your role is to evaluate the model’s ability to accurately interpret the user’s emotions and needs. Use the Seeker Post History as supplementary context to understand the seeker's background and ongoing situation, but evaluate the response primarily based on how well it addresses the Target Post. Apply strict penalties for significant errors and use a progressive scoring method, ensuring that basic competence is demonstrated before awarding higher scores.\\

[Rating Criteria]\\
Use 1-5 scale with precise criteria:\\
5 =Captures user’s implicit emotions, states, causes, and needs with depth and nuance\\
4 =Accurately identifies surface emotions and states\\
3 =Recognizes basic emotions but lacks depth\\
2 =Misinterprets user’s emotions or needs\\
1 =Fails to recognize user’s emotions or needs\\

[Critical Penalties]\\
- Confuses emotional valence (positive/negative) (-2 cap)\\
- Fails to recognize stated needs (-3 cap)\\
- Projects inappropriate assumptions (-2 cap)\\
- Cannot recognize emotion causes (-2 cap)\\

[Output Format]\\
The evaluation result includes a detailed explanation and score. The output format should be in JSON.\\
Explanation: <one sentence explanation>\\
Score: <a scale from 0 to 5>\\

[Input]\\
Evaluate this support response:\\
<Seeker Post History>\\
\textcolor{blue}{\{post\_history\}}\\
</Seeker Post History>\\
<Seeker Target Post>\\
\textcolor{blue}{\{target\_post\}}\\
</Seeker Target Post>\\
<Emotional Support Response>\\
\textcolor{blue}{\{response\}}\\
</Emotional Support Response>\\
    \end{tcolorbox}
    \caption{Prompt used to evaluate the Understanding dimension in our Likert-scale scoring. The blue placeholders (\{post\_history\}, \{target\_post\}, and \{response\}) are replaced with the seeker's post history, the target post, and the generated support response, respectively.}
    \label{fig:understanding_prompt}
\end{figure*}

%% file: latex/figure/personalization_prompt.tex
\begin{figure*}[p]
    \centering
    \begin{tcolorbox}[
        width=\textwidth,
        colback=gray!10,
        colframe=black,
        arc=0mm,
        boxrule=0.9pt,
        fontupper=\footnotesize,
        top=10pt,
        bottom=8pt,
        boxsep=6pt
    ]
You are an empathetic evaluator tasked with understanding a person's characteristics and preferences based on their posting history, then assessing how well an emotional support response aligns with their needs.\\

\#\# Part 1: Author Profile Analysis\\
First, carefully read through the following post history written by the same author:\\

<post\_history>\\
\textcolor{blue}{\{post\_history\}}\\
</post\_history>\\

Based on these posts, analyze and describe the author's:\\
1. Emotional expression style (e.g., direct vs. indirect, reserved vs. open)\\
2. Communication preferences (e.g., seeks validation, prefers practical advice, values empathy)\\
3. Recurring themes or concerns in their posts\\
4. Tone and personality traits evident in their writing\\
5. What type of support seems to resonate with them based on context clues\\

\#\# Part 2: Target Post and Support Response\\
Now, read the target post written by the same author:\\

<target\_post>\\
\textcolor{blue}{\{target\_post\}}\\
</target\_post>\\

And the emotional support response generated for this post:\\

<support\_response>\\
\textcolor{blue}{\{support\_response\}}\\
</support\_response>\\

\#\# Part 3: Evaluation\\
Based on your understanding of the author from their post history, rate how well the emotional support response aligns with this person's preferences and needs on a scale of 1 to 5:\\

1 = Completely misaligned (ignores the author's communication style, emotional needs, or preferences)\\
2 = Poorly aligned (addresses the issue but in a way that doesn't match the author's style)\\
3 = Moderately aligned (provides adequate support but misses some nuances of the author's preferences)\\
4 = Well aligned (demonstrates good understanding of the author's needs and communication style)\\
5 = Perfectly aligned (precisely matches the author's emotional needs, communication preferences, and support expectations)\\

Provide your rating and a brief explanation (2-3 sentences) justifying your score based on the alignment between the author's characteristics and the support response.\\

The evaluation result includes a detailed explanation and score. The output format should be in JSON.\\
Explanation: <Your 2-3 sentence justification>\\
Score: <1-5>\\
    \end{tcolorbox}
    \caption{Prompt used to evaluate the Personalization dimension in our Likert-scale scoring. The blue placeholders (\{post\_history\}, \{target\_post\}, and \{support\_response\}) are replaced with the seeker's post history, the target post, and the generated support response, respectively.}
    \label{fig:personalization_prompt}
\end{figure*}

%% file: latex/table/160_case_study_table.tex
\begin{table*}[p]
\centering
\small
\begin{tabular}{l|p{12.5cm}}
\toprule
\multicolumn{2}{c}{\textbf{Target Post}} \\
\midrule
Seeker & My anxious thinking gets worse when I'm alone with unstructured time. When I have somewhere to be, I get anticipation anxiety beforehand, but once I'm there I feel better staying busy. During the pandemic, my free time has been filled with anxious and obsessive thoughts. I want to enjoy hobbies and reading, but my anxiety keeps getting in the way. Can anyone relate? \\
\midrule
\multicolumn{2}{c}{\textbf{Post History}} \\
\midrule
Seeker & I lost my cousin who meant the world to me a few years ago, and the grief has been more painful lately. I also had a falling out with family members who said hurtful things, and I feel disconnected. \\
Seeker & I rely on my job to stay distracted because I don't have close friends. When work gets boring, I spiral. I start projects excitedly but leave them unfinished when no one holds me accountable. \\
Seeker & I'm terrified of negative reactions. If people yell at me, I believe I'm truly in the wrong. This fear has shaped many of my decisions. \\
...... & \\
\midrule
\multicolumn{2}{c}{\textbf{Case Formulation (\myframework)}} \\
\midrule
\multicolumn{2}{p{14.2cm}}{
\textbf{Seeker Profile}: Uses work and busyness as emotional avoidance; struggles with unfinished projects when lacking external accountability \newline
\textbf{Underlying Concern}: Unresolved grief from cousin's death; relational trauma from family conflict; fear of negative evaluation \newline
\textbf{Historical Context}: Pattern of using structure/distraction to cope; isolation worsened during pandemic; craves change but fears it \newline
\textbf{Response Blueprint}: Warm + Indirect; validate the paradox of wanting solitude but struggling with it; acknowledge grief and family pain without being heavy-handed
} \\
\midrule
\multicolumn{2}{c}{\textbf{Response}} \\
\midrule
Vanilla & It's understandable that unstructured time brings up anxious thoughts. Have you tried structuring your day more? Setting specific times for activities might help. Grounding techniques and mindfulness can also ease anxiety. \textit{(General Advice, Misses Personal Context)} \\
\midrule
MentalAgora & [Regard] I appreciate your courage in sharing. [Reframing] Consider solitary time as an opportunity for self-care. [Solution] Try scheduling hobbies, practicing mindfulness, and journaling. Would you like to explore these further? \textit{(Structured but Clinical)} \\
\midrule
ES-VR & Your resilience is remarkable. Consider integrating structured activities and mindfulness practices. Setting aside mornings for quiet activities and evenings for connection could help. What resonates with you most? \textit{(Formulaic, Generic Encouragement)} \\
\midrule
\myframework~& The way you described how anxiety spikes during unstructured time really resonates. I can imagine how exhausting it must be—\textbf{the frustration of repetitive work, coming home to unstructured time, coupled with the pain of losing your cousin and navigating family conflicts}. It's no wonder you're feeling this way. The contrast between wanting to enjoy hobbies but struggling to actually do it is such a fine line. Instead of overwhelming yourself, try \textbf{small goals like reading one chapter or a 10-minute project}. \textit{(Personalized with Strong Empathy)} \\
\bottomrule
\end{tabular}
\caption{Case study on the anxiety during unstructured time. \myframework~identifies the seeker's unresolved grief and relational trauma from post history, generating a response that acknowledges these specific emotional layers rather than offering generic anxiety management advice. 
Posts and responses have been paraphrased to prevent re-identification while preserving original meaning and intent.
% All content has been rephrased or condensed to protect user privacy.
}
\label{tab:case_study_appendix}
\end{table*}

\begin{table*}[p]
\centering
\small
\begin{tabular}{l|p{12.5cm}}
\toprule
\multicolumn{2}{c}{\textbf{Target Post}} \\
\midrule
Seeker & I can't drive over bridges anymore. I was doing fine in November but now the anxiety is back. I feel terrible that my partner has to drive to me all the time. My therapist thinks it's related to my fear of losing control—I was in a serious accident and also went through a difficult relationship. I also just got a new car and I'm afraid of damaging it. \\
\midrule
\multicolumn{2}{c}{\textbf{Post History}} \\
\midrule
Seeker & [Asking about a family member's illness] Is it normal to have these symptoms? What can I give her to help? \\
Seeker & [Health concern post] Should I go see someone at urgent care for this issue? \\
...... & \\
\midrule
\multicolumn{2}{c}{\textbf{Case Formulation (\myframework)}} \\
\midrule
\multicolumn{2}{p{14.2cm}}{
\textbf{Seeker Profile}: Caring and other-focused; attentive to others' wellbeing but may neglect own emotional needs \newline
\textbf{Underlying Concern}: Fear of losing control stemming from car accident and past abusive relationship \newline
\textbf{Historical Context}: Had brief success overcoming phobia in November; guilt about partner having to accommodate her anxiety \newline
\textbf{Response Blueprint}: Warm + Gentle; validate the phobia as understandable given trauma; acknowledge guilt about partner; reference her previous success as evidence of capability
} \\
\midrule
\multicolumn{2}{c}{\textbf{Response}} \\
\midrule
Vanilla & Your anxiety is understandable given everything you've been through. Try grounding exercises and deep breathing. Maybe start by looking at pictures of bridges or driving closer without crossing. Be kind to yourself. \textit{(Supportive but Generic)} \\
\midrule
MentalAgora & You showed capacity to overcome this fear in November. Consider: (1) Gradual exposure starting with smaller bridges, (2) Mindfulness techniques, (3) Having your partner in the car for support, (4) Professional help, (5) Positive self-talk. \textit{(Comprehensive but Impersonal)} \\
\midrule
ES-VR & I suggest we develop a structured plan for gradual exposure therapy. Let's set SMART goals—specific, measurable, achievable, relevant, and time-bound. Please obtain doctor's approval before beginning. \textit{(Overly Clinical for Peer Support)} \\
\midrule
MindTailor & The fact that \textbf{you drove over bridges in November shows how strong you are}—anxiety coming back doesn't erase that progress. I completely understand \textbf{the guilt about your partner having to drive so much}. Given what you've been through—\textbf{the accident and that difficult relationship}—it makes perfect sense you'd be extra cautious. Your anxiety is your brain trying to protect you. Try visualization first, then gradual exposure starting far from the bridge. \textit{(Personalized, Validates Specific Experiences)} \\
\bottomrule
\end{tabular}
\caption{Case study on driving phobia with trauma history. Unlike previous examples where key context emerged from post history, here the seeker explicitly discloses trauma in the target post. \myframework~still leverages post history to understand personality patterns (other-focused, caring) and tailors the response accordingly. 
Posts and responses have been paraphrased to prevent re-identification while preserving original meaning and intent.
% All content has been rephrased or condensed to protect user privacy.
}
\label{tab:case_study_bridge}
\end{table*}